\documentclass[letterpaper]{article} 
\usepackage[preprint]{aaai2027}  
\usepackage[hyphens]{url}  
\usepackage{graphicx} 
\usepackage{natbib}  
\usepackage{caption} 
\usepackage{algorithm}
\usepackage{algorithmic}

\usepackage{newfloat}
\usepackage{listings}
\DeclareCaptionStyle{ruled}{labelfont=normalfont,labelsep=colon,strut=off} 
\floatstyle{ruled}
\newfloat{listing}{tb}{lst}{}
\floatname{listing}{Listing}

\usepackage{booktabs}

\usepackage{amsfonts}       
\usepackage{nicefrac}       
\usepackage{microtype}      
\usepackage{xcolor}         
\usepackage{amsmath, amssymb}
\usepackage{multirow}
\usepackage{array}
\usepackage{enumitem}
\usepackage{tabularx}

\newcommand{\dw}{\ensuremath{\Delta w}}
\newcommand{\R}{\ensuremath{\mathbb{R}}}
\newcommand{\one}{\ensuremath{\mathbf{1}}}

\newcommand{\npast}{\ensuremath{n_{\mathrm{past}}}}
\newcommand{\DMD}{\ensuremath{\mathrm{DMD}}}

\providecommand{\appdef}{}%
\renewcommand{\appdef}[2]{\expandafter\def\csname app.#1\endcsname{#2}}
\providecommand{\appref}{}%
\renewcommand{\appref}[1]{%
  \ifcsname app.#1\endcsname
    \csname app.#1\endcsname
  \else
    \textbf{[APPREF?\,#1]}%
  \fi}
\appdef{app:experimental_details}{Appendix~A}
\appdef{app:probe_conventions}{Appendix~B}
\appdef{app:spatial_localization}{Appendix~C}
\appdef{app:pocket_persistence}{Appendix~D}
\appdef{app:state_correspondence}{Appendix~E}
\appdef{app:pythia_scale_extensions}{Appendix~F}
\appdef{app:recipe_conditioning}{Appendix~G}
\appdef{app:online}{Appendix~H}

\providecommand{\maindef}{}%
\renewcommand{\maindef}[2]{\expandafter\def\csname main.#1\endcsname{#2}}
\providecommand{\mainref}{}%
\renewcommand{\mainref}[1]{%
  \ifcsname main.#1\endcsname
    \csname main.#1\endcsname
  \else
    \textbf{[MAINREF?\,#1]}%
  \fi}
\maindef{sec:methods}{Section~3}
\maindef{sec:empirical}{Section~4}
\maindef{sec:probe_relations}{Section~4.1}
\maindef{sec:heterogeneity}{Section~4.2}
\maindef{sec:dynamics}{Section~4.3}
\maindef{sec:recipe}{Section~4.4}
\maindef{fig:family_panels}{Figure~1}
\maindef{fig:cross_family}{Figure~2}
\maindef{fig:aux_bulk_slope}{Figure~3}
\maindef{fig:pocket_scarcity}{Figure~4}
\maindef{fig:pythia70m_case}{Figure~5}
\maindef{fig:opt_axis}{Figure~6}
\maindef{eq:frr}{Eq.~(2)}
\maindef{eq:lrgf}{Eq.~(3)}
\maindef{eq:cospred}{Eq.~(4)}
\maindef{eq:sesa}{Eq.~(5)}
\maindef{eq:selectivity_lift}{Eq.~(6)}

\title{Measuring Structured Predictability in Neural Training Dynamics: A Cross-Regime Study}
\author{Fanqi Wang, Weisheng Tang, and Hairong Qi\corresponding}
\affiliations{Department of Electrical Engineering and Computer Science, University of Tennessee, Knoxville}

\begin{document}

\maketitle

\begin{abstract}
Modern deep networks are trained through long update trajectories, yet their temporal organization remains less systematically characterized than architectures, losses, or optimizers. We study short-horizon predictability as a measure of temporal redundancy: where, when, and under which training conditions recent updates contain information about near-future parameter motion. We combine three complementary probe families---displacement-direction, subspace-residual, and predictor-based probes---with convention-aware, null-calibrated
group-level readouts, and apply them to multi-pass vision training on CIFAR and public Pythia pretraining checkpoints. Across both regimes,
vector-like tensors such as normalization parameters and biases (\emph{auxiliary parameters}) exhibit simpler short-horizon dynamics than matrix-like feature-transforming weights (\emph{bulk parameters}), whose predictable behavior concentrates in localized, time-varying pockets. Agreement within and across probe families, and with independent trajectory diagnostics, indicates that these measurements capture intrinsic trajectory structure, while probe differences distinguish complementary forms of temporal organization. Controlled CIFAR comparisons further show that architecture and training recipe systematically modulate the measured structure. A Pythia-70M case study further exposes a sequence of role-, depth-, and scale-dependent events, including bulk ESA falling below the random sign-agreement level and the emergence and redistribution of predictable $qkv$ pockets across layers. These results position short-horizon predictability as a retrospective, parameter-resolved diagnostic of training dynamics.
\end{abstract}

\section{Introduction}
\label{sec:introduction}

Modern neural networks are trained through long sequences of gradient evaluations and parameter updates, making iterative optimization a central computational bottleneck as models, datasets, and training horizons grow. A large body of work has studied architectures, losses, optimizers, final representations, gradient geometry, and loss-landscape structure
\cite{gur2018gradient,li2018measuring,goodfellow2014qualitatively,draxler2018essentially,cohen2021gradient,papyan2020prevalence}.
In contrast, the temporal organization of the training trajectory remains less systematically characterized. Recent trajectory-level work \cite{singh2025directionality} has begun to characterize the global directional geometry of checkpoint paths, yet the short-horizon temporal redundancy of successive updates---the extent to which recent motion is informative about near-future displacement---has not been systematically characterized.

This gap is particularly relevant to predictive acceleration. Weight nowcasting, update extrapolation, predictive differential training, and related methods rely on recent history being informative about near-future parameter evolution \cite{sinha2017introspection,jang2023learning,knyazev2024accelerating, wang2026predictive,guan2024xgrad}. Their success suggests that such information can exist, but not that it is uniform across parameters, training stages, architectures, or optimization recipes; indeed, Predictive Differential Training showed that selective prediction can be more reliable than prediction applied uniformly across a model. This motivates a more basic measurement question: \emph{where, when, and under what training conditions do neural update trajectories exhibit exploitable short-horizon structure?}

To answer this question, we develop three complementary probe families of increasing complexity. Displacement-direction probes test whether recent movement directions persist; subspace-residual probes ask whether future displacement remains within the span of recent updates; and predictor-based probes fit explicit local dynamics to capture richer temporal structure. We evaluate them at aggregate, tensor, and within-tensor group scales, with convention-aware and null-calibrated readouts for comparing localized structure under different grouping choices. Agreement and disagreement across the probes reveal shared and complementary forms of temporal organization.

Across multi-pass vision training on CIFAR and one-pass pretraining of Pythia, the probes reveal a common spatial organization. Auxiliary, vector-like parameters exhibit simpler short-horizon dynamics than bulk, matrix-like parameters. Under the deepest probes considered here, bulk tensors are largely unpredictable in aggregate but contain localized subsets with elevated predictability whose identities shift over training. We call these localized, time-varying subsets \emph{predictable pockets}. Beyond locating this structure, the probes reveal how it evolves with optimization. In controlled vision runs, directional  readouts closely track movement directedness and gradient signal-to-noise ratio; ESA rapidly falls toward an anti-persistent floor, with later departures depending on parameter role and training condition.
In Pythia, temporal changes differ across QKV, attention-output, and MLP parameters and are not reducible to a single event in the scalar loss curve. More broadly, the short-horizon organization of training dynamics is jointly conditioned on optimization position, architecture, and training recipe; its parameter-resolved structure reveals where these differences are expressed within the model.

In summary, our contributions are:
\begin{enumerate}
    \item We introduce a convention-aware, multiscale framework for measuring short-horizon temporal structure through three graded probe families: displacement-direction, subspace-residual, and predictor-based probes. The framework resolves aggregate, tensor, and within-tensor group structure, operates retrospectively on stored checkpoints, and provides a geometric ceiling for predictors constrained to the
    recent-history span.

    \item Across multi-pass vision training and one-pass language-model pretraining, we identify a shared organization of predictability. Auxiliary and bulk parameters separate systematically in short-horizon dynamics, while bulk tensors contain localized, time-varying predictable pockets rather than uniformly predictable structure.

    \item We show that the probe readings provide a state-sensitive, parameter-resolved view of optimization. They track realized movement structure, distinguish role-specific temporal events, and respond systematically to architecture and training recipe.
    These results position short-horizon predictability as a retrospective diagnostic of temporal redundancy, distinct from optimizer quality and only conditionally relevant to predictive reuse.
\end{enumerate}

\section{Related Work and Positioning}
\label{sec:related-work}

Predicting future weights or updates has been explored as a way to accelerate neural training. Introspection learns weight-evolution patterns across runs~\citep{sinha2017introspection}; weight nowcasting and neuron-interaction nowcasting train auxiliary predictors to forecast near-future parameters~\citep{jang2023learning,knyazev2024accelerating}; Predictive Differential Training uses Koopman/DMD-style forecasts with selective training-dynamics masks~\citep{wang2026predictive}; and optimizer-rule-based prediction has also been used to modify gradient-based training~\citep{guan2024xgrad}. DMD and Koopman methods provide a natural linear-dynamical language for time-resolved trajectories~\citep{schmid2010dynamic,tu2014dynamic,dogra2020optimizing}. These works motivate the possibility of predictive training, but they are primarily predictor- or acceleration-centered. They do not systematically ask where short-horizon temporal regularity exists inside a model, whether it is localized or global, or how it changes with parameter role, training stage, architecture, optimizer, and training regime.

The closest work to ours studies the optimization trajectory as an object in its own right: \citet{singh2025directionality} show that the directional geometry of raw checkpoint paths saturates early at the macro level, oscillates at finer scales later, and is shaped in part by weight decay and momentum. We look at a different object. Rather than the path of the raw weights, we study the sequence of updates; rather than describing past geometry, we ask how well the recent past predicts the next few steps; and we answer at aggregate, tensor, and within-tensor scales. Existing trajectory-level studies characterize selected geometric properties of optimization paths, but a multiscale account of short-horizon parameter predictability spanning directional persistence, recent-history subspace structure, and explicit prediction remains missing.

A separate body of work shows that neural training has rich structure. Gradients can align with low-dimensional Hessian subspaces~\citep{gur2018gradient}, random-subspace experiments estimate intrinsic dimension~\citep{li2018measuring}, and recent work questions whether such alignment identifies trainable projected subspaces~\citep{song2024does}. Loss-landscape studies reveal interpolation structure, mode connectivity, and useful weight averaging or ensembling along trajectories~\citep{goodfellow2014qualitatively,li2018visualizing,draxler2018essentially,garipov2018loss,izmailov2018averaging,huang2017snapshot,wortsman2022model}. Stage-wise analyses study edge-of-stability behavior, neural collapse, grokking, and lazy or kernel-like regimes~\citep{cohen2021gradient,papyan2020prevalence,power2022grokking,jacot2018neural,chizat2019lazy}.
These studies establish that optimization is structured, but they mainly concern spatial geometry, trainable subspaces, final representations, or named training regimes. Our object is the short-horizon temporal organization of parameter displacements, and where inside the model that structure resides.

A further line of work uses pretraining suites with released intermediate checkpoints, most prominently Pythia \citep{biderman2023pythia}, to study how capabilities and circuits form along training, from the early emergence of induction heads \citep{olsson2022context,tigges2024llm} to compute-optimal accounts of over-training \citep{hoffmann2022training}. We use the Pythia checkpoints as our one-pass regime but measure something else: the temporal redundancy of the weight trajectory itself, not behaviors or circuits. When our readings line up in time with events reported in this literature, we note the correspondence and make no mechanistic claim.

Architecture and optimizer choices are also known to reshape optimization. Normalization, attention, and adaptive methods change Transformer optimization behavior~\citep{zhang2020adaptive,xiong2020layer,zhang2024transformers}; momentum, weight decay, sharpness-aware perturbations, slow--fast weight coupling, and matrix-structured updates alter the geometry of update sequences~\citep{kingma2014adam,loshchilov2017decoupled,foret2020sharpness,zhang2019lookahead,ma2018quasi,jordan2024muon}. Weight decay in particular has well-studied mechanisms, through its interaction with normalization and effective learning rates~\citep{vanlaarhoven2017l2,d2024we}. Existing efficient-training methods reduce cost through pruning, sparse training, data selection, sample selection, or gradient and communication compression~\citep{frankle2018lottery,evci2020rigging,mirzasoleiman2020coresets,killamsetty2021grad,jiang2019accelerating,lin2017deep,vogels2019powersgd}. We take a complementary view: the optimizer trajectory itself may contain localized temporal redundancy. We therefore study short-horizon predictability as a trajectory-level measurement problem, rather than as an optimizer leaderboard or the performance of a single predictor, and we carry the same instruments across two training regimes --- multi-pass vision training and one-pass language-model pretraining.

\section{Measurement Framework}
\label{sec:methods}

We study short-horizon predictability as a property of a training trajectory rather than the performance of a single forecasting algorithm. Let $w_t\in\R^d$ denote the trainable parameters at checkpoint $t$, $\dw_t=w_{t+1}-w_t$ the one-step update, and $\dw_a^{(\tau)}=w_{a+\tau}-w_a$ the displacement realized over horizon $\tau$ from anchor $a$. The preceding $\npast$ updates form the local history $H_a=[\dw_{a-\npast},\ldots,\dw_{a-1}]$. Probes are evaluated on a measurement unit $S$, either a whole trainable tensor or one of $K$ disjoint within-tensor groups; $\dw_{t,S}$, $\dw_{a,S}^{(\tau)}$, and $H_{a,S}$ denote the corresponding restrictions. Every probe uses only history available before $a$ and is evaluated retrospectively against the displacement after $a$. Architectural roles are assigned only after measurement, for interpreting where predictable structure occurs. The choices of $\tau$, $\npast$, $K$, and grouping convention are specified with the experiments.

We use three complementary probe families of increasing representational complexity. \emph{Displacement-direction probes} test whether the latest update persists into the near-future displacement. \emph{Subspace-residual probes} ask whether that displacement remains within the low-dimensional span of recent updates, even when no single direction persists. \emph{Predictor-based probes} test whether an explicit local model can exploit richer temporal structure. The families are therefore related but non-equivalent measurements of short-horizon organization.

\paragraph{Displacement-direction probes.}
For element $j$ of unit $S$, let
$e_{a,S,j}^{(\tau)}=\one\{\operatorname{sign}(\dw_{a-1,S,j})=\operatorname{sign}(\dw_{a,S,j}^{(\tau)})\}$.
\emph{Element-wise sign agreement} (ESA) and \emph{vector cosine similarity} (VCS) are
\begin{equation}
\begin{aligned}
\mathrm{ESA}_S(a,\tau)
&=\frac{1}{|S|}\sum_{j\in S}e_{a,S,j}^{(\tau)},\\
\mathrm{VCS}_S(a,\tau)
&=\frac{\langle\dw_{a-1,S},\dw_{a,S}^{(\tau)}\rangle}
{\|\dw_{a-1,S}\|_2\,\|\dw_{a,S}^{(\tau)}\|_2}.
\end{aligned}
\end{equation}
ESA weights coordinates equally, whereas VCS weights them by displacement magnitude.

\paragraph{Subspace-residual probes.}
Let $P_{a,S}$ project onto the retained span of the recent-update matrix $H_{a,S}$. The \emph{future residual ratio} (FRR) is
\begin{equation}
\mathrm{FRR}_S(a,\tau)
=\frac{\|(I-P_{a,S})\dw_{a,S}^{(\tau)}\|_2}
{\|\dw_{a,S}^{(\tau)}\|_2},
\label{eq:frr}
\end{equation}
where lower values indicate stronger containment in the recent-history span. An in-window control analogously evaluates the latest update against the span of the preceding $\npast-1$ updates. For groups $\{G_g\}_{g=1}^{K}$, the \emph{low-residual group fraction} (LRGF) summarizes localization,
\begin{equation}
\mathrm{LRGF}(a,\tau;\theta)
=\frac{1}{K}\sum_{g=1}^{K}
\one\{\mathrm{FRR}_{G_g}(a,\tau)<\theta\}.
\label{eq:lrgf}
\end{equation}
Because the fixed threshold $\theta$ is convention-dependent, $\mathrm{LRGF}_{\rm cal}$ instead uses the $\alpha=0.05$ quantile of a size-matched permutation-null FRR distribution; under the null, its expected fraction is approximately $\alpha$, and we report excess over $\alpha$ where indicated.

\paragraph{Predictor-based probes.}
A predictor maps the recent history to a predicted displacement $\widehat f_{\mathrm{P},a,S}^{(\tau)}$. We compare three increasingly expressive local models: inertia, $\widehat f_{\mathrm{in},a,S}^{(\tau)}=\tau\dw_{a-1,S}$; a per-element linear trend fitted to recent updates and integrated over the next $\tau$ steps; and DMD, which fits and rolls forward a local linear map between consecutive update-history matrices. Because predicted magnitudes depend strongly on horizon and optimizer state, we emphasize direction and selection. The selector-free \emph{predicted-direction cosine} is
\begin{equation}
\cos_{\mathrm{P},S}(a,\tau)
=\frac{\langle\widehat f_{\mathrm{P},a,S}^{(\tau)},\dw_{a,S}^{(\tau)}\rangle}
{\|\widehat f_{\mathrm{P},a,S}^{(\tau)}\|_2\,\|\dw_{a,S}^{(\tau)}\|_2},
\quad \mathrm{P}\in\{\mathrm{in},\mathrm{lin},\mathrm{DMD}\}.
\label{eq:cospred}
\end{equation}
Since cosine ignores scale, $\cos_{\mathrm{in}}\equiv\mathrm{VCS}$.

Predictor outputs also define a selector
$m_{a,S,j}=\one\{\operatorname{sign}(\widehat f_{a,S,j}^{(\tau)})=\operatorname{sign}(\dw_{a-1,S,j})\}$.
The \emph{selected element-wise sign agreement} (SESA) and its \emph{selectivity lift} over the unselected baseline are
\begin{align}
\mathrm{SESA}_S(m;a,\tau)
&=\frac{\sum_{j\in S}m_{a,S,j}e_{a,S,j}^{(\tau)}}
{\sum_{j\in S}m_{a,S,j}},
\label{eq:sesa}\\
\mathrm{Lift}_S(m;a,\tau)
&=\mathrm{SESA}_S(m;a,\tau)-\mathrm{ESA}_S(a,\tau).
\label{eq:selectivity_lift}
\end{align}
SESA is defined when the selected subset is nonempty; positive lift means that the predictor isolates coordinates whose latest observed direction is more reliable than the unselective baseline. Under our construction, DMD predictions remain in the recent-history span, linking their attainable directional alignment to FRR. Full predictor constructions, rank selection, null generation, grouping rules, and numerical conventions are deferred to the appendix.

\section{Empirical Analysis}
\label{sec:empirical}

\paragraph{Experimental setup.}
We analyze stored weight trajectories in two complementary training
regimes.
\emph{Multi-pass vision training.}
Our CIFAR-10 panel contains $30$ model--optimizer cells, crossing five
architectures (a CIFAR-style ResNet, two ViT scales, AlexNet, and an
MLP) with six training configurations (SGD, SGD with momentum, SGD
with weight decay, SAM, Adam, and AdaGrad).
Models are trained for $50$ epochs with batch size $128$ and a cosine
learning-rate schedule, with weights recorded at every epoch boundary.
Unless otherwise stated, probes are evaluated at anchor epochs
$\{5,10,20,30,40\}$.
Section~\ref{sec:recipe} additionally uses Muon and controlled
weight-decay and momentum dose arms.

\emph{One-pass language-model pretraining.}
We analyze the public checkpoint trajectories of Pythia
70M, 160M, and 410M~\citep{biderman2023pythia}.
For each model, checkpoints are spaced by $1{,}000$ optimization steps;
we evaluate $123$ admissible post-warmup anchors spanning steps
$7{,}000$--$129{,}000$.

Unless otherwise stated, both regimes use horizon $\tau=5$ and history
length $\npast=5$.
The measurement clock is regime-specific: one probe step corresponds
to one epoch for CIFAR and one released checkpoint for Pythia.
Vision tensors use flat $K=16$ within-tensor groups by default, whereas
Pythia tensors use architecture-aligned groups within the $qkv$,
attention-output, MLP, and vector segments.
All probes are causal at the anchor and depend only on the preceding
weight history.
FRR- and LRGF-based results always state their grouping convention;
fixed-threshold $\mathrm{LRGF}_{\theta}$ is used for matched
cross-layer comparisons, while shuffle-calibrated
$\mathrm{LRGF}_{\mathrm{cal}}$ is used along the within-layer time
axis.
Complete configurations, checkpoint accounting, and measurement details are provided in \appref{app:experimental_details} of the supplementary material.

\subsection{Probe families are coherent but non-equivalent}
\label{sec:probe_relations}

\subsubsection{Within-family coherence under explicit conventions}
\label{sec:agreeement}

\begin{figure*}[t]
  \centering
  \includegraphics[width=0.95\textwidth]
  {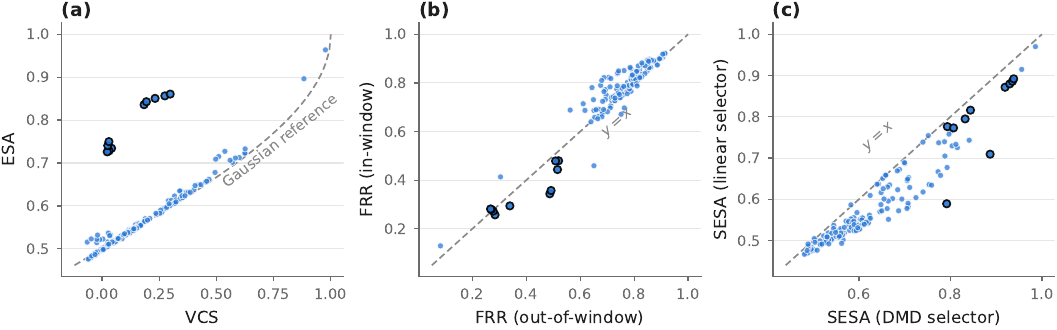}
  \caption{
  Within-family agreement on the controlled vision panel.
  (a) ESA versus group-mean VCS;
  (b) out-of-window versus in-window FRR;
  (c) SESA under DMD and linear-trend selectors.
  Circled points mark cells with near-zero-displacement coordinates.
  }
  \label{fig:family_panels}
\end{figure*}

We first examine within-family agreement to test whether alternative
readouts yield coherent measurements rather than idiosyncratic behavior
of a single readout. Figure~\ref{fig:family_panels} summarizes one
comparison for each probe family over the controlled vision panel of
$30$ model--optimizer cells and five anchors.

\emph{Displacement-direction.}
ESA and group-mean VCS weight the same displacement pair differently:
ESA gives each coordinate one sign vote, whereas VCS weights coordinates
by displacement magnitude. Under a centered bivariate Gaussian
reference, the corresponding sign agreement and correlation satisfy
$\mathrm{ESA}=1-\arccos(\mathrm{VCS})/\pi$
\cite{schmid2007nonparametric}. As shown in
Fig.~\ref{fig:family_panels}(a), 28 of the 30 cells closely follow this
parameter-free curve across anchors. Departures occur primarily in cells
with many near-zero-displacement coordinates (coordinates whose
displacement is zero or numerically negligible over one or both
intervals): these coordinates contribute little to VCS but remain
sensitive to ESA's coordinate-wise sign convention, and restricting the
comparison to coordinates that move in both intervals largely restores
agreement.

\emph{Subspace-residual.}
Out-of-window and in-window FRR are strongly correlated
(median per-cell Spearman $\rho_S=0.83$;
Fig.~\ref{fig:family_panels}(b)), consistent with local smoothness across
nearby target windows. FRR is computed within contiguous,
structure-aligned parameter groups using a model-specific, empirically
selected granularity, which determines the spatial scale at which local
subspace structure and predictable pockets can be resolved.

\emph{Predictor-based.}
DMD and linear-trend selectors induce strongly consistent SESA
orderings (median per-cell $\rho_S=0.89$), while DMD achieves
systematically higher values (Fig.~\ref{fig:family_panels}(c)),
consistent with its ability to exploit coupled low-rank dynamics in the
recent trajectory rather than extrapolating a single linear trend.
Alternative grouping granularities and additional within-family checks
are reported in \appref{app:probe_conventions}.

\subsubsection{Cross-family relations across levels of complexity}
\label{sec:cross_family}

We next compare representative readouts across probe families at the
tensor level. Correlations are computed across tensors within each
model--optimizer cell and summarized by the median over the 30 cells,
avoiding correlations induced by pooling different training recipes.

\begin{figure*}[t]
  \centering
  \includegraphics[width=0.95\textwidth]
  {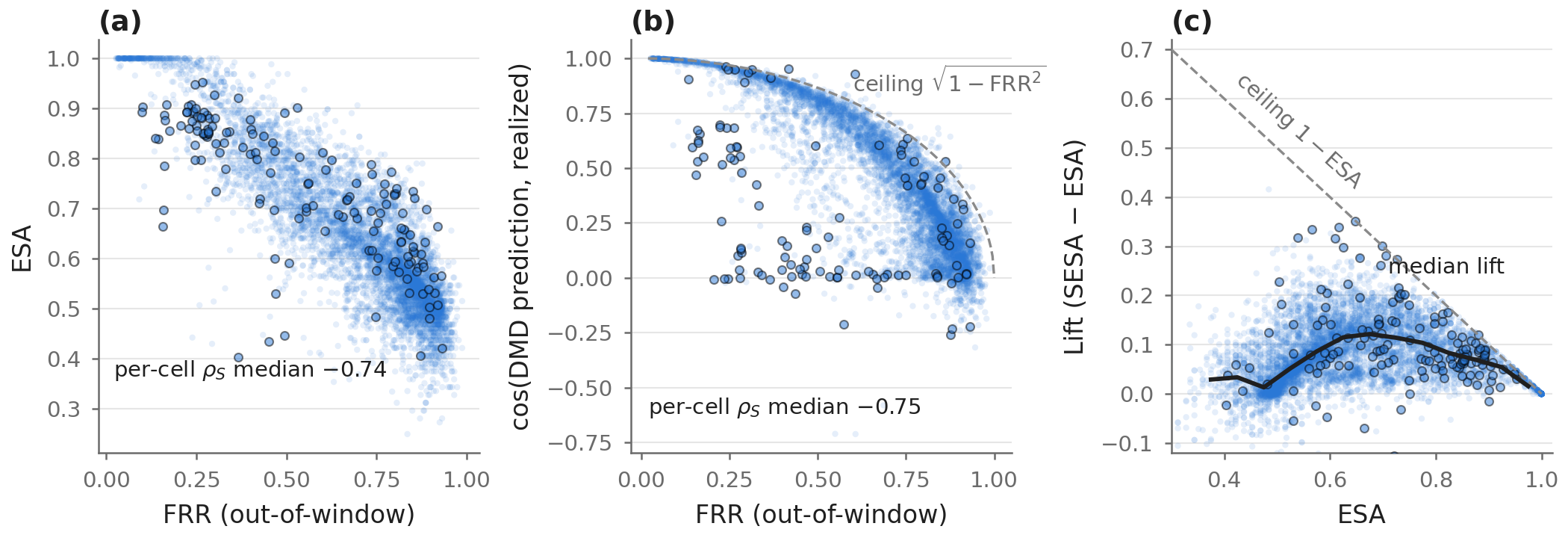}
  \caption{
  Cross-family relations on the controlled vision panel.
  (a) ESA versus out-of-window FRR;
  (b) out-of-window FRR versus DMD direction cosine;
  (c) ESA versus DMD lift.
  Circled points mark cells flagged by the near-zero-displacement
  diagnostic.
  }
  \label{fig:cross_family}
\end{figure*}

\emph{Direction--subspace.}
ESA and out-of-window FRR are strongly anticorrelated
(median per-cell Spearman $\rho_S=-0.74$;
Fig.~\ref{fig:cross_family}(a)): tensors with more persistent
coordinate-wise directions also tend to keep their future displacement
closer to the span of recent updates. The two probes therefore detect
closely related structure from different geometric perspectives.

\emph{Subspace--predictor.}
Under our construction, the DMD prediction lies in the recent-history
subspace, giving the geometric bound
\[
\cos_{\mathrm{DMD}}
\leq \sqrt{1-\mathrm{FRR}^{2}}.
\]
DMD direction cosine is correspondingly strongly anticorrelated with
FRR (median per-cell Spearman $\rho_S=-0.75$;
Fig.~\ref{fig:cross_family}(b)), with 37\% of tensor-level observations
lying within 0.15 of the bound. FRR thus characterizes the directional
headroom available to a predictor constructed from the recent-history
subspace.

\emph{Direction--predictor.}
DMD lift satisfies the headroom bound
\[
\mathrm{Lift}
=\mathrm{SESA}-\mathrm{ESA}
\leq 1-\mathrm{ESA}.
\]
Its binned median peaks at $0.118$ for
$\mathrm{ESA}\in[0.6,0.7]$, rather than at the highest ESA values
(Fig.~\ref{fig:cross_family}(c)). High ESA already indicates strong
predictability from directional persistence alone, leaving little
residual error for DMD-based selection to remove; low lift in this
region therefore does not imply low predictability. At very low ESA,
the trajectory also offers less stable structure to exploit. The
largest lift occurs at intermediate ESA, where DMD can recover
additional low-rank temporal structure beyond coordinate-wise
persistence while sufficient headroom remains.

Together, these results show that predictability can be characterized
at complementary levels of complexity: coordinate-wise directional
persistence, local history-subspace structure, and additional temporal
structure recoverable by an explicit predictor.

\subsection{Predictability is spatially localized across parameter scales}
\label{sec:heterogeneity}

We next ask where the measured short-horizon structure resides.
We partition trainable tensors into two predefined morphological
classes: \emph{auxiliary} tensors include vector-like parameters such
as normalization parameters and biases, whereas \emph{bulk} tensors
include matrix-like feature-transforming weights such as attention
projections, MLP weights, and convolutional kernels.
Exact tensor membership and partition audits are provided in
\appref{app:spatial_localization}.

\begin{figure*}[t]
  \centering
  \includegraphics[width=0.6\textwidth]
  {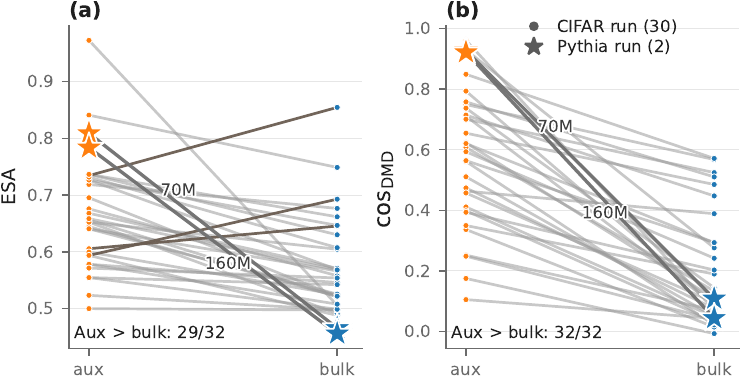}
  \caption{
  Auxiliary--bulk separation across training regimes.
  Each line connects the auxiliary and bulk summaries within a
  matched CIFAR run or Pythia trajectory; circles denote CIFAR runs
  and stars denote Pythia trajectories.
  Auxiliary tensors exceed bulk tensors in $29/32$ comparisons under
  ESA and $32/32$ under the selector-free DMD direction cosine.
  Because the two regimes use different grouping protocols, we
  compare the within-line ordering rather than absolute levels across
  regimes.
  }
  \label{fig:aux_bulk_slope}
\end{figure*}

Figure~\ref{fig:aux_bulk_slope} shows a stable tensor-scale
separation within matched trajectories. Auxiliary tensors exceed bulk
tensors in $29/32$ trajectory-level comparisons under ESA and in
$32/32$ comparisons under the DMD direction cosine. Before temporal
aggregation, the same ordering holds at all $246$ Pythia
model--checkpoint snapshots under both probes. The recurring paired
ordering shows that this coarse spatial organization is shared across
multi-pass vision training and one-pass language-model pretraining,
rather than arising from a single probe or training regime. We do not
compare the magnitudes of the CIFAR and Pythia gaps because their
grouping and checkpoint protocols differ.

Despite their lower aggregate predictability, bulk tensors contain a
small minority of within-tensor groups with substantially elevated
predictability. We refer to these localized groups as
\emph{predictable pockets}. Under the reported structural cuts and
reference convention, they account for approximately $6\%$ of pooled
group--anchor observations. Most pockets are transient: their
identities change across training anchors, although the degree of
persistence varies by parameter role. Detailed pocket trajectories and
persistence analyses are reported in
\appref{app:pocket_persistence}.

\begin{figure}[t]
  \centering
  \includegraphics[width=0.9\columnwidth]
  {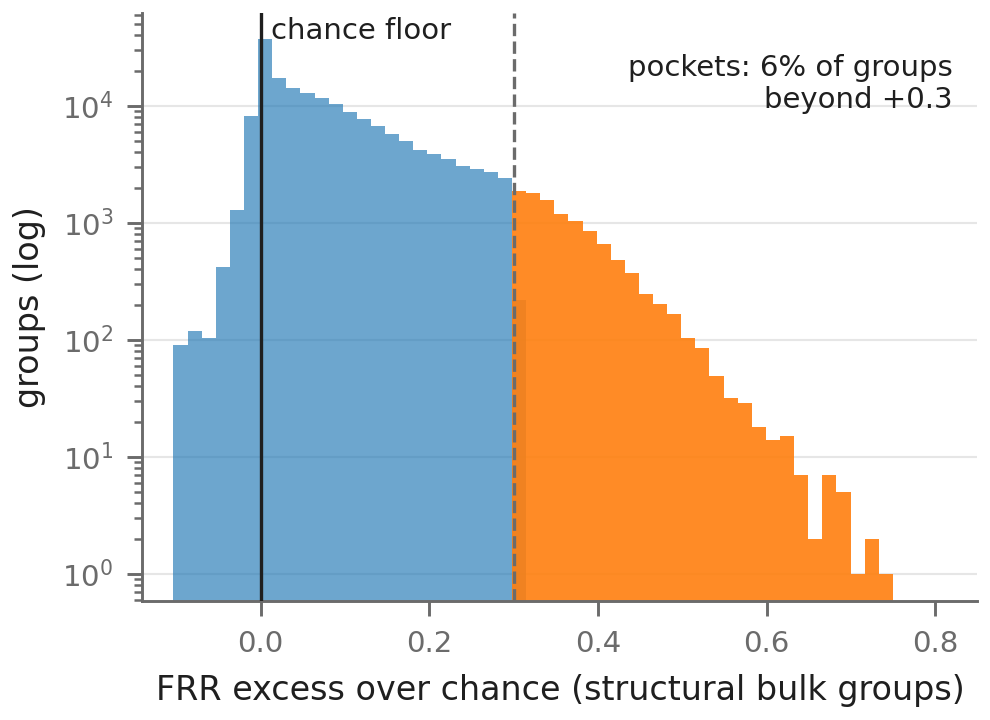}
  \caption{
  Distribution of FRR excess above the approximate isotropic reference
  $\sqrt{(D-r)/D}-\mathrm{FRR}$ over structurally grouped bulk
  parameters, with $r=5$. The highlighted tail
  ($\mathrm{excess}>0.3$) contains approximately $6\%$ of pooled
  group--anchor observations.
  }
  \label{fig:pocket_scarcity}
\end{figure}

\subsection{Probe readings track the realized optimization state}
\label{sec:dynamics}

The proposed probes need not be interpreted in isolation. We compare
their readings with complementary diagnostics derived from realized
trajectory motion and training telemetry. For ESA, we use movement
directedness---the ratio of net displacement to path length over the
future window---and a loss-side estimate of gradient coherence based
on loss decrease, learning rates, and gradient norms. Across
heterogeneous CIFAR training conditions, ESA correlates with these
quantities at $\rho=0.978$ and $\rho=0.804$, respectively. Formal
definitions, their relation to prior gradient-coherence and
signal-to-noise measures, and the complete comparisons are provided in
\appref{app:state_correspondence}.

\begin{figure}[t!]
  \centering
  \includegraphics[width=\columnwidth]
  {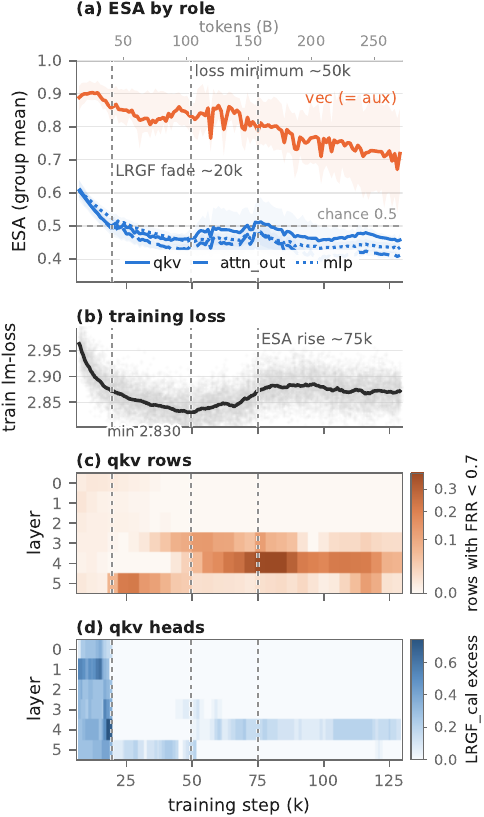}
  \caption{
  Pythia-70M case study.
  (a) ESA by parameter role.
  (b) Released training loss and its smoothed trajectory.
  (c) Fraction of $qkv$ row groups satisfying $\mathrm{FRR}<0.7$.
  (d) Head-aligned $qkv$ $\mathrm{LRGF}_{\mathrm{cal}}$ excess.
  Vertical lines mark the early fading of head-aligned excess, the
  loss minimum, and the later $qkv$ ESA rise.
  }
  \label{fig:pythia70m_case}
\end{figure}

We next use Pythia-70M as a case study to examine how these
state-level correspondences unfold within a one-pass pretraining
trajectory. Figure~\ref{fig:pythia70m_case} jointly resolves the
public loss trajectory and probe readings across parameter roles,
layers, and grouping scales.

The auxiliary vector segment remains clearly separated from the bulk
roles throughout training. Although the bulk tensors contain most of
the model parameters, all three bulk roles undergo a rapid initial
decline in ESA and cross below the $0.5$ sign-agreement reference.
They continue to evolve within this sub-$0.5$ regime rather than
settling to a common floor. The clearest mid-to-late turn occurs in
$qkv$: its ESA bends upward near the smoothed loss minimum around
$50$k steps, reaches a local maximum near $75$k after the loss has
begun to rise, and subsequently declines again. Attention-output and
MLP exhibit smaller, non-identical variations over the same interval.

The finer-resolution views expose an earlier reorganization within
$qkv$ at a different spatial scale. Head-aligned
$\mathrm{LRGF}_{\mathrm{cal}}$ excess is concentrated near the
beginning of training and largely fades around $20$k steps. Row-level
low-FRR pockets, however, are initially sparse and emerge only after
some training, first in deeper layers and later across deep and
intermediate layers. The fading of head-aligned excess therefore
should not be read as a global disappearance of predictable
structure. Rather, the detectable $qkv$ structure changes its
localization and its relation to the chosen grouping scale.

The temporal proximity between the loss reversal and the later
$qkv$ ESA excursion is a retrospective correspondence, not evidence
that either event causes or uniquely explains the other. The
role-, depth-, and grouping-resolved readings nevertheless reveal
trajectory reorganizations that are not localized by the scalar loss
curve alone. Complete results for the other Pythia scales, together
with cross-group diagnostics and pocket-evolution analyses, are
reported in \appref{app:pythia_scale_extensions}.

\subsection{Training conditions modulate state-dependent temporal redundancy}
\label{sec:recipe}

Across our controlled CIFAR experiments, temporal redundancy varies
systematically with architecture, optimizer family, weight decay, and
momentum. Size-controlled architecture comparisons and controlled
weight-decay and momentum dose studies are reported in
\appref{app:recipe_conditioning}; here we use optimizer-family
variation as a compact main-text example.

\begin{figure}[t]
  \centering
  \includegraphics[width=\columnwidth]
  {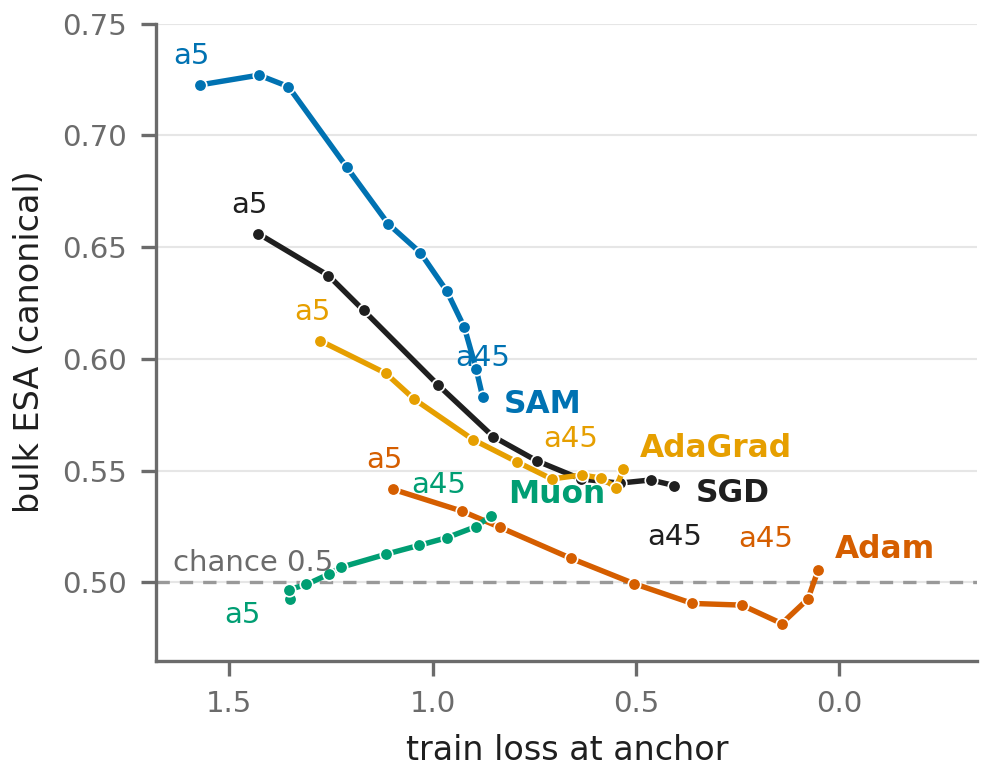}
  \caption{
  Optimizer-family profiles for ViT-mid along training loss
(horizontal axis reversed), with canonical bulk ESA on the vertical
axis. All-family comparisons use the common support
$\mathrm{loss}\approx0.9$--$1.4$; matched loss does not imply matched
optimizer state. Dashed line: $0.5$ sign-agreement reference.
  }
  \label{fig:opt_axis}
\end{figure}

Because optimizer families traverse the loss trajectory at different
rates, Figure~\ref{fig:opt_axis} compares them along training loss
rather than checkpoint index. Distinct profiles remain visible over the
shared range: SAM occupies the highest-ESA regime, Muon remains near the
$0.5$ reference, Adam lies below SGD over most of their overlap, and SGD
and AdaGrad approach a similar low-loss plateau. Thus, optimization
position organizes the evolution of predictability, while the training
recipe also conditions its level and local directional structure.

Higher predictability should not be interpreted as better optimization:
optimizers with strong final performance occupy both high- and
low-predictability regimes. The complete optimizer panel,
predictability--accuracy comparison, and the architecture,
weight-decay, and momentum results are provided in
\appref{app:recipe_conditioning}.

\section{Discussion, Limitations, and Future Directions}

The proposed measurements provide a parameter-resolved and time-local view of the temporal organization of training dynamics. They localize short-horizon structure across parameter roles, layers, tensors, and within-tensor groups, while tracking how that structure emerges, disappears, or changes its spatial organization. Directional, subspace, and predictor-based probes provide complementary views: their agreement supports shared trajectory structure, whereas their differences distinguish non-equivalent forms of temporal organization. The framework can therefore support retrospective analysis of checkpoint-rich runs with sufficiently fine temporal resolution.

The probes nevertheless capture a restricted class of dynamics, emphasizing low-order directional relations and structure representable within short, locally low-dimensional history spaces rather than arbitrary nonlinear, long-memory, or high-rank dependence. Their readings also depend on checkpoint spacing, horizon, history length, grouping convention, and null calibration, which define the scale of observation. Because ESA and FRR use realized post-anchor displacement, they are retrospective or delayed measurements rather than immediately available online signals, and temporal alignment with loss or other events should not be interpreted as mechanistic evidence. Our current coverage of controlled vision training and public Pythia trajectories also does not establish generality across larger models or other learning paradigms.

Future work should extend these measurements across broader scales and regimes, and study the emergence, migration, and disappearance of predictable pockets together with their alignment to representation formation, circuit development, and interpretable parameter communities. The sub-$0.5$ ESA regimes further motivate examining possible connections to edge-of-stability behavior \cite{cohen2021gradient} and Ornstein--Uhlenbeck-like mean reversion\cite{mandt2017stochastic}. Another direction is to develop delayed-online or history-only diagnostics that complement scalar loss. Our preliminary online example suggests that restricting predictive write-back to a higher-confidence auxiliary subset can be more reliable than broad prediction (\appref{app:online}), motivating future use of these instruments for adaptive prediction, selective updating, skipping, and compression.

\bibliography{citation}

\clearpage
\setcounter{page}{1}
\setcounter{figure}{0}
\setcounter{table}{0}
\setcounter{equation}{0}
\graphicspath{{supp_figures/}}

\def\addcontentsline#1#2#3{\addtocontents{#1}{\protect\contentsline{#2}{#3}{\thepage}{}}}

\renewcommand{\topfraction}{0.85}
\renewcommand{\dbltopfraction}{0.9}
\renewcommand{\textfraction}{0.07}
\renewcommand{\floatpagefraction}{0.75}
\renewcommand{\dblfloatpagefraction}{0.75}
\setcounter{totalnumber}{4}
\setcounter{dbltopnumber}{3}

\twocolumn[
\begin{center}
{\LARGE\bfseries Supplementary Material for\\[0.25em]
\emph{Measuring Structured Predictability in Neural Training Dynamics: A Cross-Regime Study}\par}
\vspace{0.7em}
{\Large\bfseries Fanqi Wang, Weisheng Tang, and Hairong Qi\textsuperscript{*}\par}
\vspace{0.25em}
{\normalsize Department of Electrical Engineering and Computer Science, University of Tennessee, Knoxville\par}
{\normalsize \textsuperscript{*}Corresponding author\par}
\vspace{0.8em}
\end{center}
]
\noindent
This document is the supplementary material for the submission
\emph{Measuring Structured Predictability in Neural Training Dynamics:
A Cross-Regime Study}. Notation, probe definitions, and terminology are
those of \mainref{sec:methods}; references prefixed by ``Section'',
``Figure'', or ``Eq.'' point into the main paper, and ``Appendix~A'' to
``Appendix~H'' point into this document.

\begingroup
\small
\setcounter{tocdepth}{2}
\tableofcontents
\endgroup

\appendix

\section{Experimental Coverage and Protocol Details}
\label{app:experimental_details}

This appendix records the training corpus behind
\mainref{sec:empirical}, the checkpoint accounting for both regimes, and
the measurement constants shared by every probe.
Appendix~\ref{app:cifar_configurations} covers the controlled vision
panel and Appendix~\ref{app:cifar_families} the further run families that
\mainref{sec:recipe} draws on. Appendix~\ref{app:pythia_coverage} covers
the pretraining checkpoints, and
Appendix~\ref{app:anchor_construction} the measurement clocks, anchor
construction, and seeds.

\subsection{Vision Panel: Architectures, Optimizers, and Recipes}
\label{app:cifar_configurations}

The controlled vision panel crosses five architectures with six training
configurations. The architectures span three structural families: an MLP;
AlexNet and a CIFAR-style ResNet-18 variant with its last stage removed;
and two ViT scales, written ViT and ViT-mid. The six configurations are
plain SGD, SGD with momentum, SGD with weight decay, Adam, AdaGrad, and
SAM. Table~\ref{tab:appA_recipes} gives the shared recipe, the optimizer
definitions, and the learning rate of every cell together with the
parameter count of each architecture.

\begin{table*}[tp]
\centering
\small

{\centering (a) Shared recipe\par}
\vspace{0.3em}

\begin{tabular}{l l}
\toprule
Item & Setting \\
\midrule
Dataset & CIFAR-10 (full 50k train / 10k test); CIFAR-100 only in the transfer family \\
Loss / batch size & cross-entropy; batch size 128 everywhere \\
Epochs / schedule & 50 epochs; schedules stepped once per epoch \\
Cosine schedule & \texttt{CosineAnnealingLR} to $0$ with $T_{\max}$ = epochs \\
Augmentation & \texttt{RandomCrop(32, padding=4)} $+$ \texttt{RandomHorizontalFlip} \\
AlexNet exception & \texttt{Resize(128)}, \texttt{Normalize(0.5, 0.5)}, no augmentation \\
Normalization & CIFAR-10: mean $(0.4914, 0.4822, 0.4465)$, std $(0.2023, 0.1994, 0.2010)$ \\
 & CIFAR-100: mean $(0.5071, 0.4865, 0.4409)$, std $(0.2673, 0.2564, 0.2762)$ \\
Statistics pinning & the transfer and calibration families pin CIFAR-10 statistics for
 \emph{all} their runs, \\
 & including the CIFAR-100 ones, so transfer pairs see identical input statistics \\
Seed & 42 for the panel; 43 and 44 for replicates \\
Snapshots & weights recorded at every epoch boundary \\
\bottomrule
\end{tabular}

\vspace{1.0em}
{\centering (b) Optimizer definitions\par}
\vspace{0.3em}

\begin{tabular}{l l l l l}
\toprule
Name & Rule & Momentum & Weight decay & Other \\
\midrule
SGD & SGD & $0$ & $0$ & --- \\
SGD-mom & SGD & $0.9$ & $0$ & --- \\
SGD-wd & SGD & $0$ & $5\times10^{-4}$ & --- \\
Adam & Adam & $\beta=(0.9, 0.999)$ & $0$ & --- \\
AdaGrad & AdaGrad & --- & $0$ & --- \\
SAM & SAM over plain SGD & $0$ & $0$ & $\rho=0.05$ \\
\bottomrule
\end{tabular}

\vspace{1.0em}
{\centering (c) Architectures and per-cell learning rates\par}
\vspace{0.3em}

\begin{tabular}{l r cccccc}
\toprule
Architecture & Params & SGD & SGD-mom & SGD-wd & Adam & AdaGrad & SAM \\
\midrule
MLP & 3{,}805{,}450 & $0.1$ & $0.01$ & $0.1$ & $10^{-3}$ & $10^{-2}$ & $0.05$ \\
AlexNet & 57{,}044{,}810 & $0.05$ & $0.01$ & $0.05$ & $10^{-3}$ & $5\times10^{-3}$ & $0.01$ \\
ResNet & 2{,}777{,}674 & $0.1$ & $0.1$ & $0.1$ & $10^{-3}$ & $10^{-2}$ & $0.1^{s}$ \\
ViT & 546{,}186 & $0.1$ & $0.01$ & $0.1$ & $10^{-3}{}^{s}$ & $10^{-2}$ & $0.05^{s\dagger}$ \\
ViT-mid & 2{,}855{,}562 & $0.05$ & $0.05$ & $0.05$ & $5\times10^{-4}$ & $5\times10^{-3}$ & $0.025^{s\dagger}$ \\
\bottomrule
\end{tabular}

\smallskip
{\footnotesize
All $30$ cells are trained from scratch under the shared recipe of panel~(a) with
seed $42$. Parameter counts are for the $10$-class heads; ViT-mid has
$2{,}875{,}812$ parameters with the $100$-class head used in the transfer family.
Most cells take the model-default learning rate; a few were hand-tuned, including
ViT with SAM ($0.05$) and ViT-mid with SAM ($0.025$).
$^{s}$~Cell additionally re-trained with seeds $43$ and $44$.
$^{\dagger}$~Those two SAM re-runs used the model-default learning rate
($0.1$ and $0.05$), twice the hand-tuned seed-$42$ value, so these two cells are
not learning-rate-matched across seeds.}

\caption{Vision panel: shared recipe, optimizer definitions, and per-cell
learning rates. \textbf{(a)} the recipe held fixed across every run in this
appendix; \textbf{(b)} the six training configurations of the panel;
\textbf{(c)} the five architectures with their parameter counts and the
learning rate of each cell.}
\label{tab:appA_recipes}
\end{table*}

All cells train from scratch for $50$ epochs at batch size $128$ under a
cosine schedule stepped once per epoch, with seed $42$, and weights are
recorded at every epoch boundary. That epoch-level snapshot cadence is
what sets the probe time step on this side. Most cells take the
model-default learning rate; the hand-tuned exceptions are listed in the
table. AlexNet is the one architecture with a different input pipeline,
since it resizes to $128$ pixels and trains without augmentation.

\subsection{Additional Run Families}
\label{app:cifar_families}

Around the panel, six further families vary one ingredient at a time
while holding the shared recipe fixed
(Table~\ref{tab:appA_families}). Two of them supply the controlled
comparisons of \mainref{sec:recipe}: a Muon arm over the same five
architectures, and a weight-decay and momentum dose ladder on plain SGD.
The remaining families support the robustness and calibration checks:
seed replicates on five cells, a transfer-dynamics family between
CIFAR-10 and CIFAR-100, and a probe-calibration batch covering a schedule
zoo, basin slices, a learning-rate dose ladder, and a blind audit whose
configurations were withheld from the analysis side. The corpus is $139$
training runs and one frozen source checkpoint.

\begin{table*}[tp]
\centering
\small

\begin{tabular}{l r p{0.60\textwidth}}
\toprule
Family & Runs & Recipe deltas with respect to the panel \\
\midrule
Panel & 30 &
The five architectures of Table~\ref{tab:appA_recipes}(c) crossed with the six
configurations of Table~\ref{tab:appA_recipes}(b); from scratch, seed $42$. \\
\addlinespace
Muon arm & 5 &
The same five architectures under Muon (momentum $0.95$,
\texttt{ns\_steps}${=}5$), learning rates $0.02$ (MLP), $0.01$ (AlexNet),
$0.02$ (ResNet), $0.005$ (ViT), $0.01$ (ViT-mid). One additional stalled
diagnostic run at the ViT model-default learning rate $0.02$ is kept on disk
and excluded from readouts. \\
\addlinespace
Weight-decay and momentum dose & 26 &
Plain SGD at the model-default learning rate, cosine, $50$ epochs, seed $42$.
Weight-decay ladder $\{10^{-4}, 10^{-3}, 2\times10^{-3}\}$ over the five
architectures, with the override applied to plain SGD only; momentum ladder
$\{0.5, 0.95\}$ over the five architectures; one further MLP run at momentum
$0.7$. Two runs collapsed and are kept on disk but excluded from every
readout: MLP at momentum $0.95$ (final test accuracy $0.170$) and AlexNet at
momentum $0.95$ (final test accuracy $0.100$, chance level). \\
\addlinespace
Seed replicates & 10 &
Five panel cells re-trained at seeds $43$ and $44$, recipe otherwise
identical: ResNet--SAM, ResNet--Muon, ViT--Adam, ViT--SAM, ViT-mid--SAM. The
two SAM re-runs are not learning-rate-matched (Table~\ref{tab:appA_recipes}
note $\dagger$). \\
\addlinespace
Transfer dynamics & 22 &
All ViT-mid under plain SGD (momentum $0$, weight decay $0$), CIFAR-10
statistics pinned, plus one frozen source checkpoint. CIFAR-10 to CIFAR-100
and CIFAR-100 to CIFAR-10 at constant learning rate, two arms ($0.005$ and
$0.025$), seeds $42/43/44$ with weight seed $k$ and data-order seed
$1000{+}k$; scratch controls on both datasets; a stationary continuation
($+50$ epochs at $0.025$) and step learning-rate switches at epoch $25$. \\
\addlinespace
Probe calibration & 45 &
Plain SGD, seed $42$, CIFAR-10 statistics pinned. A $16$-run schedule zoo
(ResNet and ViT-mid over linear, two step schedules, exponential,
warmup-cosine and cosine-with-floor, two ResNet constant-rate runs, a
$100$-epoch cosine run and a basin anneal); $10$ basin slices (early and late
basins crossed with five constant rates, $12$ epochs); a $5$-run
learning-rate dose ladder ($\times\{\tfrac{1}{25},\tfrac{1}{10},\tfrac{1}{2},
2,4\}$ of base $0.025$); two stationary continuations; and a $12$-run blind
audit whose configurations were withheld from the analysis side. \\
\midrule
Total & 139 & training runs, plus one frozen source checkpoint \\
\bottomrule
\end{tabular}

\caption{Run families of the vision corpus. The panel is the set analyzed
throughout \mainref{sec:probe_relations} to \mainref{sec:dynamics}; the Muon
arm and the dose ladders supply the controlled comparisons of
\mainref{sec:recipe}; the remaining families support the robustness and
calibration checks. The two collapsed momentum runs are kept on disk and
excluded from every readout.}
\label{tab:appA_families}
\end{table*}

Two runs in the momentum ladder collapsed and are excluded from every
readout while being kept on disk: MLP at momentum $0.95$, which ends at
$0.170$ test accuracy, and AlexNet at momentum $0.95$, which ends at
$0.100$, the chance level for ten classes. Neither contributes to the
momentum comparisons.

\subsection{Pretraining Checkpoint Coverage}
\label{app:pythia_coverage}

\begin{figure*}[tp]
  \centering
  \includegraphics[width=\textwidth]{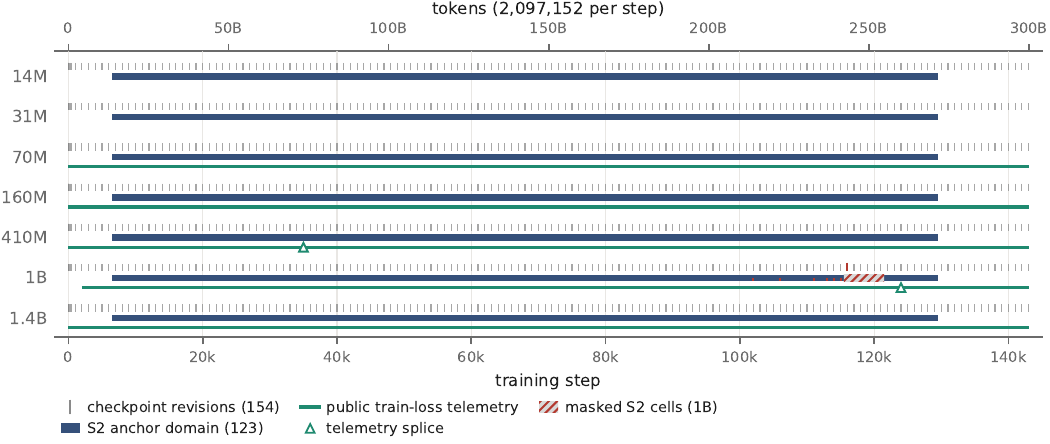}
  \caption{Checkpoint and anchor coverage across the pretraining scales.
  Each row is one model scale on a shared training-step axis, with tokens on
  the top axis at $2{,}097{,}152$ per step. Ticks give the $154$ public
  checkpoint revisions, released on the schedule $0,1,2,4,\ldots,512$ and
  every $1$k steps thereafter up to step $143$k. Bars give the anchor domain
  actually measured, $123$ anchors on a $1$k grid spanning $7$k to $129$k for
  every scale. The domain is trimmed at both ends for protocol reasons rather
  than for data availability: the head drops the warmup phase and leaves room
  for the backward history window, and the tail reserves a complete future
  window for the longest horizon ($143\mathrm{k}-14\mathrm{k}=129\mathrm{k}$
  at $\tau{=}14$). Lines give the span of public training-loss telemetry, and
  the two smallest scales carry no line because no public run exists for
  them. Triangles mark splices where the released telemetry is assembled from
  more than one run. Hatching on the 1B row marks the step-$116$k checkpoint
  defect and the anchors masked because of it, with short ticks at the
  horizon-specific future endpoints; the telemetry line and the official
  evaluations are unaffected by the masking.}
  \label{fig:appA_coverage_map}
\end{figure*}

Table~\ref{tab:appA_pythia} lists the released checkpoint series we
analyze, and Figure~\ref{fig:appA_coverage_map} places the measured
anchor domain on the training-step axis of each size. The main panel is
Pythia-70M, 160M, and 410M. The extended sizes 14M, 31M, 1B, and 1.4B are
processed under the identical protocol. Six of the seven sizes
enter the cross-scale readings of
Appendix~\ref{app:pythia_scale_extensions}; 1B is covered by the coverage
accounting here but is excluded from every cross-scale reading, on the three
grounds given there.

\begin{table*}[tp]
\centering
\small
\setlength{\tabcolsep}{4pt}

{\centering (a) Sizes, architecture, and coverage on disk\par}
\vspace{0.3em}

\begin{tabular}{l rr c l}
\toprule
 & \multicolumn{2}{c}{Parameters} & & \\
\cmidrule(lr){2-3}
Model & total & non-embedding & $L \times d_{\mathrm{model}}$ & Coverage \\
\midrule
\multicolumn{5}{@{}l}{\emph{Main panel}} \\
Pythia-70M & 70{,}426{,}624 & 18{,}915{,}328 & $6\times512$ & 154 revisions; 123 anchors \\
Pythia-160M & 162{,}322{,}944 & 85{,}056{,}000 & $12\times768$ & 154 revisions; 123 anchors \\
Pythia-410M & 405{,}334{,}016 & 302{,}311{,}424 & $24\times1024$ & 123 anchors (246 shards); spectrum 154 rev. \\
\midrule
\multicolumn{5}{@{}l}{\emph{Extended set, reported in Appendix~\ref{app:pythia_scale_extensions}}} \\
Pythia-14M & 14{,}067{,}712 & 1{,}189{,}888 & $6\times128$ & 123 anchors (246 shards); spectrum 154 rev. \\
Pythia-31M & 30{,}494{,}720 & 4{,}739{,}072 & $6\times256$ & 123 anchors (246 shards); spectrum 154 rev. \\
Pythia-1B$^{\dagger}$ & 1{,}011{,}781{,}632 & 805{,}736{,}448 & $16\times2048$ & 123 anchors (246 shards); spectrum 154 rev. \\
Pythia-1.4B & 1{,}414{,}647{,}808 & 1{,}208{,}602{,}624 & $24\times2048$ & 123 anchors (246 shards); spectrum 154 rev. \\
\bottomrule
\end{tabular}

\smallskip
{\footnotesize
``Total'' counts pure parameters of the released fp16 weights, with no buffers;
``non-embedding'' excludes the untied input and output embeddings (vocabulary
$50{,}304$). The 70M and 160M values are read from the verified reference card;
the five remaining sizes are recomputed from the on-disk group sizes, by a method
that reproduces the reference card exactly on 70M. The extended sizes were
processed at a second compute site under the identical engine and protocol, with
per-size equivalence gates passing at zero maximum relative error ($1260$ units at
14M and 31M, $4824$ at 410M and 1.4B, $3240$ at 1B); the gate tables carry
non-empty exemption counts for degenerate units, and every non-exempt row passes.
Anchors are steps $7000$ to $129000$ at spacing $1000$, and each anchor produces
one panel and one segment shard.}

\vspace{1.0em}
{\centering (b) Public telemetry and evaluation provenance\par}
\vspace{0.3em}

\begin{tabular}{l p{2.7in} l l}
\toprule
Model & Public train-loss telemetry & Splice check & Official zero-shot evals \\
\midrule
Pythia-14M & none public & --- & none public \\
Pythia-31M & none public & --- & none public \\
Pythia-70M & \texttt{32t0zbcs} (to 143k) & single run & 27 JSON records \\
Pythia-160M & \texttt{3mvtbwii} (to 143k) & single run & 27 JSON records \\
Pythia-410M & \texttt{3jg1upg7} $[1,35\mathrm{k})$ $+$ \texttt{12j05401} $[35\mathrm{k},143\mathrm{k}]$ & overlap $|\Delta\mathrm{loss}|\le0.019$ & 28 JSON records \\
Pythia-1B$^{\dagger}$ & \texttt{339s0ka6} $+$ \texttt{1ena81fo} $+$ \texttt{318z89xb} (restart chain, bf16) & 49-step overlaps, $\Delta\mathrm{loss}\le4\times10^{-16}$ & 28 JSON records \\
Pythia-1.4B & \texttt{qiu9n7a6} (fp16) & single run & 28 JSON records \\
\bottomrule
\end{tabular}

\smallskip
{\footnotesize
All telemetry comes from the public training project of the checkpoint release, in
the run group matching the released weights, with configurations cross-checked item
by item against the published training configurations. The 1B chain starts at step
$2001$; steps $0$ to $2000$ are missing and the analysis domain starts at $7000$, so
the gap is outside it. Official zero-shot evaluations come from the release
repository and were produced by an older harness, so only within-harness trends are
read from them.}

\smallskip
{\footnotesize
$^{\dagger}$\,\textbf{Pythia-1B checkpoint defect.} Revision \texttt{step116000} of
the 1B series has wrong content upstream: it is the only 1B revision exported by a
different release pipeline, and tensor forensics place it as an inserted state from
a different run, with $w_{115\mathrm{k}}\approx w_{117\mathrm{k}}$. The defect dates
from the first weight upload and is not repairable downstream. Affected cells are
masked at read time, by probe family: base-anchor rows at $a{=}116$k for all $\tau$;
history-window anchors $a{=}116$k to $121$k for all $\tau$; the future-endpoint
cells $(a,\tau)\in\{(115\mathrm{k},1),(114\mathrm{k},2),(113\mathrm{k},3),
(111\mathrm{k},5),(106\mathrm{k},10),(102\mathrm{k},14)\}$; and the
\texttt{step116000} row of the embedding-spectrum series. Public telemetry and the
official evaluations are unaffected, since neither is derived from local
checkpoints.}

\caption{Pretraining checkpoint coverage. \textbf{(a)} parameter counts,
architecture, and on-disk coverage, split into the main panel and the
extended set reported in Appendix~\ref{app:pythia_scale_extensions};
\textbf{(b)} public telemetry and evaluation provenance, including the
splice checks for the two series whose released telemetry is assembled from
more than one run. The 1B rows carry the \texttt{step116000} checkpoint
defect and its masking windows.}
\label{tab:appA_pythia}
\end{table*}

Checkpoints are released every $1{,}000$ steps after a dense early grid,
giving $154$ revisions up to step $143{,}000$. We evaluate $123$ anchors
on a $1{,}000$-step grid spanning steps $7{,}000$ to $129{,}000$. Both
edges are protocol choices rather than data limits. The lower edge keeps
the five-checkpoint history window, which reaches back to step $2{,}000$,
clear of the $1{,}430$-step learning-rate warmup. The upper edge leaves a
complete future window for the longest horizon, since
$143{,}000-14{,}000=129{,}000$ at $\tau=14$ checkpoints.

Public training-loss telemetry and the official zero-shot evaluations are
used only as external context, and their provenance is recorded in panel
(b) of the table, including the two sizes whose released telemetry is
assembled from more than one run. One size carries a checkpoint defect:
revision \texttt{step116000} of the 1B series is wrong upstream, and the
cells whose history or future window touches it are masked at read time.
The masking windows are listed in the table footnote and are applied
wherever 1B appears.

\subsection{Measurement Clocks, Anchors, and Seeds}
\label{app:anchor_construction}

\paragraph{Clocks.}
One probe step is one epoch on the vision side and one released
checkpoint on the pretraining side. A checkpoint step spans $1{,}000$
optimization steps, or about $2.10$ billion tokens at $2{,}097{,}152$
tokens per step. The two clocks index different amounts of optimization,
which is why cross-regime readings compare the ordering within a matched
trajectory rather than levels across regimes.

\paragraph{Anchors and windows.}
Both regimes use horizon $\tau=5$ and history length $\npast=5$ unless
stated otherwise, with the reference direction $\dw_{a-1}$ and the future
target $\dw_a^{(\tau)}$ as defined in \mainref{sec:methods}. Vision
probes are evaluated at anchor epochs $\{5,10,20,30,40\}$. An anchor is
admissible only when its full history window and its full future window
lie inside the recorded trajectory, which is what fixes the edges of the
pretraining anchor domain above. The pretraining panel additionally
records $\tau\in\{1,2,3,5,10,14\}$ so that horizon sensitivity can be
read from the same shards.

\paragraph{Seeds and determinism.}
The panel trains at seed $42$, with replicates at $43$ and $44$ on the
five cells listed in Table~\ref{tab:appA_families}. On the measurement
side, three seeds are fixed and shared across every lane so that a
recomputation reproduces the same numbers: the rank-selection sample
(Appendix~\ref{app:rank_selection}), the per-layer group subsample
(Appendix~\ref{app:grouping_definitions}), and the permutation-null draws
(Appendix~\ref{app:lrgf_calibration}), whose seed is composed from the
draw index, anchor, layer, and segment.

\paragraph{Compute.}
The vision corpus was trained on a single workstation with two NVIDIA RTX
A6000 GPUs ($48$\,GB each), one run per GPU. Median wall-clock for a
$50$-epoch run, taken over each architecture's optimizer cells, is about $12$
minutes for the MLP, $15$ for ViT, $22$ for AlexNet, $23$ for ViT-mid, and
$38$ for the ResNet; the slowest single run is AlexNet under Muon at about
$4.8$ hours, where the orthogonalization step dominates on the largest weight
matrices. Probe and panel computation is CPU-only and in double precision, on
a $16$-core machine with $128$\,GB of memory: one pretraining anchor of the
full panel protocol, including its permutation nulls, takes about $10$ minutes
at 160M and $25$ at 410M. The panels for the two largest sizes ran on
large-memory CPU nodes ($64$ cores, $2$\,TB) with anchors processed in $8$ to
$10$ parallel lanes, at roughly $17$ to $19$ hours per size; the
embedding-spectrum sweep is checkpoint-bound, at seconds to about a minute per
revision. The stored checkpoint packs occupy $24$, $54$, $126$, $311$, and
$436$\,GB for 70M, 160M, 410M, 1B, and 1.4B. Software: Python~3 with PyTorch,
used only to deserialize checkpoints, together with NumPy, SciPy, and pandas.

\section{Probe Conventions, Calibration, and Robustness}
\label{app:probe_conventions}

\mainref{sec:methods} defers five implementation details to this appendix:
the exact predictor constructions, the rank-selection rule, the null
generation procedure, the grouping rules, and the numerical conventions.
Appendices~\ref{app:grouping_definitions} to~\ref{app:zero_handling} give
them, in that order. Appendix~\ref{app:cut_sensitivity} then reports what
the group-level readouts do when the grouping convention is changed,
together with the full versions of the within-family comparisons
summarized in \mainref{sec:probe_relations}, and
Appendix~\ref{app:reporting_discipline} states the reporting rules that
follow from those measurements.

\subsection{Measurement Units and Grouping Conventions}
\label{app:grouping_definitions}

Every group-level readout in this paper is conditional on a rule that
partitions a tensor's scalars into measurement units. We call such a rule a
\emph{cut}. Tables~\ref{tab:appB_cuts_vision} and~\ref{tab:appB_cuts_pythia}
list all cuts used in the paper together with their group sizes and their
status with respect to chance passing.

\begin{table*}[tp]
\centering
\small
\setlength{\tabcolsep}{4pt}
\renewcommand{\arraystretch}{1.15}
\begin{tabularx}{\textwidth}{@{} l >{\raggedright\arraybackslash}p{0.17\textwidth} l X >{\raggedright\arraybackslash}p{0.19\textwidth} @{}}
\toprule
Cut & One group is & $D$ & What it reads & Chance status at $\theta{=}0.7$ \\
\midrule
\multicolumn{5}{@{}l}{\emph{Flat protocol, the main measurement convention}} \\
flat $K{=}16$
 & $16$ contiguous scalars of the row-major flattened tensor
 & $16$
 & Between-group ESA dispersion sits at the binomial sampling floor
   ($1.1\times$, median over $30$ cells), so blocks are statistically
   homogeneous. Bulk group-level FRR hugs the chance floor (median excess
   $.014$, against $.148$ for the auxiliary class). Cross-architecture
   ordering claims do not survive recutting.
 & Not immune: of the $31.2\%$ bulk pass rate, about $25$\,pp is chance
   ($\mathrm{null}_{\theta}$ median $.23$ to $.25$), so the calibrated
   readout is required. \\
\midrule
\multicolumn{5}{@{}l}{\emph{Structural cuts, physical and functional axes}} \\
filter $3{\times}3$
 & one ResNet output channel's full $3{\times}3$ kernel stack
 & $27$--$2304$
 & Names the unit behind ResNet's flat readout: $66$ to $87\%$ of ResNet
   flat $\mathrm{LRGF}$ is intra-kernel coherence, destroyed by
   within-tensor permutation ($1{\times}1$ shortcut $\Delta{=}.0025$).
   $\mathrm{FRR}{<}0.7$: $1.5\%$. ESA dispersion $1.9\times$ floor, not
   coupled to FRR.
 & Immune. \\
row $1{\times}1$
 & one $1{\times}1$-convolution row
 & $64$--$128$
 & Essentially no pockets: $\mathrm{FRR}{<}0.7 = 0.1\%$.
 & Immune. \\
head $qkv$
 & per-head q/k/v row block of the fused $qkv$ tensor
 & $4096$ / $12544$
 & Strongest structural unit: ESA dispersion $4.4\times$ floor;
   $\mathrm{FRR}{<}0.7$ of $19.8\%$ (ViT-mid) and $5.2\%$ (ViT); restores
   group-level ESA--FRR coupling (cell-wide median $-.42$ against $-.16$
   for flat).
 & Immune (pass rate ${\approx}0$ for $D{\geq}448$). \\
head proj
 & per-head projection rows
 & $4096$ / $12544$
 & $\mathrm{FRR}{<}0.7$: $10.0\%$ (ViT-mid), $1.2\%$ (ViT).
 & Immune. \\
neuron pair
 & hidden neuron fan-in and fan-out concatenated
 & $256$ / $448$
 & $\mathrm{FRR}{<}0.7$: $12.7\%$ (ViT-mid), $6.9\%$ (ViT).
 & Immune. \\
\midrule
\multicolumn{5}{@{}l}{\emph{Magnitude cut, causal activity axis}} \\
magnitude, 16 bins
 & per-tensor rank by pre-anchor window $\lVert\dw\rVert_2$, $16$ equal
   bins (causal)
 & $108$--$36864$
 & Strongest single-axis density sorter and not a raiser of totals:
   monotone $16$-bin gradient ($\mathrm{FRR}{<}0.7$ pooled $0\%$ to
   $20.1\%$; $\mathrm{ESA}{>}0.7$ $0.1$ to $27.2$; $\mathrm{VCS}{>}0.4$
   $0.2$ to $36.4$); the top three bins concentrate $2$ to $3\times$
   (ViT-mid $27.9\%$ against head $qkv$ $19.8\%$); totals slightly lower
   than structural ($0.4/3.2/8.1$ against $1.3/6.7/12.9$ for ResNet, ViT,
   ViT-mid).
 & Immune (all $D{\geq}108$). \\
\bottomrule
\end{tabularx}
\caption{Grouping conventions on the vision panel. $D$ is the number of
scalars per group. ``Immune'' means the membership-resample null passes
$\theta{=}0.7$ at a rate near zero, so the raw pass fractions carry a negligible chance contribution. The flat $K{=}16$ protocol is the one cut where a large share of
raw passes is chance, which is what motivated the calibrated readout
$\mathrm{LRGF}_{\mathrm{cal}}$. Levels are not comparable across cuts,
since group sizes and chance floors differ, so every level claim in this
paper names its cut. Structural and magnitude readouts are pooled over the
$18$ recipe cells of the group-level lanes and anchors $5$ to $40$.}
\label{tab:appB_cuts_vision}
\end{table*}

\begin{table*}[tp]
\centering
\small
\setlength{\tabcolsep}{4pt}
\renewcommand{\arraystretch}{1.15}
\begin{tabularx}{\textwidth}{@{} l >{\raggedright\arraybackslash}p{0.17\textwidth} l X >{\raggedright\arraybackslash}p{0.19\textwidth} @{}}
\toprule
Cut & One group is & $D$ & What it reads & Chance status at $\theta{=}0.7$ \\
\midrule
positional (reference)
 & $qkv$: per-head block; MLP: neuron pair
 & $10^{3}$--$10^{5}$
 & Reference convention of the Pythia protocol. Within-tensor permutation
   \emph{raises} the flat $qkv$ reading ($\Delta{=}-.140$, structural
   $.099$ against permuted $.239$), the signature of comoving mass
   concentrated in a few strong coordinates.
 & Bare $\theta$ blind at large $D$; calibrated readout used. \\
\texttt{attn\_out} head
 & per-head output block
 & $32768$
 & Excess $+.263$ at step $70$k while the bare-$\theta$ column reads
   $.000$, the sharpest case of large-$D$ blindness.
 & Calibrated only. \\
$W_{qkv}$ row
 & one $qkv$ row, the natural coherence scale
 & $512$ (70M)
 & Absolute pockets revive at row granularity: $\mathrm{FRR}{<}0.7$ stable
   at $4$ to $5\%$, against $\approx 0$ at head level. Row-length
   invariance across $D{=}128$ to $3072$ (410M peaks $.432/.434/.438$ at
   $D{=}1024/512/256$; 160M scan; 1B below $0.5\%$ after masking).
 & $\theta$ legal (analytic floor ${\approx}0$). \\
magnitude
 & within (layer, segment), coordinates sorted by pre-anchor $L_2$ and
   chunked to the same $(n,D)$
 & as reference
 & Loses to head grouping on $qkv$ at step $10$k
   ($\mathrm{LRGF}_{\mathrm{cal}}$ $.333/.389$ against $.104/.069$ at 70M
   and 160M); MLP sensitivity is local (70M layer 5 mid-training, about
   $3\times$) with $7$ of $10$ grid cells negative; at most $0.12\%$ on
   absolute $\theta$ across five sizes.
 & Null shared with the positional cut (grouping-independent). \\
MLP input half
 & neuron input row ($w_{\mathrm{in}}$)
 & $d_{\mathrm{model}}$
 & Reads near zero (mean $0.04\%$).
 & $\theta$ legal. \\
MLP output half
 & neuron output column ($w_{\mathrm{out}}$)
 & $d_{\mathrm{model}}$
 & Real emergence carried by the deepest layer: the 70M layer-5 wave rises
   from $64$k to a peak of $5.4\%$ at $96$k. Across five model sizes the
   output half is the only MLP cut with signal (peaks $1.8$ to $15.7\%$),
   with the input half and the magnitude cut near zero.
 & $\theta$ legal. \\
\bottomrule
\end{tabularx}
\caption{Grouping conventions on the Pythia side. All Pythia readings are
registered as correspondences and are not used to certify one another.
Column conventions follow Table~\ref{tab:appB_cuts_vision}. The last two
rows are the recut used for the granularity ladder in
Appendix~\ref{app:cut_sensitivity}.}
\label{tab:appB_cuts_pythia}
\end{table*}

\paragraph{Flat $K{=}16$, the main protocol.}
Each tensor's parameter vector is flattened in row-major order and split
into consecutive disjoint blocks of $K{=}16$ scalars. A trailing block
shorter than $K$ is dropped, and a tensor holding fewer than $K$ scalars
carries no group at all; on the vision panel this removes only the
ten-parameter classifier bias of each model. The cut is cheap, uniform in
group size, and identical across architectures, which is what keeps the
group size, and therefore the chance geometry, matched in cross-architecture
comparisons. It is not aligned to the architecture. In row-major layout a
flat block of a convolutional tensor is approximately one $3{\times}3$
kernel, whereas for a feature-dimension linear tensor it is $16$ adjacent
input dimensions with no functional identity, and the two readings are not
the same measurement.

\paragraph{Structural cuts (vision).}
Structural cuts take the unit from the architecture instead of from the
memory layout. For the ViT family a group is one head's q, k, or v row
block of the fused $qkv$ tensor, one head's projection rows, or one hidden
neuron's fan-in and fan-out concatenated. For ResNet a group is one output
channel's full $3{\times}3$ kernel stack, or one $1{\times}1$ convolution
row. Group sizes run from $27$ to $12{,}544$.

\paragraph{Magnitude cut (vision).}
Within each tensor, coordinates are ranked by the $L_2$ norm of their
pre-anchor update window and split into $16$ equal bins. The ranking reads
only pre-anchor history, so the cut is causal and carries no information
from the target window.

\paragraph{Pythia segments.}
Each transformer layer contributes four segments. The fused $[3d,d]$ $qkv$
tensor is stored head-major, and one group is the block belonging to one
head and one of q, k, v, of size $d_{\mathrm{head}}\,d$ ($32{,}768$ at
70M); groups are enumerated part-major, that is, all q blocks, then all k,
then all v. In \texttt{attention.dense} the input axis carries the
concatenated head outputs, so one group is one head's column block. In the
MLP one group is a neuron pair: row $j$ of \texttt{dense\_h\_to\_4h}
joined with column $j$ of \texttt{dense\_4h\_to\_h}, of size $2d$. In the
vector segment each vector parameter is its own group. The two embedding
matrices form separate segments and stay outside every aggregation
(Appendix~\ref{app:pythia_partition}). Two finer conventions are used for
the granularity studies in Appendix~\ref{app:cut_sensitivity}: one $W_{qkv}$
row ($D{=}d$), and the two MLP halves taken separately.

\paragraph{Group subsampling.}
Layers with more than $1{,}500$ groups are subsampled to that cap before
group-level statistics are formed, with a seed derived from the layer name
so that the same groups are drawn on any recomputation. Tensor-level and
pooled readouts use all groups.

\subsection{Predictor Constructions}
\label{app:predictor_constructions}

All three predictors read only the $\npast$ one-step updates before the
anchor. For a unit $S$ write the history columns as $h_1,\ldots,h_m$ with
$m=\npast$, so that $h_m=\dw_{a-1,S}$.

\paragraph{Inertia.}
$\widehat f_{\mathrm{in}}=\tau\,h_m$.

\paragraph{Linear trend.}
Per coordinate, an affine model $h_s\approx\alpha+\beta s$ is fitted by
least squares over $s=0,\ldots,m-1$, and the fitted updates are summed over
the next $\tau$ steps,
\begin{equation}
\widehat f_{\mathrm{lin}}
=\sum_{q=0}^{\tau-1}\bigl(\alpha+\beta(m+q)\bigr)
=\tau\alpha+\beta\sum_{q=0}^{\tau-1}(m+q).
\end{equation}
The fit is on updates rather than on weights, so the output is a
displacement and not an extrapolated checkpoint.

\paragraph{DMD.}
The construction has three steps.
\emph{(i) Standardization.} Each coordinate row of the history is divided
by its standard deviation over the $m$ columns. Rows whose standard
deviation falls below $10^{-30}$ are left unscaled, and any unit containing
such a row is predicted as zero. Write $\tilde H$ for the standardized
history.
\emph{(ii) Fit.} With $X=[\tilde h_1,\ldots,\tilde h_{m-1}]$ and
$Y=[\tilde h_2,\ldots,\tilde h_m]$, take the thin SVD
$X=U\Sigma V^{\top}$, truncate to $r=\min(|S|,m-1)$, which is four columns
under the default protocol, and form the reduced map
\begin{equation}
\tilde A=U_r^{\top}\,Y\,V_r\,\Sigma_r^{-1},
\end{equation}
with singular values below $10^{-10}$ inverted to zero.
\emph{(iii) Rollout.} Eigendecompose $\tilde A=W\Lambda W^{-1}$, lift the
eigenvectors to $\Phi=U_rW$, and expand the last standardized update in
them, $b=\Phi^{\dagger}\tilde h_m$. The $\tau$-step prediction is
\begin{equation}
\widehat f_{\DMD}
=\Re\Bigl\{\Phi\Bigl(\sum_{k=1}^{\tau}\Lambda^{k}\Bigr)b\Bigr\},
\label{eq:dmd_rollout}
\end{equation}
with each coordinate rescaled by the standard deviation removed in step
(i). Summing $\Lambda^{k}$ over $k=1,\ldots,\tau$ instead of applying
$\Lambda^{\tau}$ is what makes the output the cumulative displacement over
the horizon, which is the quantity the target $\dw_{a,S}^{(\tau)}$
measures.

\paragraph{Containment in the recent-history span.}
Because $\Phi=U_rW$, the prediction lies in the column space of $U_r$, and
the row standardization cancels exactly between the fit in step (ii) and
the rescaling in step (iii). The DMD prediction therefore lies in the span
of the first $m-1$ raw history columns, which is contained in the span the
FRR projector retains whenever that projector covers the full history
rank. This is the containment behind reading FRR as the directional
headroom available to a predictor built from the recent history, as in
\mainref{sec:probe_relations}: the residual fraction that FRR measures
bounds how well any vector drawn from that span can align with the
realized displacement.

\subsection{Rank Selection}
\label{app:rank_selection}

The FRR projector retains a rank chosen by the Gavish--Donoho optimal hard
threshold. For a unit whose history matrix has singular values
$s_1\ge\cdots\ge s_p$ and aspect ratio
$\beta=\min(|S|,m)/\max(|S|,m)$, the threshold is
\begin{equation}
\begin{split}
\theta_{\mathrm{GD}}&=\omega(\beta)\,\mathrm{median}(s),\\
\omega(\beta)&=0.56\beta^{3}-0.95\beta^{2}+1.82\beta+1.43,
\end{split}
\end{equation}
and the retained rank is $\max\bigl(2,\lvert\{i:s_i>\theta_{\mathrm{GD}}\}
\rvert\bigr)$. The floor of two is part of the protocol rather than a
numerical guard, and the rank is capped at the number of history columns.

Applying the rule per unit would force one projection per group. The
default rule instead draws $256$ units per model--anchor cell with a fixed
seed, computes their Gavish--Donoho ranks, and applies the median rank
uniformly to all units of that cell, which keeps the projection batched at
one rank per cell. The flat-protocol group readouts reported in this paper
are computed at the full history rank $r=\npast=5$. The pooled bulk
group-level FRR median is $0.815$, just below the isotropic reference
$\sqrt{(D-r)/D}=0.829$ that this rank and $D{=}16$ imply. This is the same
isotropic-reference construction that \mainref{fig:pocket_scarcity} measures
excess against, evaluated there at each structural group's own size rather
than at $D{=}16$.

\subsection{Null Generation and the Calibrated Readout}
\label{app:lrgf_calibration}

FRR compares a $D$-dimensional target with an $r$-dimensional subspace, so
a target with no temporal structure at all still passes a fixed threshold
at a rate that depends on $D$ and $r$. A readout compared across group
sizes therefore needs a chance reference computed at its own geometry.

\paragraph{Membership resampling.}
Within one pool of comparable units, a tensor on the vision side and a
(layer, segment) block on the Pythia side, all scalar coordinate histories
are collected, permuted, and re-chunked into groups whose size multiset is
exactly that of the observed cut. FRR is then recomputed on the permuted
groups under the same window, horizon, rank rule, and normalization. The
permutation leaves every coordinate's own trajectory untouched and destroys
only which coordinates share a group, so the resulting distribution
isolates the contribution of group membership and nothing else.

\paragraph{Pooling and seeds.}
Null values are pooled per (segment, group size, horizon, protocol). The
number of draws per segment is
$\max\bigl(5,\lceil 200/n_{\mathrm{groups}}\rceil\bigr)$, so that at least
about $200$ null values stand behind every threshold. Each draw takes a
seed composed from the draw index, anchor, layer, and segment, so any lane
that recomputes the null reproduces it.

\paragraph{The two readouts.}
The fixed-threshold readout $\mathrm{LRGF}_{\theta}$ counts groups with
$\mathrm{FRR}<\theta$, with $\theta=0.7$ throughout. The calibrated readout
$\mathrm{LRGF}_{\mathrm{cal}}$ replaces $\theta$ by the $\alpha=0.05$
quantile of the matched null pool and is reported as excess over $\alpha$.
$\mathrm{LRGF}_{\mathrm{cal}}$ is a comparability convention: it places
segments whose group size and null geometry differ on a single axis. It
does not correct individual FRR values, and it does not make a level
comparable to a level read under a different cut. Where the fixed threshold
is used, we also report the null pool's own pass rate at $\theta$, written
$\mathrm{null}_{\theta}$, which is the share of the observed fraction that
membership alone would produce.

\subsection{Numerical Conventions and the Sign Rule}
\label{app:zero_handling}

\paragraph{The sign rule and zero displacement.}
Signs are compared exactly on the stored floating-point values, with
$\operatorname{sign}(0)=0$ and no tolerance band. A coordinate that does
not move in either interval therefore counts as an agreement, and a
coordinate that moves in exactly one of the two counts as a disagreement.
We keep the exact rule because it leaves ESA a parameter-free function of
the realized trajectory, but it makes ESA sensitive to coordinates whose
displacement is zero or numerically negligible over one or both intervals,
in a way that VCS is not: VCS weights each coordinate by its displacement
magnitude, so those coordinates barely enter it.

This is the mechanism behind the two departures from the parameter-free
reference curve $\mathrm{ESA}=1-\arccos(\mathrm{VCS})/\pi$ in
\mainref{fig:family_panels}(a). At the five reported anchors the two
departing cells are AlexNet with AdaGrad and AlexNet with Adam, in which up
to $79\%$ of coordinates have zero reference displacement and up to $69\%$
are zero over both intervals. Their median vertical departure from the
reference curve is $+0.276$ and $+0.221$. Restricting both probes to the
coordinates that move in both intervals reduces those departures to
$+0.008$ and $+0.007$, while the median absolute departure over the
remaining $28$ cells is $0.003$ and is unchanged by the restriction.

\paragraph{Degenerate units.}
VCS is undefined when either vector has norm below $10^{-30}$; such units
are recorded as missing and dropped from denominators rather than counted
as zero. SESA is undefined when the selector is empty, and those units are
dropped in the same way. Non-finite FRR values leave the $\mathrm{LRGF}$
denominator.

\paragraph{Other numerical choices.}
The probe layer works in double precision. FRR denominators carry an
additive $10^{-12}$ guard. FRR is computed on the raw history, whereas the
DMD fit standardizes coordinate rows. On the Pythia side a second FRR
protocol normalizes each history column to unit norm within the group, with
a $10^{-30}$ guard, and degenerate columns are counted rather than
absorbed; both FRR protocols are reported side by side wherever the
column-normalized variant is used.

\paragraph{Group-mean and full-vector VCS.}
\mainref{fig:family_panels}(a) plots one point per model--configuration cell
and anchor, pairing that cell's ESA with its \emph{group-mean} VCS, the mean
of the per-group cosines. The full-vector cosine of the same cell is recorded
alongside it and differs because it weights coordinates by magnitude across
the whole unit rather than within each group. The two are close on most cells
(median difference $+0.006$, interquartile range $-0.017$ to $+0.023$) but
diverge sharply on a few, up to $\pm0.45$. The group mean is the readout that
belongs on the parameter-free reference curve: its median absolute departure
from $1-\arccos(\mathrm{VCS})/\pi$ is $0.005$, against $0.011$ for the
full-vector cosine on the same cells.

\subsection{Grouping Sensitivity and Within-Family Checks}
\label{app:cut_sensitivity}

\subsubsection{What the Flat Protocol Reads}

\begin{figure*}[tp]
  \centering
  \includegraphics[width=\textwidth]{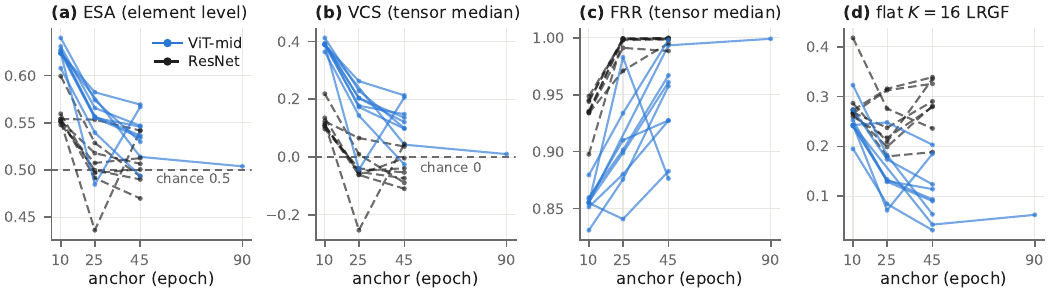}
  \caption{Probes that use no grouping convention, against the flat
  readout (bulk scope; one line per run, $8$ matched ViT-mid and ResNet
  pairs; the long-cosine ViT-mid run extends to epoch $90$). On all three
  cut-free probes, (a) element-level ESA, (b) tensor-level VCS, and
  (c) tensor-level FRR, where lower means more predictable, ViT-mid reads
  as more predictable than ResNet at every anchor, while (d) the flat
  $K{=}16$ $\mathrm{LRGF}$ reference orders the same runs and anchors the
  other way ($7$ to $8$ of $8$ pairs). Canonical protocol
  ($\npast{=}5$, $\tau{=}5$, $\theta{=}0.7$); dashed grey lines mark
  chance (ESA $0.5$, VCS $0$).}
  \label{fig:appB_e1}
\end{figure*}

A dedicated audit compared the flat readout with three probes that involve
no grouping at all: element-level ESA, tensor-level VCS, and tensor-level
FRR. On all three, ViT-mid reads as more predictable than ResNet at every
anchor of every matched pair, whereas the flat $\mathrm{LRGF}$ reference
orders the same runs and anchors the other way
(Figure~\ref{fig:appB_e1}). Under a within-tensor membership permutation
the flat ordering reverses to ViT-mid at or above ResNet in $24$ of $24$
pair--anchor comparisons.

\begin{figure*}[tp]
  \centering
  \includegraphics[width=\textwidth]{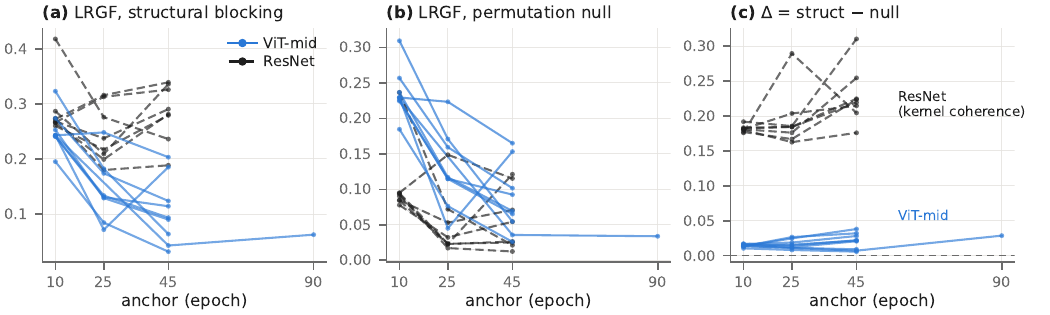}
  \caption{Permutation-null decomposition of the flat $K{=}16$
  $\mathrm{LRGF}$ (bulk scope; one line per run): (a) structural blocking,
  (b) the membership-permutation null (mean of $5$ draws, shading gives min
  to max), and (c) their difference $\Delta$, which is the intra-block
  coherence contribution. Permutation destroys $66$ to $87\%$ of ResNet's
  flat readout and the destroyed part localizes on the $3{\times}3$
  kernels ($1{\times}1$ shortcut convolutions: $\Delta\approx.003$),
  whereas $82$ to $94\%$ of ViT-mid's readout survives and the
  architecture ordering flips to ViT-mid at or above ResNet on the null
  column ($24/24$ pairs). A flat-$\mathrm{LRGF}$ level is therefore a
  statement about a grouping convention as well as about an
  architecture.}
  \label{fig:appB_e2}
\end{figure*}

The decomposition names the mechanism (Figure~\ref{fig:appB_e2}).
Permuting block membership destroys $66$ to $87\%$ of ResNet's flat
$\mathrm{LRGF}$, and the destroyed part localizes on the $3{\times}3$
kernels, with $1{\times}1$ shortcut convolutions contributing
$\Delta\approx.003$. On ViT-mid, $82$ to $94\%$ of the reading survives the
same permutation. A flat-$\mathrm{LRGF}$ level is therefore a statement
about a grouping convention as much as about an architecture.

Two further properties of the flat protocol matter for reading the main
paper. Between-group ESA dispersion sits at the binomial sampling floor
($1.1\times$, median over $30$ cells), so flat blocks are statistically
homogeneous, and group-level coupling between ESA and FRR is largely erased
(cell-wide median $-.16$). And bulk group-level FRR sits close to its
chance floor: of the $31.2\%$ of bulk groups passing $\theta{=}0.7$, about
$25$ percentage points are chance passing ($\mathrm{null}_{\theta}$ median
$.23$ to $.25$). This is the concrete reason the time-axis readouts use
$\mathrm{LRGF}_{\mathrm{cal}}$.

\subsubsection{Structural and Magnitude Cuts on the Vision Panel}

\begin{figure*}[tp]
  \centering
  \includegraphics[width=\textwidth]{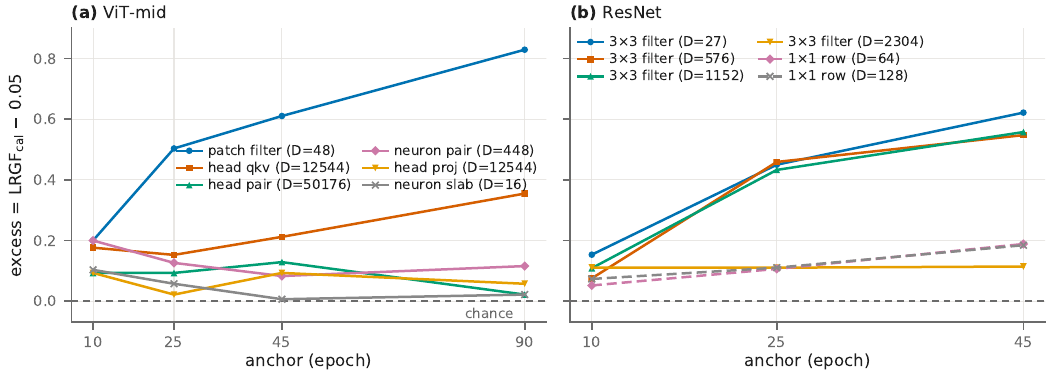}
  \caption{Structural-axis cuts under the calibrated readout: excess
  $=\mathrm{LRGF}_{\mathrm{cal}}-\alpha$ ($\alpha{=}0.05$, so zero is
  chance), median over runs. (a) ViT-mid: physically defined groups surface
  above-chance structure that the flat axis does not resolve, with whole
  patch filters reaching $.61$ and head-level $qkv$ groups $.21$ at anchor
  $45$, while slab and projection cuts sit near chance. (b) ResNet: whole
  $3{\times}3$ filters reach $.55$ to $.62$ and $1{\times}1$ rows $.18$.
  Each cut carries a size-matched membership-permutation null that
  preserves the group-size multiset, and groups use only pre-anchor
  information. Group size differs across cuts, so levels are comparable
  only through the calibrated excess.}
  \label{fig:appB_c1}
\end{figure*}

Recutting along physical or functional axes changes what the instrument
resolves. Structural units raise between-group ESA dispersion off the floor
(head $qkv$ $4.4\times$, filter $3{\times}3$ $1.9\times$, neuron pair
$1.7\times$, row $1{\times}1$ $1.2\times$, against $1.1\times$ for flat)
and restore group-level ESA--FRR coupling (cell-wide median $-.42$ against
$-.16$). The restoration is architecture-resolved: ViT and ViT-mid couple
in every recipe cell ($-.28$ to $-.77$), while ResNet's plain recipes stay
flat (SGD $-.14$; momentum, Adam, and Muon near $0$) and only the recipe
cells couple (weight decay $-.45$, SAM $-.35$, AdaGrad $-.30$). Under the
calibrated readout the same reordering is visible over training anchors
(Figure~\ref{fig:appB_c1}): whole patch filters and head-level $qkv$ groups
carry clear above-chance excess, while slab and projection cuts sit near
chance.

\begin{figure}[tbp]
  \centering
  \includegraphics[width=\columnwidth]{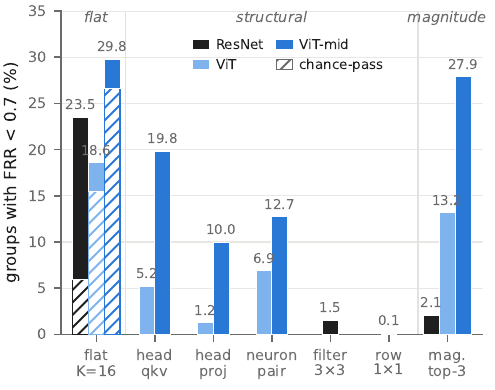}
  \caption{Pocket density by grouping cut and architecture on the vision
  panel: share of groups with $\mathrm{FRR}<0.7$, pooled over the $18$
  recipe cells and anchors $5$ to $40$ of the group-level lanes (bulk
  tensors, out-of-window target). The structural cuts and the magnitude cut
  (top three of $16$ causal bins) are chance-immune at $\theta{=}0.7$, so the chance contribution to their bars is negligible. The flat $K{=}16$ bars are pooled
  over the same $18$ cells weighted by group count and are stacked: the
  hatched portion is the measured chance-pass fraction under the
  membership-resample null ($6.0$, $15.5$, and $26.6$\,pp of the
  $23.5/18.6/29.8\%$ totals for ResNet, ViT, and ViT-mid). Most of the
  ViT-family flat readout is chance passing, and most of the ResNet flat
  readout is intra-kernel coherence. Group size differs across cuts, so
  levels are comparable within a bar group only in the chance-immune
  sense.}
  \label{fig:appB_density}
\end{figure}

Pocket density under structural cuts is architecture-stratified
(Figure~\ref{fig:appB_density}): $\mathrm{FRR}<0.7$ reaches $19.8\%$ of
ViT-mid head-$qkv$ groups and $12.7\%$ of neuron pairs, against $1.5\%$ for
ResNet kernels and $0.1\%$ for $1{\times}1$ rows. These cuts have large
group sizes, where the membership-resample null passes $\theta{=}0.7$ at a
rate near zero, so essentially none of the passes is attributable to chance, unlike the flat bars in the same figure. Structural excess decays over training, from a
median of $.070$ at anchor $5$ to $.036$ at anchor $40$, which places
pockets in an early-to-mid-training population.

\begin{figure*}[tp]
  \centering
  \includegraphics[width=\textwidth]{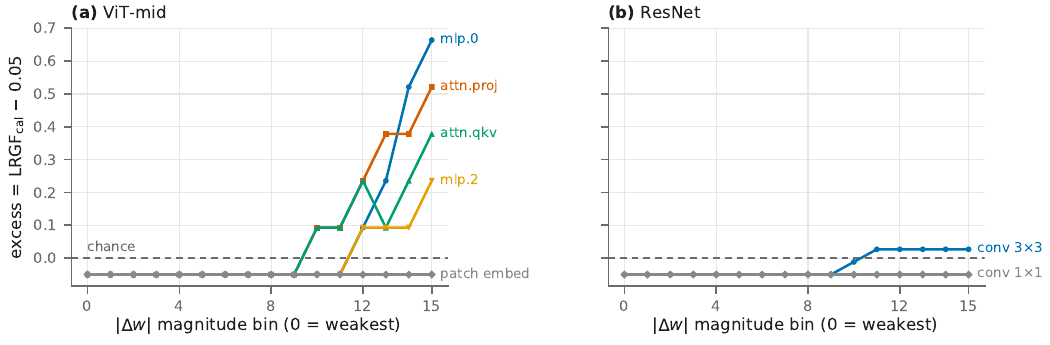}
  \caption{Magnitude-stratified cut at anchor $45$: groups are $16$
  equal-width bins of the per-tensor pre-anchor $\lVert\dw\rVert_2$ rank
  (bin $0$ weakest, bin $15$ strongest), reported as excess over chance,
  median over runs; the floor at $-0.05$ means no group passes.
  (a) ViT-mid feature-dimension tensors show a monotone concentration
  signature (top bins $.38$ to $.66$), while the patch embedding stays at
  the floor because its structure lies on the spatial axis
  (Figure~\ref{fig:appB_c1}). (b) ResNet reads near zero under the same
  cut. This is the strongest single-axis sorter on the ViT side, and its
  architecture contrast runs opposite to the flat readout.}
  \label{fig:appB_c2b}
\end{figure*}

The magnitude cut is the strongest single-axis sorter of pocket density and
not a raiser of totals. Totals sit slightly below the structural cut
($0.4/3.2/8.1\%$ against $1.3/6.7/12.9\%$ for ResNet, ViT, and ViT-mid),
but the density gradient is monotone across all $16$ bins and steep at the
top: pooled $\mathrm{FRR}<0.7$ rises from $0\%$ at bin $0$ to $20.1\%$ at
bin $15$, $\mathrm{ESA}>0.7$ from $0.1\%$ to $27.2\%$, and
$\mathrm{VCS}>0.4$ from $0.2\%$ to $36.4\%$, three instrument families
ordered the same way along one causal axis (Figure~\ref{fig:appB_c2b}). The
top three bins concentrate density beyond the best structural cut (ViT-mid
$27.9\%$ against head $qkv$ $19.8\%$). ResNet stays low under every probe of
the magnitude cut ($2.1/4.4/12.4\%$ for FRR, ESA, and VCS on the top three
bins), so the architecture stratification is itself cut-robust.

\begin{figure}[tbp]
  \centering
  \includegraphics[width=\columnwidth]{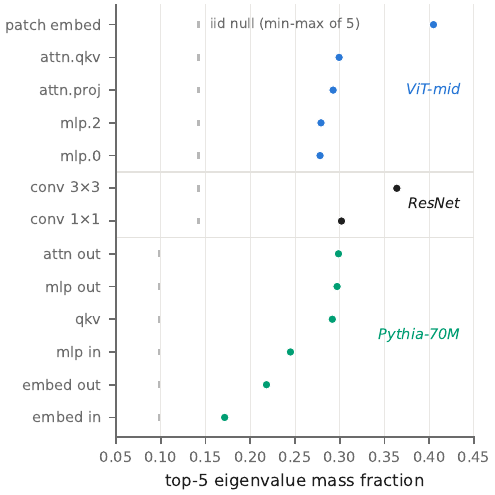}
  \caption{Spectrum diagnostic at the late anchors (anchor $45$ on the
  vision panel, step $70$k on Pythia): top-five eigenvalue mass of the
  coordinate-trajectory correlation spectrum per tensor family, against a
  simulated independent null of matching shape (grey band gives min to max
  over $5$ draws), median over runs. Every family of all three
  architectures exceeds the null by two to four times, so co-moving
  structure is present in every family; which grouping axis can access it
  is what differs by architecture.}
  \label{fig:appB_c3}
\end{figure}

An axis-independent check bounds what any cut could reach
(Figure~\ref{fig:appB_c3}): the top-five eigenvalue mass of the
coordinate-trajectory correlation spectrum exceeds a matched independent
null by two to four times in every tensor family of all three
architectures. Co-moving structure is present everywhere; which grouping
axis can access it is what differs by architecture.

\subsubsection{Pythia Cuts and the Granularity Ladder}

The Pythia readings are registered as correspondences and are not used to
certify one another. The flat audit finds a third permutation signature
here, distinct from both vision architectures: on $qkv$, permuting block
membership \emph{raises} the flat reading ($\Delta=-.140$, structural
$.099$ against permuted $.239$), which is what comoving mass concentrated
in a few strong coordinates produces when random blocks inherit it and
contiguous blocks isolate it.

Head grouping and magnitude grouping are adjudicated at anchor $10$k. Head grouping retains far stronger calibrated signal on $qkv$ ($\mathrm{LRGF}_{\mathrm{cal}}$
$.333/.389$ against $.104/.069$ at 70M and 160M), so the head unit is an
organizational axis of its own rather than a shadow of coordinate
magnitude. On the MLP, magnitude sensitivity is local (70M layer 5 in
mid-training, about $3\times$) and the grid is negative in $7$ of $10$
cells, so the MLP's modest calibrated readout is modest under both
conventions.

Granularity is a separate axis. At head granularity ($D{=}32{,}768$) the
absolute $\theta$ columns read near zero while calibrated excess is large,
for instance $+.263$ for \texttt{attn\_out} at anchor $70$k, which is the
sharpest case of large-$D$ blindness in the fixed threshold. Stepping down
to one $W_{qkv}$ row ($D{=}512$) revives an absolute pocket population,
with $\mathrm{FRR}<0.7$ stable at $4$ to $5\%$ of rows and the $D{=}512$
and $D{=}128$ distributions nearly coincident. The row convention is
invariant across $D{=}128$ to $3072$ (410M peak fractions $.432/.434/.438$
at $D{=}1024/512/256$; a 160M scan over the same range; 1B below $0.5\%$
throughout after masking), so no per-size adjustment is needed. Recutting
the MLP resolves an asymmetry between its halves: the input half and the
magnitude cut read near zero on absolute $\theta$ (means $0.04\%$ and
$0.01\%$), while the output half carries a late wave in the deepest layer
(70M layer 5: rising from $64$k to a peak of $5.4\%$ at $96$k, fading by
$128$k). Across five model sizes the output half is the only MLP cut with
signal (peaks $1.8$ to $15.7\%$).

\subsubsection{Within-Family Agreement, per Cell}

The two within-family summaries quoted in \mainref{sec:probe_relations}
are medians over per-cell Spearman correlations, each computed across the
tensors of one model--configuration cell. Their dispersion is as follows.
Out-of-window and in-window FRR correlate at a median $\rho_S=0.83$, with
per-cell values from $0.53$ to $0.97$; the low end is occupied by AdaGrad
cells (ViT $0.62$, ViT-mid $0.63$) together with one further cell at
$0.53$. SESA under the DMD and linear-trend selectors correlates at a
median $\rho_S=0.89$, with per-cell values from $0.44$ to $1.00$; the one
low value is AlexNet with AdaGrad ($0.44$), which is also one of the two
cells identified in Appendix~\ref{app:zero_handling}. The two displacement-
direction readouts correlate at a median $\rho_S=0.93$. Correlations
between the DMD and linear-trend \emph{lift} readouts are much weaker
(median $0.15$); lift is a within-predictor quantity, and the two
selectors define different denominators, so the comparison is reported
only as dispersion and not as an agreement check.

\subsection{Reporting Discipline}
\label{app:reporting_discipline}

Five rules follow from the measurements above, and every group-level
statement in this paper obeys them.

\begin{enumerate}[leftmargin=1.4em,itemsep=0.25em]
\item \textbf{Claims travel with the grouping convention.} Any group-level
ordering or level statement names its cut, and an ordering that reverses
under recutting is reported as conditional on the convention rather than as
a property of the architectures.
\item \textbf{Levels are not comparable across cuts.} Paired flat and
structural FRR readouts rank consistently ($\rho_S=.79$) but carry a
systematic level offset (median $\Delta=+.082$, structural above flat), so
cross-cut comparisons use ranks or calibrated excess and never raw levels.
\item \textbf{Chance immunity is cut-dependent.} Large-group cuts make
$\theta{=}0.7$ essentially chance-free, while the flat protocol passes
about $25$ percentage points of bulk groups by chance. Raw $\theta$
fractions are quoted only for cuts whose null pass rate is near zero, and
calibrated excess is used otherwise.
\item \textbf{Time-axis readouts use $\mathrm{LRGF}_{\mathrm{cal}}$.} Where
a bare $\theta$ fraction is reported across layers, it is accompanied by
its $\mathrm{null}_{\theta}$ reference.
\item \textbf{Selectivity lift is not compared across predictors.} Lift is
a within-predictor quantity, since SESA is evaluated on the selector that
predictor produces.
\end{enumerate}

\section{Spatial Localization: the Auxiliary--Bulk Partition}
\label{app:spatial_localization}

\mainref{sec:heterogeneity} splits trainable tensors into two morphological
classes and reads the auxiliary--bulk separation within matched
trajectories. This appendix gives the membership rule, the tensor-by-tensor
attribution it produces in each training regime, and the audits that bound
how much the choice of rule could matter.

\subsection{The Partition Rule}
\label{app:partition_rule}

The primary definition is morphological. A parameter tensor is
\emph{auxiliary} if it is vector-like, which covers normalization scales
and shifts, biases, and token and positional embedding vectors, and
\emph{bulk} if it is a matrix-like weight, which covers linear and
attention projections, MLP weights, and convolutional kernels. The
definition refers to the shape and function of the tensor, not to its size.

The vision pipeline realizes this rule through a size threshold on coarse
role classes: a tensor is auxiliary if the total parameter count of its
role class within that model falls below $10^{5}$. The threshold applies to
role-class totals rather than to individual tensors, so a small individual
matrix inside a large class, such as the $1{,}280$-parameter ViT classifier
head, stays bulk. Every quantitative auxiliary--bulk readout in the paper
uses this operational assignment, and
Appendix~\ref{app:cifar_attribution} audits it against the morphological
definition. On Pythia the same rule is expressed through a segment taxonomy
rather than a threshold, because the segment structure already separates
matrices from vectors exactly (Appendix~\ref{app:pythia_partition}). The Pythia partition additionally coincides with optimizer treatment: normalization parameters and all biases are exempt from weight decay under the released recipe, so auxiliary--bulk contrasts at that scale register a parameter-role difference under the recipe rather than an isolated effect of tensor shape.

\subsection{Vision Panel: Attribution and Dual-Definition Audit}
\label{app:cifar_attribution}

Table~\ref{tab:appC_attribution} lists the attribution per model, the
composition of the auxiliary class by coarse role, and the audit of the two
definitions against each other.

\begin{table*}[tp]
\centering
\small

{\centering (a) Per-model attribution\par}
\vspace{0.3em}

\begin{tabular}{l rr rr r l}
\toprule
 & \multicolumn{2}{c}{Tensors} & \multicolumn{2}{c}{Parameters} & & \\
\cmidrule(lr){2-3}\cmidrule(lr){4-5}
Model & aux & bulk & aux & bulk & aux share & aux composition (tensors) \\
\midrule
MLP & 3 & 4 & 1{,}792 & 3{,}803{,}648 & 0.05\% & bias\,3 \\
ViT & 37 & 18 & 14{,}464 & 531{,}712 & 2.65\% & ln\,18, bias\,17, tok\,2 \\
ViT-mid & 61 & 30 & 32{,}704 & 2{,}822{,}848 & 1.15\% & ln\,30, bias\,29, tok\,2 \\
ResNet & 31 & 15 & 7{,}040 & 2{,}770{,}624 & 0.25\% & bn\,26, bias\,4, linear\,1 \\
AlexNet & 7 & 8 & 9{,}344 & 57{,}035{,}456 & 0.02\% & bias\,7 \\
\midrule
All & 139 & 75 & 65{,}344 & 66{,}964{,}288 & 0.10\% & --- \\
\bottomrule
\end{tabular}

\vspace{1.0em}
{\centering (b) Dual-definition audit\par}
\vspace{0.3em}

\begin{tabular}{l rrr r}
\toprule
Audit level & total & agree & defect & agreement \\
\midrule
model $\times$ tensor & 214 & 213 & 1 & 99.53\% \\
cell $\times$ tensor & 1{,}203 & 1{,}196 & 7 & 99.42\% \\
panel rows (5 anchors) & 6{,}015 & 5{,}980 & 35 & 99.42\% \\
\bottomrule
\end{tabular}

\caption{Auxiliary--bulk attribution and dual-definition audit on the
vision panel. \textbf{(a)} Per-model tensor attribution under the
operational split, with the composition of the auxiliary class by coarse
role (ln/bn/bias/tok; the single defector appears as \emph{linear}).
\textbf{(b)} Agreement between the morphological and operational
definitions at three audit levels; the complete defector list is the single
tensor discussed in Appendix~\ref{app:cifar_attribution}. Definitions are
given in Appendix~\ref{app:partition_rule}. Data: the tensor-level table of
the canonical group-level panel, with attribution fields constant across
anchors and optimizer cells. Parameter counts cover the parameter vector
only, with no batch-normalization running statistics.}
\label{tab:appC_attribution}
\end{table*}

The two definitions agree on $213$ of $214$ unique model--tensor pairs
($99.5\%$), and on $1{,}196$ of $1{,}203$ cell--tensor instances
($99.4\%$). The defector list is a single tensor: the ResNet final
classifier head \texttt{linear.weight} ($2{,}560$ parameters), which is
matrix-like under the morphological definition but auxiliary under the
operational one, because its \texttt{linear} role class totals only those
same $2{,}560$ parameters. It recurs in all seven ResNet optimizer cells,
which is $35$ of $6{,}015$ panel rows. Because the reported readouts use
the operational assignment, this tensor is counted as auxiliary throughout.

Two properties of the audit are worth stating explicitly. First, the
agreement is a property of the role-class form of the threshold: applying
the same $10^{5}$ threshold per tensor instead of per role class would
agree with the morphological definition on only $175$ of $214$ tensors
($81.8\%$). Second, the four ResNet tensors filed under the \texttt{bias}
bucket are the shortcut batch-normalization affine parameters, which the
coarse name rule places there because their names carry no \texttt{bn}
marker. They are vector-like under either reading, so the audit is
unaffected.

The panel covers the parameter vector only, with no batch-normalization
running statistics, and excludes tensors holding fewer than $16$ scalars,
because the $K{=}16$ grouping cannot be formed there. On the vision side
this excludes exactly one tensor per model, the ten-parameter classifier
bias, for $50$ parameters in total.

\subsection{Pythia: Segment Taxonomy and the Vector Segment}
\label{app:pythia_partition}

\begin{figure*}[tp]
  \centering
  \includegraphics[width=5in]{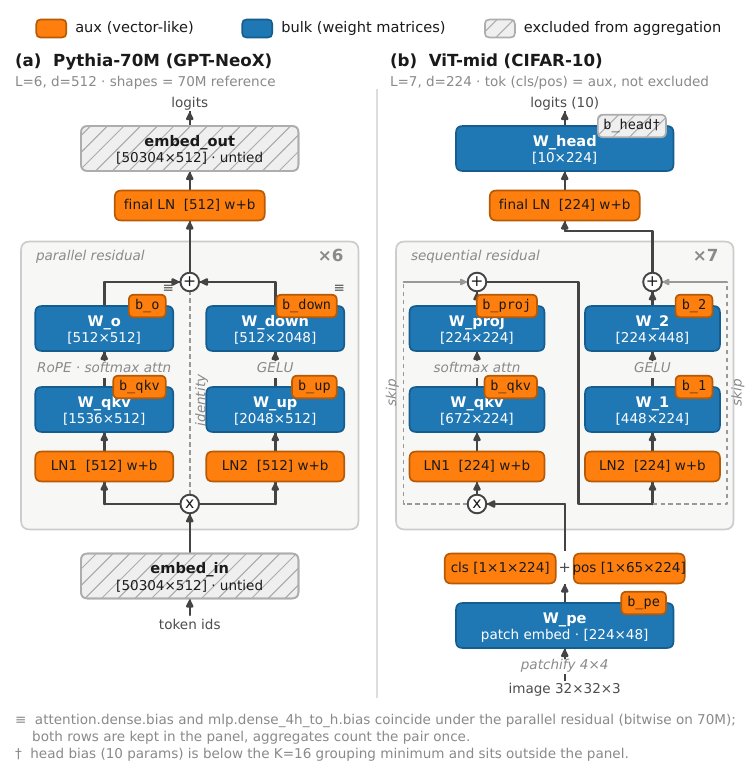}
  \caption{Where the auxiliary--bulk partition sits in the architecture.
  One box per parameter tensor (LayerNorm boxes carry weight and bias
  jointly; linear biases are drawn as chips on their matrix box), with fill
  colour giving the partition class: orange for auxiliary, which is the
  vector segment and covers all vector-like parameters including the final
  LayerNorm; blue for bulk, the weight matrices ($qkv$, \texttt{attn\_out},
  MLP); hatched grey for tensors excluded from the aggregation.
  \textbf{(a)} Pythia-70M: one parallel-residual block, repeated six times,
  plus the top level; the two untied embedding matrices are their own
  segments and are excluded. The marked bias pair ($\equiv$) coincides
  under the parallel residual, so both rows are kept in the panel while
  aggregates count the pair once. \textbf{(b)} ViT-mid: sequential
  residuals, repeated seven times; here the token and positional embedding
  vectors are vector-like and therefore auxiliary, which is the one
  cross-regime difference from (a), and the ten-parameter head bias
  ($\dagger$) is below the $K{=}16$ grouping minimum and is the only
  tensor excluded on the vision side. Shapes are the 70M reference.}
  \label{fig:appC_structure}
\end{figure*}

Table~\ref{tab:appC_pythia} lists the segment scheme with shapes, tensor
counts, and parameter shares at 70M, and Figure~\ref{fig:appC_structure}
places every parameter tensor in the architecture.

\begin{table*}[tp]
\centering
\small

\begin{tabular}{l l l l r r r}
\toprule
Segment & Class & Tensor patterns & Shape (70M) & $n$ & Params (70M) & Share \\
\midrule
$qkv$ & bulk & \texttt{attention.query\_key\_value.weight} & $1536\times512$ & 6 & 4{,}718{,}592 & 6.70\% \\
\texttt{attn\_out} & bulk & \texttt{attention.dense.weight} & $512\times512$ & 6 & 1{,}572{,}864 & 2.23\% \\
mlp & bulk & \texttt{mlp.dense\_h\_to\_4h.weight} & $2048\times512$ & 12 & 12{,}582{,}912 & 17.87\% \\
 & & \texttt{mlp.dense\_4h\_to\_h.weight} & $512\times2048$ & & & \\
vec & aux & \texttt{input\_layernorm.\{weight,bias\}} & $512$ & 50 & 40{,}960 & 0.058\% \\
 & & \texttt{post\_attention\_layernorm.\{weight,bias\}} & $512$ & & & \\
 & & \texttt{attention.query\_key\_value.bias} & $1536$ & & & \\
 & & \texttt{attention.dense.bias}\,$^{\equiv}$ & $512$ & & & \\
 & & \texttt{mlp.dense\_h\_to\_4h.bias} & $2048$ & & & \\
 & & \texttt{mlp.dense\_4h\_to\_h.bias}\,$^{\equiv}$ & $512$ & & & \\
 & & \texttt{final\_layer\_norm.\{weight,bias\}}\,$^{\ast}$ & $512$ & & & \\
\texttt{embed\_in} & excluded & \texttt{gpt\_neox.embed\_in.weight}\,$^{\ast}$ & $50304\times512$ & 1 & 25{,}755{,}648 & 36.57\% \\
\texttt{embed\_out} & excluded & \texttt{embed\_out.weight}\,$^{\ast}$ & $50304\times512$ & 1 & 25{,}755{,}648 & 36.57\% \\
\midrule
bulk & \multicolumn{3}{l}{$=qkv + \texttt{attn\_out} + \mathrm{mlp}$} & 24 & 18{,}874{,}368 & 26.80\% \\
aux & \multicolumn{3}{l}{$=$ vec (all vector-like parameters incl.\ final LN)} & 50 & 40{,}960 & 0.06\% \\
excluded & \multicolumn{3}{l}{$=\texttt{embed\_in} + \texttt{embed\_out}$ (untied)} & 2 & 51{,}511{,}296 & 73.14\% \\
total & & & & 76 & 70{,}426{,}624 & 100\% \\
\bottomrule
\end{tabular}

\vspace{0.5em}

{\footnotesize
Per-layer patterns carry the prefix \texttt{gpt\_neox.layers.$i$.} and
repeat over the $L=6$ layers; $^{\ast}$~marks whole-model tensors, with
\texttt{final\_layer\_norm} counted inside vec and the two embedding
matrices forming their own segments. Shapes are the 70M reference and scale
with model size (160M: $d{=}768$, $L{=}12$, $d_{\mathrm{mlp}}{=}3072$, so
the $qkv$ weights become $2304\times768$); shares are of the 70M parameter
total $70{,}426{,}624$. $^{\equiv}$~\texttt{attention.dense.bias} and
\texttt{mlp.dense\_4h\_to\_h.bias} coincide under the parallel residual, so
both rows are kept in the panel with a duplicate flag and all aggregates
count the pair once. The auxiliary series of the aggregation is an alias of
the vec segment, with a per-anchor difference identically $0.0$. The vec
segment is exempt from weight decay in the released pretraining recipe,
while the matrix segments and both embeddings take $\mathrm{wd}=0.1$.}

\caption{Pythia segment scheme and its mapping to the auxiliary--bulk
partition, with shapes and parameter counts at 70M.}
\label{tab:appC_pythia}
\end{table*}

Each transformer layer contributes three matrix segments, $qkv$
(\texttt{attention.query\_key\_value.weight}), \texttt{attn\_out}
(\texttt{attention.dense.weight}), and \texttt{mlp}
(\texttt{dense\_h\_to\_4h} and \texttt{dense\_4h\_to\_h} weights), which
together constitute bulk, and one vector segment collecting the two
per-layer LayerNorms with their weights and biases, all four linear biases,
and the final LayerNorm. The auxiliary series of the aggregation is an
alias of the vector segment, with a per-anchor difference identically
$0.0$, so the vector segment is the auxiliary class: all vector-like
parameters including the final LayerNorm. At 70M this is $50$ tensors and
$40{,}960$ parameters, against $24$ tensors and $18{,}874{,}368$ parameters
for bulk.

\paragraph{Embeddings.}
The two untied embedding matrices, \texttt{embed\_in} and
\texttt{embed\_out}, form their own segments and stay outside the
auxiliary--bulk aggregation. They hold $73.1\%$ of the parameter count, so
the aggregation covers the remaining $26.9\%$. One labeling imprecision is
registered: in the existing level--time aggregation the embedding rows
enter the bulk side with $0.016\%$ row weight, which has no numerical
effect on any reported readout.

\paragraph{The duplicate bias pair.}
Under the parallel residual of GPT-NeoX, \texttt{attention.dense.bias} and
\texttt{mlp.dense\_4h\_to\_h.bias} receive identical per-step gradients,
and their trajectories coincide: bitwise on 70M over all released
checkpoints, and within one half-precision unit in the last place on 160M.
Both rows are kept in the panel with a duplicate flag, and every aggregate
counts the pair once.

\paragraph{Weight decay.}
The vector segment is entirely exempt from weight decay in the released
pretraining configuration, while the matrix segments and both embeddings
take $\mathrm{wd}=0.1$. This is a property of the public training recipe
rather than a choice of ours, and it is one of the conditions under which
the auxiliary--bulk contrast in \mainref{fig:aux_bulk_slope} is read.

\subsection{Cross-Regime Comparability}
\label{app:partition_cross_regime}

The membership rule, vector-like to auxiliary, is the same in both regimes.
The one difference is which embedding-adjacent tensors exist and where they
land. On the vision side the token and positional embedding \emph{vectors}
are vector-like and therefore auxiliary. On Pythia the embedding
\emph{matrices} are not vector-like, and they are excluded as separate
segments rather than assigned to bulk.

Two further protocol differences keep absolute levels from being compared
across regimes, and \mainref{fig:aux_bulk_slope} accordingly compares the
within-line ordering rather than levels. The vision panel uses flat
$K{=}16$ groups on an epoch clock, whereas the Pythia panel uses
architecture-aligned segment groups on a released-checkpoint clock, and the
two clocks index different amounts of optimization per probe step.

\section{Pocket Trajectories, Persistence, and Overlap Registrations}
\label{app:pocket_persistence}

\mainref{sec:heterogeneity} reports that predictable pockets are a small
minority of bulk group--anchor observations, that most of them are
transient, and that the degree of persistence varies by parameter role.
This appendix gives the trajectories behind those statements: what the
pocket population looks like at a given time, when it appears and how
long individual pockets survive, the one event on the pretraining side at
which the population changes discontinuously, and the measured overlaps
between pocket membership and indicators defined outside the
subspace-residual family.

\paragraph{Units.}
As everywhere in this paper, each readout below names the unit it is read
on and the threshold convention it uses
(Appendix~\ref{app:reporting_discipline}). Two units appear. At
\emph{group level} a pocket is a functional group whose FRR falls below
the size-matched $5\%$ null quantile at that anchor ($\tau{=}5$, raw
protocol); this is the census unit, and it is read as excess over the
null. At \emph{row level}, meaning one $W_{qkv}$ row, one
$d_{\mathrm{head}}$ slice of the attention output projection, or one
output column of an MLP neuron, the reported quantity is the fraction of
rows with $\mathrm{FRR}<0.7$; at those group sizes the
membership-resample pass rate is near zero
(Appendix~\ref{app:lrgf_calibration}), so that fraction is read as an
absolute population. The two are not interchangeable and are never
pooled.

\paragraph{Reading discipline.}
The overlap subsection is written as a registry. Each entry is a measured
overlap between pocket membership and one indicator; the entries are not
treated as evidence for one another, and none of them supports a
statement about what pocket membership represents.

\subsection{Census and Residence}
\label{app:pocket_census}

\paragraph{Group-level census.}
The census covers $123$ anchors at $1$k-step spacing for both 70M and
160M. The population is stratified by segment and, within a segment,
concentrated in a few layers: averaging the bulk-segment share over
anchors, 70M runs from $0.056$ at layer 2 to $0.173$ at layer 4, and 160M
from $0.046$ at layer 5 to $0.155$ at layer 11, a factor of about three
between the quietest and the busiest layer at both sizes.

Stability follows the same stratification.
Table~\ref{tab:appD_jaccard} reports membership overlap between
consecutive anchors together with residence run lengths. Vector and $qkv$
pockets are close to persistent (vector Jaccard $1.000$ at $k{=}1$, with
residence median $7.5$ anchors at 70M and maximum $62$ at 160M; $qkv$
Jaccard $.875$ and $.842$, with residence median $5$ at 70M, $p_{95}$ of
$19$ to $20$, and maximum $56$), attention-output is intermediate
($.333$ and $.476$), and MLP pockets are strongly transient (Jaccard
$.141$ at both sizes, residence median $1$ anchor).

Counted over residence runs rather than over segments, the population is
dominated by the MLP: $103{,}465$ of $104{,}030$ runs at 70M and
$239{,}191$ of $240{,}808$ at 160M, more than $99\%$ at both sizes, and
those runs have median length one anchor. The transient majority reported
in \mainref{sec:heterogeneity} is therefore a statement about the pocket
population by count, and the segment-resolved view here is what the same
sentence means by persistence varying with parameter role.

\paragraph{Multi-pass recomputation.}
The same census recomputed on two vision runs with the same functional
units, over anchors $5$ to $45$ at $\tau{=}5$ epochs, moves in opposite
directions for the two architectures. On ViT-mid the MLP, vector, and
$qkv$ shares decay ($.233\to.104$, $.250\to.095$, $.219\to.182$ from the
early to the tail period) while the patch share rises steeply
($.194\to.487\to.791$). On ResNet both convolutional families rise
($3{\times}3$ $.143\to.381\to.472$, $1{\times}1$ $.106\to.162\to.261$)
while the vector share decays ($.283\to.100$). Adjacent-anchor Jaccard is
$.658$ for ViT-mid $qkv$, $.733$ to $.750$ for vector groups, $.480$ for
ViT-mid MLP, $.446$ for ResNet $3{\times}3$, and $.244$ for $1{\times}1$.
The frozen-coordinate fraction is zero throughout both runs, so no
reading in this census is carried by inactive units.

Two riders travel with the cross-regime comparison. The time unit
differs: one pretraining anchor spans about $0.8\%$ of training against
about $2\%$ for one vision epoch, so the sparser vision sampling
mechanically depresses its Jaccard values. The contrast that survives
that bias is the transient MLP population ($.141$ against ViT-mid MLP
$.480$), which runs against the direction of the bias; the $qkv$ contrast
is partly attributable to it. And a blanket reading that one-pass pockets
are more stable does not hold, because the stratification pattern itself
differs between the two regimes.

\paragraph{Dependence on the unit.}
Persistence is a property of the unit as much as of the run. Under a
top-$q$ \emph{group}-level agreement definition, the transfer family gives
cross-run Jaccard $0.115$ to $0.135$ against a chance level of $0.111$,
with a within-run adjacent-anchor ceiling of only $0.127$ to $0.156$: at
that unit the objects have almost no memory. Moving to \emph{tensor}-level
units on the same runs, at the same chance level, gives within-run
persistence $0.57$ to $0.73$ in every condition. On the tensor-level
definition the late-training pocket location is also conserved across
transfer regimes: the three continuing-descent arms converge to $0.57$ to
$0.71$ against a same-anchor cross-seed reference band of $0.58$ to
$0.80$, while at early anchors every cross-regime overlap sits at $0.23$
to $0.38$, far below the band, and the two anti-persistent arms stay
below it throughout. We register that as a time-resolved reading: early
pocket locations are regime-specific and late ones are not.

\subsection{Emergence and Survival}
\label{app:pocket_emergence}

\begin{figure*}[tp]
  \centering
  \includegraphics[width=\textwidth]{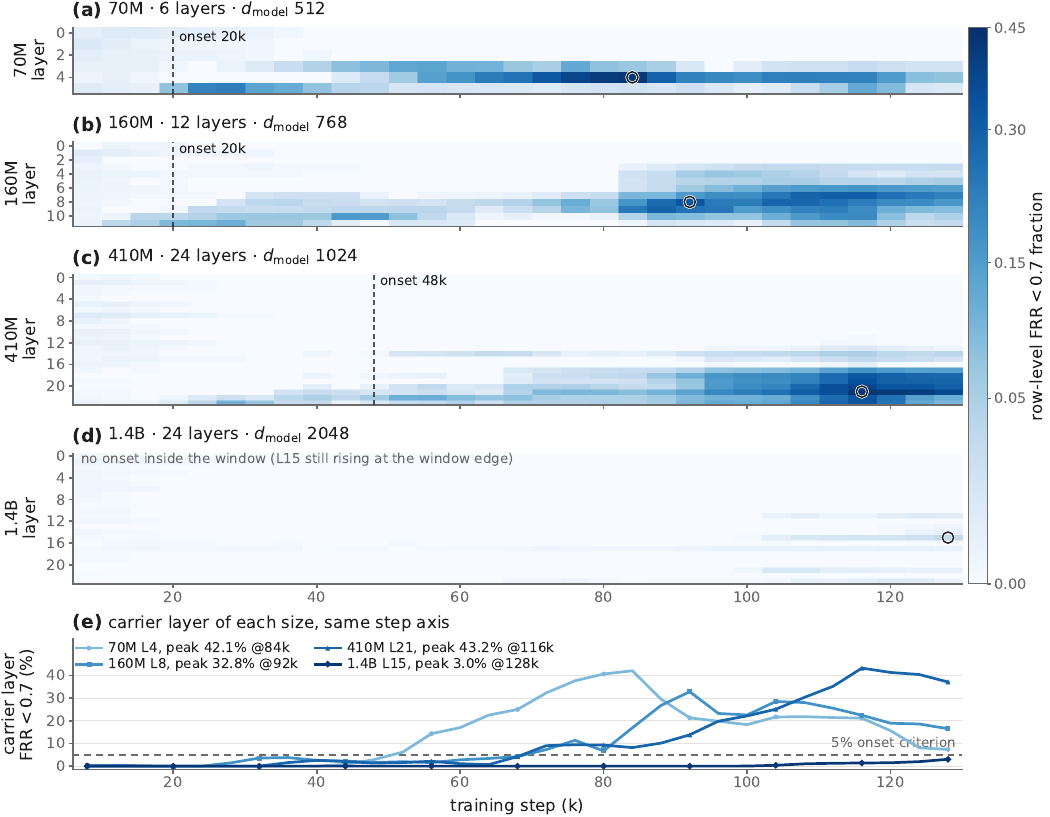}
  \caption{Row-level pocket emergence across four pretraining scales
  ($qkv$). Every panel shows the fraction of rows with $\mathrm{FRR}<0.7$
  resolved by depth and training step, where one row is one $W_{qkv}$ row
  ($D=d_{\mathrm{model}}$); at these group sizes the membership-resample
  pass rate at $\theta{=}0.7$ is near zero, so the fraction is read as an
  absolute population rather than as excess over a null. Panels (a) to (d)
  share one colour scale and one colour bar, so the four sizes are directly
  comparable; the pooled maximum is $0.4323$ (410M layer 21 at $116$k).
  Dashed vertical lines mark the row-level pocket onset, the earliest anchor
  at which some layer stays above $5\%$ for three consecutive anchors of the
  $4$k scan grid: $20$k at 70M and 160M, $48$k at 410M, and none inside the
  window at 1.4B, whose layer-15 population is still rising at the edge.
  Rings mark the maximum over layer and anchor. Panel (e) shows the carrier
  layer of each size on the same step axis with the $5\%$ criterion as a
  reference line. The 1B series is excluded from this figure on the grounds
  registered in Appendix~\ref{app:pythia_scale_extensions}, and its row-level
  fraction stays at or below $0.33\%$ throughout in any case. All entries are
  single-run readings on the public checkpoint series, registered as
  descriptions of those runs; no ordering between the row-level onset and the
  pool-coherence onset is asserted here.}
  \label{fig:appD_emergence}
\end{figure*}

\paragraph{Size-ordered timing.}
Table~\ref{tab:appD_timeline} collects the row-level emergence readings
for five sizes and Figure~\ref{fig:appD_emergence} shows the underlying
depth-by-time maps. Both the pool-coherence onset, which is where the
null line descends, and the row-level pocket onset move later with model
size, and the two do not move together. 70M has both at $20$k. 160M has a
two-stage null-line descent, moderate at $13$k and deep over about $72$
to $76$k, with pocket onset at $20$k. 410M has pocket onset at $48$k with
the null-line onset only at $92$k. Neither 1B nor 1.4B meets either
criterion inside the $129$k window. The ordering is not a function of
parameter count alone: at equal width $2048$ the deeper model starts
(1.4B, $24$ layers, rising from about $100$k) while the shallower one
does not (1B, $16$ layers), and at equal depth $24$ the narrower model
starts much earlier (410M at $48$k against 1.4B at the window edge).

\begin{table}[tbp]
\centering
\small

\begin{tabular}{@{}lccl@{}}
\toprule
 & \multicolumn{2}{c}{Pythia (one-pass)} & CIFAR (multi-pass) \\
\cmidrule(lr){2-3}
Segment & 70M & 160M & counterpart \\
\midrule
qkv       & $0.875$ & $0.842$ & $0.658$ (ViT-mid qkv) \\
vec       & $1.000$ & $1.000$ & $0.733$--$0.750$ (ViT-mid / ResNet vec) \\
mlp       & $\mathbf{0.141}$ & $\mathbf{0.141}$ & $\mathbf{0.480}$ (ViT-mid mlp) \\
attn\_out & $0.333$ & $0.476$ & --- \\
\bottomrule
\end{tabular}

\medskip

\begin{tabular}{@{}lrrrrrrrr@{}}
\toprule
 & \multicolumn{4}{c}{Pythia 70M} & \multicolumn{4}{c}{Pythia 160M} \\
\cmidrule(lr){2-5} \cmidrule(l){6-9}
Segment & p50 & p95 & max & $n$ & p50 & p95 & max & $n$ \\
\midrule
qkv       & $5$   & $19$   & $56$  & $175$      & $2$ & $20$ & $56$  & $612$ \\
vec       & $7.5$ & $25.5$ & $32$  & $14$       & $2$ & $20$ & $62$  & $50$ \\
mlp       & $1$   & $3$    & $123$ & $103\,465$ & $1$ & $3$  & $123$ & $239\,191$ \\
attn\_out & $1$   & $6$    & $40$  & $376$      & $1$ & $10$ & $30$  & $955$ \\
\bottomrule
\addlinespace[2pt]
\multicolumn{9}{@{}p{0.97\linewidth}@{}}{\footnotesize
Run lengths count consecutive anchors ($1$ anchor $=1000$ steps) that a group
spends inside the pocket; $n$ is the number of residence runs. Key readings:
vec p50 $7.5$ anchors (70M) with max $62$ (160M); qkv p50 $5$ (70M), p95
$19$--$20$, max $56$; mlp p50 $=1$ at both sizes.}\\
\end{tabular}

\caption{Pocket stability across training regimes. Top: Jaccard overlap of
pocket membership between consecutive anchors ($k{=}1$, median over
anchors), at the group-level unit defined at the start of
Appendix~\ref{app:pocket_persistence}. Bottom: residence run lengths on the
pretraining side, counting consecutive anchors a group spends inside the
pocket set, with $n$ the number of runs. Stability is stratified by segment
and the stratification pattern differs between the two regimes, so a blanket
ordering between them does not hold. At $k{=}1$ one pretraining anchor spans
about $0.8\%$ of training against about $2\%$ for one vision epoch, so the
sparser vision sampling mechanically depresses its values; the transient MLP
contrast runs against that bias and is the most robust entry. These overlaps
are registered as descriptions of pocket dynamics under the two regimes and
are not treated as cross-regime confirmation of a shared mechanism.}
\label{tab:appD_jaccard}
\end{table}

\begin{table*}[tp]
\centering
\small
\setlength{\tabcolsep}{3pt}

\begin{tabular}{@{}lccllcc@{}}
\toprule
& \multicolumn{4}{c}{qkv, row level ($D$ = one $W_{\mathrm{qkv}}$ row)}
& attn\_out & mlp (out-half) \\
\cmidrule(lr){2-5} \cmidrule(lr){6-6} \cmidrule(l){7-7}
Model & Null-line & Pocket & Main peak & Later wave & Peak & Peak \\
($L$ / $d_{\mathrm{model}}$) & onset & onset & (layer @ step) & (layer @ step)
& (layer @ step) & (layer @ step) \\
\midrule
70M ($6$ / $512$)
 & $20$k & $20$k & $\mathbf{42.1\%}$ (L4 @ $84$k)
 & $12.3\%$ (L5 @ $116$k) & $19.3\%$ (L4 @ $80$k) & $6.9\%$ (L5 @ $84$k) \\
160M ($12$ / $768$)
 & $13$k$^{\dagger}$ & $20$k & $\mathbf{32.8\%}$ (L8 @ $92$k)
 & $30.3\%$ (L7 @ $116$k) & $15.7\%$ (L8 @ $104$k) & $15.7\%$ (L11 @ $108$k) \\
410M ($24$ / $1024$)
 & $92$k & $48$k & $\mathbf{43.2\%}$ (L21 @ $116$k)
 & $32$--$38\%$ (L18--L20 @ $116$k) & not scanned & $5.2\%$ (L23 @ $116$k) \\
1B ($16$ / $2048$)
 & none in window & none & $0.3\%$ (L1 @ $8$k)
 & --- & not scanned & $1.8\%$ (L15 @ $8$k) \\
1.4B ($24$ / $2048$)
 & none in window & none$^{\ddagger}$ & $3.0\%$ (L15 @ $128$k)
 & --- & not scanned & $1.8\%$ (L23 @ $12$k) \\
\bottomrule
\addlinespace[2pt]
\multicolumn{7}{@{}p{0.97\linewidth}@{}}{\footnotesize
\textbf{Criteria.} Null-line onset: the pooled membership-resample
$q_{05}$ of qkv FRR (layer level, $\tau{=}5$, raw; median over layers)
stays below $0.90$ for five consecutive $1$k anchors. Pocket onset: some
layer's row-level $\mathrm{FRR}{<}0.7$ fraction stays above $5\%$ for
three consecutive anchors of the $4$k row-scan grid. Peak = maximum over
(layer, anchor) of the row-level $\mathrm{FRR}{<}0.7$ fraction; ``later
wave'' is the largest reading after the main peak's anchor. At these
group sizes the membership-resample pass rate at $\theta{=}0.7$ is
${\approx}0$, so the fractions are read as absolute rather than as excess
over a null. \textbf{Notes.} $^{\dagger}$160M descends in two stages
(moderate at $13$k, deep at ${\sim}72$--$76$k); the deep stage sits near
this run's smoothed loss minimum ($77.3$k). $^{\ddagger}$1.4B does not
meet the three-anchor criterion inside the window: its row-level fraction
starts rising at ${\sim}100$k in L11--L15/L21/L23 and reaches $3.0\%$ at
the window edge ($128$k) while still increasing, so this is a window-edge case rather
than a negative reading. 1B is read with the checkpoint-$116000$ masking
window applied (anchors whose $\tau{=}5$ window uses that checkpoint as
base or history, $116$--$121$k, plus the future endpoint $111$k; $7$ of
$123$ anchors dropped); its row-level fraction stays ${\leq}0.33\%$
throughout the remaining window. 1B is registered
here for completeness but is left out of the cross-scale trend
figures and of the cross-scale appendix: it is the one suite member
trained in bf16 rather than fp16, and at $16\times2048$ its depth
ordering crosses the parameter-count ordering, so it does not sit on
the depth/width axis those readings use. Released checkpoint dtype is
fp16 at every size, so the container precision our pipeline reads is
not special for 1B. attn\_out row scans exist for 70M/160M
only (410M/1B/1.4B requested); groups are flattened on
$(d_{\mathrm{model}}, d_{\mathrm{head}})$ with $d_{\mathrm{head}}$ as the
unit, a convention less clean than for qkv (70M $D{=}64$ gives $19.3\%$
vs.\ $13.1\%$ at $D{=}512$; 160M $15.7\%$ vs.\ $16.7\%$). The mlp column
reports the out-half ($w_{\mathrm{out}}$ columns) cut, the only mlp cut
carrying signal: in-half and magnitude cuts peak at ${\leq}1.6\%$ and
${\leq}0.12\%$ across all five sizes, and the mlp carrier is the deepest
layer in every size. The row-length convention is invariant over
$D{=}128..3072$, so no per-size adjustment is applied. Large-size mlp
readings are quoted from full-protocol scans only; the $10\%$-sampled
preview lane is not quoted as an S2 reading. All entries are single-run
readings on the public Pythia checkpoint series, registered as
descriptions of those runs.}\\
\end{tabular}

\caption{Row-level emergence readings across five pretraining scales, with
the onset criteria and the per-segment conventions given in the table
footer.}
\label{tab:appD_timeline}
\end{table*}

At 410M the row-level population starts at $48$k while the pool-coherence
onset is at $92$k, so the row-level population is not gated by the
pool-level onset. The two are separable events and no ordering between
them is asserted beyond what each run's readings show.

\paragraph{Layer relay within a run.}
At 70M the row-level population is a sequence of layer-localized waves
rather than a diffuse few percent: layer 5 rises first ($7.6\%$ at $20$k,
peaking at $22.1\%$ at $28$k, then decaying), layer 3 takes over from
$32$k with a double-humped trace ($13.2\%$ at $56$k and $12.3\%$ at
$76$k), the main layer-4 wave follows (peaking at $42.1\%$ at $84$k and
staying elevated over $50$ to $120$k), and a later wave arrives near
$115$k. At 160M the relay runs layer 8 (peak $32.8\%$ at $92$k) into
layer 7, which reaches $30.3\%$ at $116$k. At 410M the deep block, layers
18 to 21, rises together into a broad $110$ to $120$k band with the
maximum at layer 21 ($43.2\%$ at $116$k).

\begin{figure*}[tp]
  \centering
  \includegraphics[width=\textwidth]{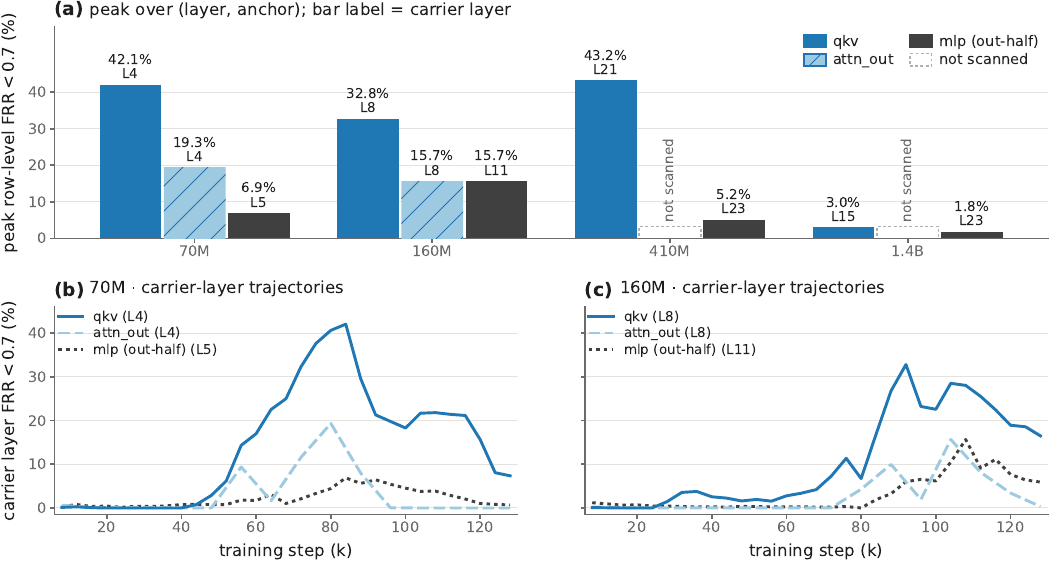}
  \caption{The same row-level population on three segments. (a) peak
  fraction of rows with $\mathrm{FRR}<0.7$, maximized over layer and anchor,
  for $qkv$, the attention output projection, and the MLP output half at
  four sizes, with the bar label giving the carrier layer; row scans of the
  attention output projection exist for 70M and 160M only, and the missing
  cells are drawn as explicit markers rather than as gaps. (b) and (c)
  trajectories of the carrier layer of each segment at 70M and 160M. Two
  conventions travel with the figure: the attention-output granularity is
  less clean than $qkv$'s, since groups are flattened on
  $(d_{\mathrm{model}}, d_{\mathrm{head}})$ with $d_{\mathrm{head}}$ as the
  unit and 70M reads $19.3\%$ at $D{=}64$ against $13.1\%$ at $D{=}512$; and
  group size differs between segments, so levels are compared within a
  segment across sizes rather than across segments. The 1B series is
  excluded on the grounds registered in
  Appendix~\ref{app:pythia_scale_extensions}.}
  \label{fig:appD_segments}
\end{figure*}

\paragraph{The attention block moves together.}
Figure~\ref{fig:appD_segments} places the three bulk segments on a common
scale. At head granularity the attention output projection shows no
pocket population at all (median FRR $0.993$ to $0.9996$ across sizes).
At row granularity it does: $19.3\%$ at 70M layer 4 at $80$k against
$qkv$'s $42.1\%$ at layer 4 at $84$k, and $15.7\%$ at 160M layer 8 at
$104$k against $qkv$'s $32.8\%$ at layer 8 at $92$k, about half the $qkv$
fraction in the same layer over the same period. The carrier layers of
the three segments agree closely within a size ($qkv$, attention output,
MLP: 70M 4/4/5; 160M 8/8/11), and their peak times separate only at 160M,
where $qkv$ leads at $92$k and the other two follow at $104$k and
$108$k. We register this at the level of the attention block: the
population appears on the read-in and the write-out projection together,
not on $qkv$ alone.

\paragraph{The MLP asymmetry.}
Re-cutting the MLP separates its two halves. Across all five sizes the
output half is the only cut carrying signal, with peaks from $1.8$ to
$15.7\%$, while the input half and the magnitude cut stay at or below
$1.6\%$ and $0.12\%$. This is a third independent negative reading for
the magnitude hypothesis on the MLP
(Appendix~\ref{app:cut_sensitivity}). The carrier is the deepest layer in
every size without exception. The late waves occur only in the models
that have crossed their transition ($6.9\%$ at $84$k for 70M, $15.7\%$ at
$108$k for 160M, $5.2\%$ at $116$k for 410M), whereas the 1B and 1.4B
maxima sit at $8$k and $12$k and are early-training residue.

\paragraph{Head-scale and row-scale do not coincide.}
Within a layer, the head-granularity calibrated readout fades before the
row-level population appears. Measured with a symmetric $\pm32$k window
around each layer's pocket onset, so that the search is not restricted to
the pre-onset side, the median lag from the calibrated trough to
row-level onset is $+20$k steps at 1.4B (four of four layers) and $+24$k
at 410M (eight of eleven layers earlier, three reversed). The horizontal
counterpart runs the other way and must be reported alongside: across
cells the Spearman correlation between the row-level fraction and the
calibrated readout is \emph{positive} ($+0.10$ overall, $+0.34$ late),
and the cells where the row-level population emerges have the same median
calibrated readout as the rest ($0.208$ against $0.208$). A low
calibrated readout therefore does not imply a pocket. What is registered
here is a within-layer temporal ordering, not a horizontal association.

\paragraph{Form of the emerging population.}
At 70M $qkv$ the coordinates carrying the coherent motion rotate
continuously: adjacent-anchor membership Jaccard of the comoving family
is $.14$ to $.18$, while cross-head timing sharing stays at $0.75$ to
$0.99$ throughout. Within a head the family condenses into two or three
row blocks, at four to eight times chance adjacency and $20$ times at
layer 5. In shallow and middle layers the per-head block directions are
private (cross-head direction cosine near chance at $.035$) and only the
timing is shared; at layer 5 both are shared (cross-head cosine $.82$,
effective rank $1$). The granularity ladder settles the unit, since the
one-row and quarter-row distributions nearly coincide, so within the tested ladder the coherent object is row-scale. Pooled over all layers the
model-wide row-level fraction is stable across training ($3.7\%$,
$4.8\%$, $4.4\%$ at $24$k, $60$k, $120$k), so the layer-localized waves
ride on a small standing population rather than replacing it.

\subsection{The One Discontinuous Event}
\label{app:pocket_collapse}

\begin{figure*}[tp]
  \centering
  \includegraphics[width=\textwidth]{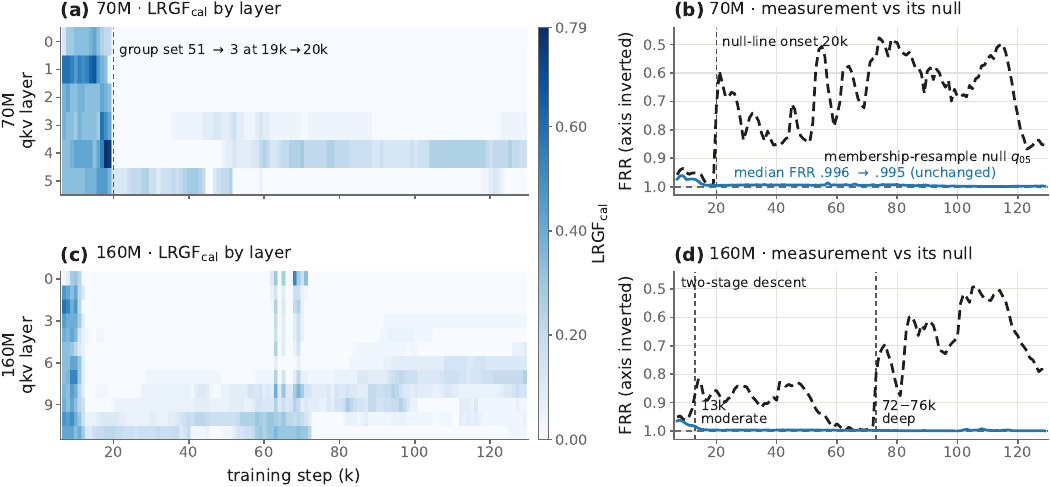}
  \caption{Where the ${\sim}20$k $qkv$ event sits, in the measurement or in
  its null. (a) and (c) layer-level calibrated readout for $qkv$
  ($\tau{=}5$, raw) at 70M and 160M on a common scale. (b) and (d) the same
  lane read as two curves: the measured median FRR of those layers (solid)
  and the pooled membership-resample $q_{05}$ (dashed), with the FRR axis
  inverted so that upward means lower FRR and the null pool merged across
  layers per segment and group size. At 70M the pocket set collapses from
  $51$ groups to $3$ between anchors $19$k and $20$k and the layer-level
  median calibrated readout falls from $.354$ to $.000$, while the measured
  median FRR of the same groups is unchanged ($.9959$ to $.9952$). What
  moves is the null, whose $q_{05}$ falls from $.9937$ to $.7201$ and stays
  at $.59$ to $.72$ through $25$k. The calibrated readout is a difference
  between a measurement and its null, so the event is located on the null
  side of that difference and is not described here as a change in the
  predictability of the groups. The 160M null line descends in two stages, a
  moderate descent at $13$k and a deep one over about $72$ to $76$k, the
  latter in the neighbourhood of that run's smoothed loss minimum ($77.3$k);
  the time co-location with the loss landmark is registered as a
  correspondence only.}
  \label{fig:appD_nullq05}
\end{figure*}

\paragraph{Where the ${\sim}20$k event sits.}
Between anchors $19$k and $20$k at 70M the $qkv$ pocket set collapses
from $51$ groups to $3$ and the layer-level median calibrated readout
drops from $.354$ to $.000$, while the measured median FRR of the same
groups does not move at all ($.9959$ to $.9952$, staying near $.995$
before and after). What moves is the null: the pooled
membership-resample $q_{05}$ falls from $.9937$ to $.7201$ and stays at
$.59$ to $.72$ through $25$k (Figure~\ref{fig:appD_nullq05}). The
calibrated readout is a difference between a measurement and its null, so
this event is located on the null side of that difference. This is what
lies behind the caution in \mainref{sec:dynamics} that the fading of
head-aligned excess should not be read as a global disappearance of
predictable structure.

\paragraph{Composition of the null pool.}
A single-anchor diagnostic at 70M at $25$k finds the null FRR
distribution long-tailed and bimodal ($q_{10}=.67$ with a jump to
$q_{25}=.992$), and across resampled groups the FRR tracks the share of
squared update energy carried by the top $1\%$ of coordinates at
$r=-.996$. The reading is that about $1\%$ of large-magnitude coordinates
move coherently across heads and dominate the resampled groups.

\begin{figure*}[tp]
  \centering
  \includegraphics[width=\textwidth]{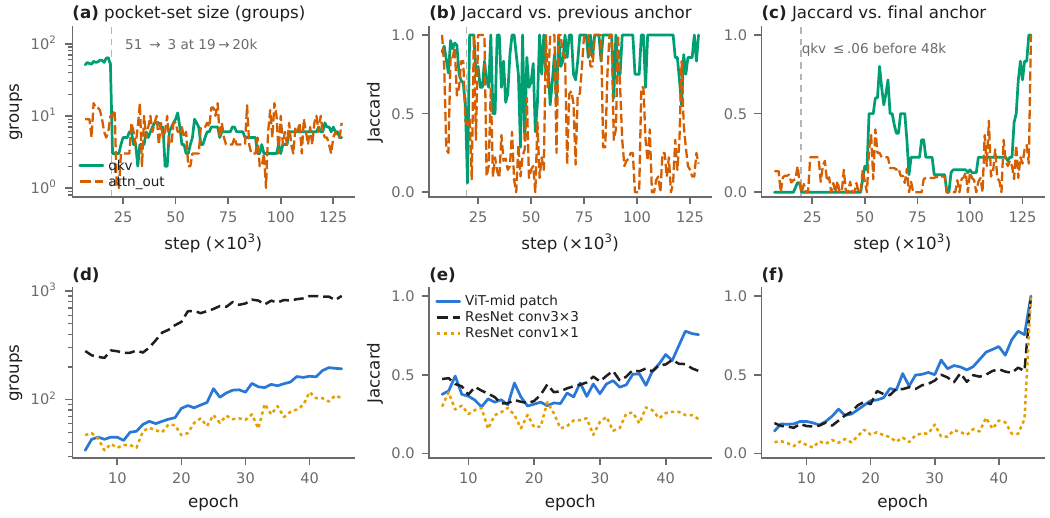}
  \caption{Turnover of the pocket set itself, at group level, for the
  one-pass 70M run (top row, dashed vertical at the $19$ to $20$k anchor
  pair) and the two multi-pass vision runs (bottom row). Columns: set size
  on a log scale, Jaccard against the previous anchor, and Jaccard against
  the final anchor. On $qkv$ the set collapses from $51$ groups to $3$ and
  then holds $2$ to $11$ groups, with Jaccard against the final set at or
  below $.06$ before $48$k, rising to $.5$ to $.8$ over $52$ to $70$k, while
  the adjacent-anchor Jaccard after the collapse has median $.88$. The
  attention output projection never stabilizes. The multi-pass runs show no
  event of this kind, with counts growing monotonically and the Jaccard
  against the final set rising steadily, which is cumulative expansion
  rather than replacement. This panel sets the reading conditions for the
  two-anchor movement overlaps: because the pocket set changes both size and
  membership between anchors, an anchor-to-anchor difference in overlap
  contains both a change in overlap and turnover of the set, and the two are
  separated only by reporting them side by side.}
  \label{fig:appD_turnover}
\end{figure*}

\paragraph{Turnover across the event.}
The set that survives is not a subset of the one that collapsed
(Figure~\ref{fig:appD_turnover}). Jaccard against the final set is near
zero before $48$k (maximum $.061$), rises to $.5$ to $.8$ over $52$ to
$70$k, and settles at $1$ in the tail, while the adjacent-anchor Jaccard
after the collapse has median $.88$: the surviving set is stable from
anchor to anchor but shares almost no membership with the pre-collapse
population. On the attention output projection the sets never stabilize
(adjacent-anchor median $.33$, versus-final median $.09$).

\paragraph{A co-located depth reading.}
An independent lane measures the effective rank of the $qkv$ update
column space, as participation ratio divided by output dimension, in
phase-diagram windows. In the ballistic window both $qkv$ and MLP read
$0.24$ to $0.44$ at every depth. In the floor windows the MLP stays at
$0.75$ to $0.77$ (70M) and $0.65$ to $0.85$ (160M) and shallow $qkv$ at
$0.59$ to $0.70$, while deep $qkv$ collapses, to $.011$ and $.007$ at 70M
layers 3 and 4 and to $.007$ to $.030$ at 160M layers 6 to 10. Those
depths overlap the layers that carry the persistent calibrated band after
the event. The two lanes read the same update sequence, so this is
registered as a co-located depth pattern between two readouts of one
record.

\paragraph{No multi-pass counterpart.}
The vision runs show no event of this kind. Pocket counts grow
monotonically (ViT-mid patch $34$ to $192$, ResNet $3{\times}3$ $280$ to
$903$) and the Jaccard against the final set rises steadily from about
$0.2$ to about $0.8$, which is cumulative expansion rather than
replacement; the $1{\times}1$ family is the exception, with a
versus-final median of $.12$.

\subsection{Overlap Registrations}
\label{app:pocket_overlap}

Table~\ref{tab:appD_overlap} is the registry. Every entry is a measured
overlap between pocket membership and an indicator that is not a probe of
the subspace-residual family.

\begin{table*}[tp]
\centering
\footnotesize
\setlength{\tabcolsep}{3pt}

\begin{tabular}{@{}p{0.135\linewidth}p{0.105\linewidth}p{0.085\linewidth}p{0.245\linewidth}p{0.145\linewidth}p{0.185\linewidth}@{}}
\toprule
Indicator pair & Paradigm side & Direction & Key readings & Null convention
& Caveat registered \\
\midrule

Movement top-$K$ $\times$ pocket coordinates
 & Pythia qkv (70M)
 & above the null band
 & precision $.195$ at a$70$k ($4.0\times$ null $.0486$) and $.312$ at
   a$129$k ($9.0\times$ null $.0347$); time-resolved ratio steps
   $0.98 \to 3.77$ across $19$k$\to$$20$k, then stays high (cumulative
   median $7.5\times$, recent-window median $9.6\times$, max $16.2\times$
   at $23$k); pocket coordinates sit at family movement quantile
   $p_{50}{=}.93$ at a$129$k
 & hypergeometric, equal cardinality ($K$ = number of pocket scalars),
   with $95\%$ interval
 & $S_i=\sum_t w_i(t)\Delta w_i(t)$ from checkpoint differences only;
   AdamW preconditioning not removed. Pocket sets change both membership
   and size between anchors, so a two-anchor comparison mixes overlap
   change with set turnover \\
\addlinespace[2pt]

 & Pythia attn\_out (70M), late anchors
 & below the null band on signed-top
 & precision $.093$ vs.\ null $.167$ at a$129$k; for anchors
   ${\geq}100$k the cumulative top ratio has median $0.34$ and the
   cumulative bottom ratio $1.90$, while both recent-window ratios sit
   at ${\approx}0.6$
 & same
 & small sets ($1$--$15$ groups) with heavy turnover (adjacent-anchor
   Jaccard median $.33$, Jaccard vs.\ final set median $.09$); the two
   time conventions disagree, and both are reported \\
\addlinespace[2pt]

 & CIFAR (ViT-mid, ResNet)
 & above the upper bound, small margin
 & ViT-mid magnitude-bin families $.142$--$.217$ vs.\ null $.125$
   (five families, both anchors); ResNet conv$3{\times}3$ $.304$/$.382$
   vs.\ null $.282$/$.367$ ($1.08\times$/$1.04\times$); patch-filter
   inside the band at both anchors
 & same
 & ViT-mid patch at a$45$: the cut is near saturation
   ($K/N{=}86\%$), which narrows the null interval mechanically; ResNet
   conv$1{\times}1$ falls below the lower bound at a$25$ \\
\midrule

LN outlier coordinates $\times$ per-coordinate dynamics
 & Pythia 70M ($13$ LN gains); ViT-mid control ($15$)
 & co-located
 & outlier vs.\ non-outlier median ESA higher in $12/13$ tensors
   (e.g.\ post5 $.992$ vs.\ $.854$); median coherence higher in $11/13$
   (e.g.\ in2 $.898$ vs.\ $.114$); $\rho_S(z,\text{coherence})$ positive
   in $11/13$ ($.78$--$.94$), negative for in0 ($-.37$) and post0
   ($-.48$); cross-layer outlier Jaccard concentrates on adjacent LN
   pairs (median $.023$, $p_{90}$ $.32$, max $.56$ for
   post3$\times$post4)
 & robust $z>3$ on $|\text{gain}|$ at the late anchor; a top-$1\%$
   control cut reproduces every direction
 & the final LN is saturated (ESA and coherence $=1.0$ on all $512$
   coordinates, no resolution). ViT-mid gain range is narrow
   ($|\text{gain}|\leq1.05$) and only $4/105$ cross-layer pairs are
   non-zero. Outlier-set formation times are non-monotone (final LN set
   size $49$ at $88$k, $9$ at $129$k) \\
\midrule

M8 low-partici\-pation heads $\times$ pocket membership
 & Pythia 70M qkv
 & co-located
 & $18$ heads fall below the pooled-bulk Tukey fence
   ($\mathrm{pr}_{\mathrm{norm}}<.9047$); $8$ of them are layer-4
   q-heads; the seven lowest ($\mathrm{pr}_{\mathrm{norm}}$
   $.262$--$.622$) are exactly the a$70$k pocket set
   $\{$L4:q\_h0,1,2,4,5,6,7$\}$
 & Tukey fence (pooled bulk $Q_1-1.5\,\mathrm{IQR}$) on
   dimension-normalised participation ratio
 & participation ratio and pocket flag are computed from the same
   update history, so this is registered as co-location of two readouts
   of one record, not as two independent measurements \\
\midrule

qkv column-space effective rank $\times$ calibrated-readout extinction
 & Pythia 70M, 160M
 & co-located in depth
 & floor window: 70M qkv L3/L4 participation ratio per dimension
   $.011$/$.007$ against $.59$--$.66$ in shallow qkv layers and
   $.75$--$.77$ in mlp; 160M hump-rise window L6--L10 $.007$--$.030$
   against $.65$--$.85$ in mlp. The same depths carry the persistent
   $\mathrm{LRGF_{cal}}$ band after the $20$k step
 & none (descriptive ratio; no resampling null defined for this
   quantity)
 & window boundaries are operational (ballistic $[1\mathrm{k},
   10\mathrm{k})$ plus phase-diagram floor windows), the two readouts
   share the same $\Delta W$ sequence, and fp16 quantisation noise on
   per-coordinate increments is registered for this lane \\
\midrule

Group-level FRR $\times$ ESA $\times$ VCS
 & CIFAR structural cuts ($3$ architectures $\times$ $7$ recipes
   $\times$ $5$ anchors)
 & co-elevated
 & $P(\mathrm{VCS}{>}0.4 \mid \mathrm{FRR}{<}0.7)=88.4\%$ against a base
   rate of $38.9\%$ (ViT-mid); $P(\mathrm{ESA}{>}0.7 \mid
   \mathrm{FRR}{<}0.7)$ lifts by $3.4\times$ (ViT-mid) to $23.8\times$
   (ResNet); all three thresholds are passed by $6.7\%$ / $3.3\%$ /
   $0.3\%$ of ViT-mid / ViT / ResNet groups
 & membership-resample null at large $D$: $\sigma_{\text{null}}$ is
   $.005$--$.024$ for ESA and $.011$--$.047$ for VCS, so all three
   thresholds are chance-immune
 & the three probes read the same $\Delta w$ history and are not
   independent instruments. Their time profiles separate: pooled ESA and
   VCS shares decay monotonically ($18.4\%$/$49.0\%$ at a$5$ to
   $0.9\%$/$6.1\%$ at a$40$) while the FRR share peaks at a$10$--a$20$
   ($3.7/10.0/10.0/5.9/3.4\%$); learning-rate schedule is not
   disentangled \\
\midrule

Loss-curve events $\times$ qkv transition
 & Pythia 70M, 160M
 & co-located in time
 & 70M: calibrated qkv readout fades at ${\sim}19.5$--$20$k (layer-level
   median $\mathrm{LRGF_{cal}}$ $.354\to.000$ across $19$k$\to$$20$k),
   smoothed loss minimum at $49.7$k, main row-level pocket wave at
   $80$--$85$k. 160M: smoothed loss minimum at $77.3$k, null-line deep
   descent at ${\sim}72$--$76$k
 & none (event co-location on the time axis)
 & one run per size; loss minima are read from smoothed curves with a
   rule-dependent resolution, and the event windows are wide relative to
   the anchor spacing. Detail in the main-text section on loss
   co-location \\
\bottomrule
\addlinespace[2pt]
\multicolumn{6}{@{}p{0.97\linewidth}@{}}{\footnotesize
Every row is a registration of a measured overlap between pocket
membership and an indicator defined outside the FRR probe family. The
rows are not treated as evidence for one another, and no row is used to
support a statement about what pocket membership represents. ``Above /
below the null band'' refers to the stated null only. Pythia pocket sets
are taken from the S2 panel group-level flag ($\tau{=}5$, raw, head
granularity, same cut and same $\alpha$); two cross-checks against the
independent census agree (qkv at a$70$k: $7$ groups, all layer-4
q-heads; attn\_out at a$70$k: $15$ of $48$ heads, matching the
calibrated share $.313\times48$).}\\
\end{tabular}

\caption{Registry of measured overlaps between pocket membership and
indicators defined outside the subspace-residual probe family. Each row
records the direction, the key readings, the null convention, and the
caveats that travel with it. The rows are not treated as evidence for one
another, and none of them supports a statement about what pocket membership
represents.}
\label{tab:appD_overlap}
\end{table*}

\begin{figure*}[tp]
  \centering
  \includegraphics[width=\textwidth]{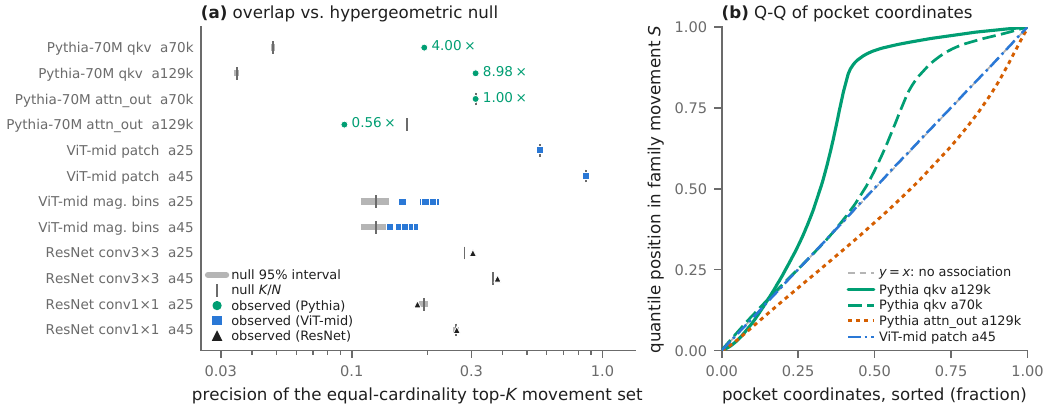}
  \caption{Overlap between pocket coordinates and the top-$K$ set of the signed norm-change proxy, labelled movement in the panels, registered against an equal-cardinality hypergeometric null.
  The proxy $S_i=\sum_t w_i(t)\dw_i(t)$ is built from checkpoint
  differences alone; preconditioning is not removed and $S$ is not a
  curvature-based importance measure. (a) observed precision of the top-$K$
  movement set restricted to pocket coordinates, against the null $K/N$ and
  its $95\%$ interval, with the pretraining rows annotated by the ratio.
  (b) quantile position of the pocket coordinates inside their family's
  movement distribution, with the diagonal as the no-association reference;
  the $qkv$ set at $129$k sits at median quantile $.93$ against about $.50$
  for the vision reference, so the whole set is displaced rather than a tail
  of it. Pocket sets are the group-level flag at $\tau{=}5$, raw, head
  granularity. The two anchors carry different pocket sets in both size and
  membership, so a two-anchor comparison mixes a change in overlap with
  turnover of the set itself. Each reading is a registration of a measured
  overlap and is not used to support any statement about what pocket
  membership represents.}
  \label{fig:appD_o1}
\end{figure*}

\paragraph{Signed norm-change proxy.}
A signed norm-change proxy $S_i=\sum_t w_i(t)\dw_i(t)$ is built from checkpoint differences alone; preconditioning is not removed, and $S$ is not a
curvature-based importance measure. Against an equal-cardinality
hypergeometric null, $qkv$ pocket coordinates overlap the top-$K$
movement set well above the null band ($4.0\times$ at $70$k, $9.0\times$
at $129$k), the ViT-mid magnitude-bin families sit just above their upper
bound ($.142$ to $.217$ against $.125$), ResNet $3{\times}3$ exceeds its
bound by only $1.04$ to $1.08\times$, and $1{\times}1$ falls below it
(Figure~\ref{fig:appD_o1}). The attention output projection reverses at
late anchors, with its terminal pocket set displaced towards the low end
of the signed movement axis.

\begin{figure*}[tp]
  \centering
  \includegraphics[width=\textwidth]{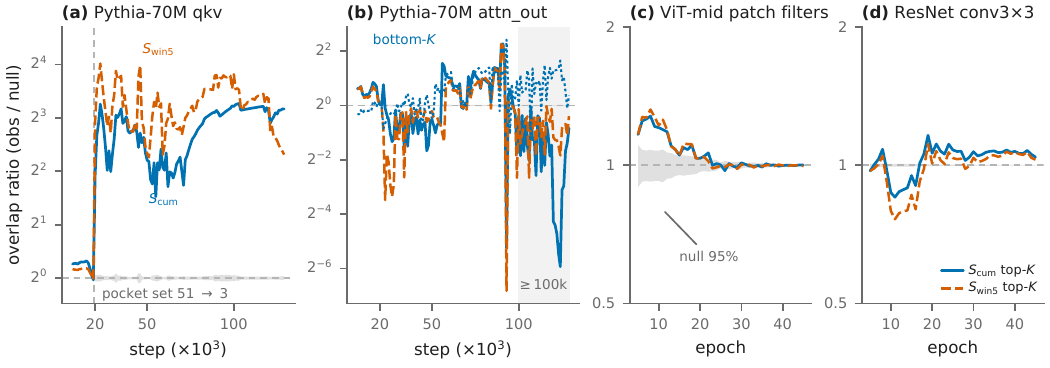}
  \caption{The same overlap resolved anchor by anchor, with the pocket set
  re-matched at every anchor. The readout is the ratio of the observed
  overlap to its hypergeometric expectation, with the grey band giving the
  null $95\%$ interval. Two time conventions for the movement proxy are
  reported side by side: one accumulates from the start of training, the
  other over the last five checkpoint or epoch pairs, matching the probe's
  history window. (a) $qkv$: the ratio hugs the random line over $7$ to
  $19$k, steps to $3.77$ at $20$k in one anchor, and stays high for the rest
  of training. (b) the attention output projection, on small sets with heavy
  turnover, where the two time conventions disagree and both are reported.
  (c) ViT-mid patch filters sit slightly above the band early and inside it
  from epoch $23$ apart from single-epoch excursions; the late approach to
  one has a mechanical component, since the cut approaches saturation.
  (d) ResNet $3{\times}3$ sits below the band over epochs $9$ to $16$ and
  above it from epoch $17$. Every curve is a registration of a measured
  overlap over time.}
  \label{fig:appD_o3time}
\end{figure*}

Because the pocket sets themselves change size and membership over time,
the overlap is also measured anchor by anchor
(Figure~\ref{fig:appD_o3time}). On $qkv$ the ratio hugs the random line
before the collapse ($0.98$ to $1.25$ over $7$ to $19$k, when the set
holds $51$ to $64$ groups) and steps to $3.77$ at $20$k, staying high for
the rest of training (cumulative median $7.5\times$, recent-window median
$9.6\times$ with a maximum of $16.2\times$ at $23$k). ViT-mid patch
filters sit slightly above the band early and inside it from epoch $23$
apart from single-epoch excursions; ResNet $3{\times}3$ sits
\emph{below} the band over epochs $9$ to $16$ and above it from epoch
$17$.

\begin{figure*}[tp]
  \centering
  \includegraphics[width=\textwidth]{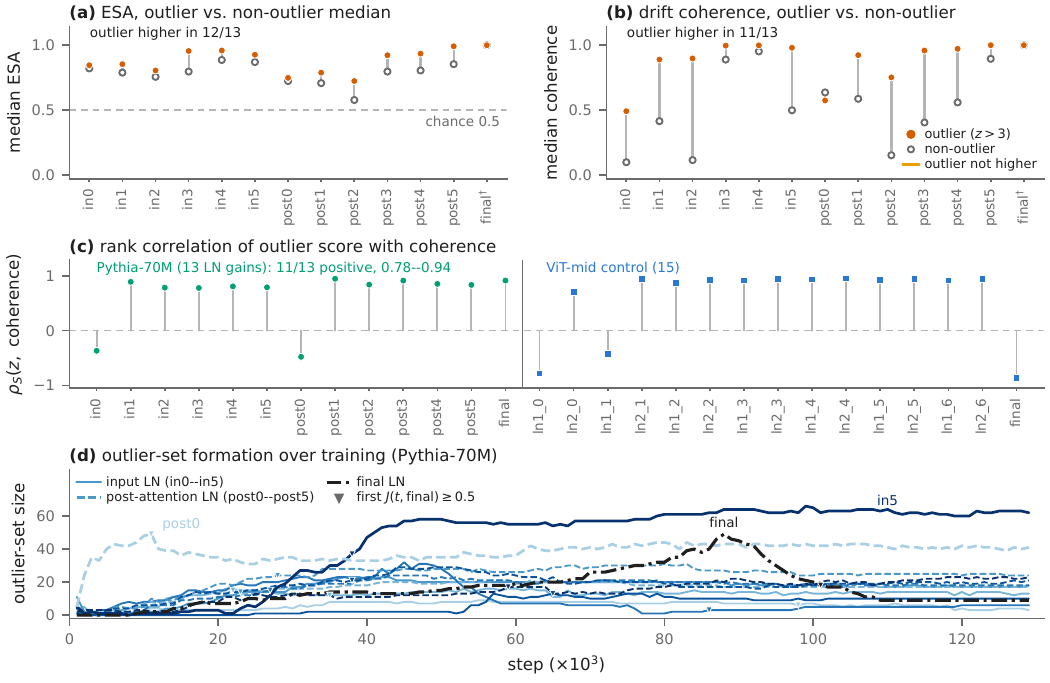}
  \caption{Normalization gain outliers and per-coordinate dynamics on the
  same gain trajectories. A coordinate is an outlier when the robust $z$ of
  its absolute gain at the late anchor exceeds $3$. (a) and (b) per-tensor
  medians for outlier against non-outlier coordinates across the $13$
  pretraining gain vectors: median ESA is higher for the outliers in $12$ of
  $13$ tensors and median drift coherence in $11$ of $13$, with connectors
  marking the tensors where it is not. The final normalization tensor is
  saturated, with agreement and coherence equal to $1.0$ on all $512$
  coordinates, so it carries no resolution and is drawn only for
  completeness. (c) rank correlation between the outlier score and coherence
  per tensor, positive in $11$ of $13$ pretraining tensors and negative in
  two. On the ViT-mid control one chain reads with the same sign and the
  others with the opposite sign; its gain range is narrow and its outlier
  sets hold at most four coordinates. A top-$1\%$ control cut reproduces
  every direction in both regimes.}
  \label{fig:appD_o2}
\end{figure*}

\paragraph{Normalization gain outliers.}
Coordinates whose normalization gain is a robust-$z$ outlier above $3$ at
the late anchor have higher median ESA than their non-outlier neighbours
in $12$ of $13$ pretraining tensors and higher median coherence in $11$
of $13$, with the rank correlation between the outlier score and
coherence between $.78$ and $.94$ in those $11$ and negative in the two
exceptions (Figure~\ref{fig:appD_o2}). A top-$1\%$ control cut reproduces
every direction including the exceptions. Outlier sets overlap across
layers mainly between adjacent normalization pairs (median Jaccard
$.023$, $p_{90}$ $.32$). Their formation times are not monotone: sets
grow from early training and first reach half of their final membership
between $11$k and $101$k depending on the tensor. On the ViT-mid control
the gain range is narrow and the outlier sets hold at most four
coordinates, and two of its three chains read with the opposite sign. The
final normalization tensor is saturated, with agreement and coherence
both at $1.0$ on all $512$ coordinates, and carries no resolution.

\paragraph{Low-participation heads.}
Eighteen $qkv$ heads fall below the pooled-bulk fence on
dimension-normalized participation ratio; eight of them are 70M layer-4
q-heads, and the seven lowest are exactly the $70$k pocket set.
Participation ratio and pocket flag are computed from the same update
history, so this is registered as a co-location of two readouts of one
record.

\paragraph{Cross-probe overlap at group level.}
On the vision structural cuts the three group-level probes concentrate on
overlapping populations. Conditioning on $\mathrm{FRR}<0.7$ raises the
$\mathrm{VCS}>0.4$ rate from a base of $38.9\%$ to $88.4\%$ on ViT-mid,
and raises the $\mathrm{ESA}>0.7$ rate by $3.4\times$ on ViT-mid to
$23.8\times$ on ResNet; all three thresholds are passed simultaneously by
$6.7\%$, $3.3\%$, and $0.3\%$ of ViT-mid, ViT, and ResNet groups. At
these group sizes the thresholds are chance-immune. The three probes read
the same update history and are not independent instruments, and their
time profiles separate: pooled ESA and VCS shares decay monotonically
from $18.4\%$ and $49.0\%$ at anchor $5$ to $0.9\%$ and $6.1\%$ at anchor
$40$, while the FRR share peaks at anchors $10$ to $20$. The
learning-rate schedule is not disentangled from these profiles.

\paragraph{Loss-curve landmarks.}
The transition markers co-locate in time with loss-curve landmarks at
both sizes (70M: calibrated fade near $20$k, smoothed loss minimum
$49.7$k, main row-level wave $80$ to $85$k; 160M: loss minimum $77.3$k,
deep null-line descent about $72$ to $76$k). This is one run per size,
the loss minima are read from smoothed curves with a rule-dependent
resolution, and the event windows are wide relative to the anchor
spacing. The co-location is discussed in \mainref{sec:dynamics} and is
registered here only for completeness.

\section{Movement Directedness and Gradient-Coherence Correspondence}
\label{app:state_correspondence}

\mainref{sec:dynamics} compares the probe readings with two quantities
that are defined without reference to any probe, and reports the
correlations $\rho=0.978$ and $\rho=0.804$. This appendix gives the two
definitions, their relation to existing gradient-coherence and
signal-to-noise measures, and the comparison in full.

\subsection{Two Externally Defined Quantities}
\label{app:state_definitions}

\paragraph{Movement directedness.}
For an anchor $a$ and horizon $\tau$, movement directedness is the net
displacement of the future window divided by the path length actually
travelled inside it,
\begin{equation}
\mathrm{MD}_S(a,\tau)
=\frac{\bigl\lVert w_{a+\tau,S}-w_{a,S}\bigr\rVert_2}
{\sum_{t=a}^{a+\tau-1}\bigl\lVert\dw_{t,S}\bigr\rVert_2}.
\label{eq:netpath}
\end{equation}
It lies in $[0,1]$, equals one when the window is traversed in a straight
line, and falls towards zero when the steps cancel. It is a property of
the realized trajectory alone: no probe, no history window, and no model
of the dynamics enters it. It is also not causally prior to the probes,
since both read the same stored weights; the comparison below is a
correspondence between two summaries of one record.

\paragraph{Loss-side gradient coherence.}
The second quantity is built only from scalar training telemetry. Over
the same window it compares the realized loss decrease with the loss
decrease that the same gradient magnitudes would have produced had every
step been perfectly aligned,
\begin{equation}
\mathrm{GC}(a,\tau)
=\frac{L_{a-1}-L_{a+\tau-1}}
{\bigl(\sum_{s\in W}\eta_s\bigr)\;
\bigl\langle\lVert g_s\rVert_2^{2}\bigr\rangle_{s\in W}},
\label{eq:losssnr}
\end{equation}
where $W$ is the set of optimization steps inside the window, $\eta_s$ is
the learning rate at step $s$, and $g_s$ is the gradient restricted to the
bulk backbone. The numerator is read from the loss curve and the
denominator from recorded per-step learning rates and gradient norms, so
the estimate contains no parameter-direction information at all. This is
what makes it a non-circular comparison for a directional probe.

\subsection{Relation to Existing Coherence and Noise Measures}
\label{app:state_related}

Both quantities have close relatives in the literature, and the
differences matter for how the comparison should be read.

Gradient coherence in the sense of \citet{chatterjee2020coherent} is
defined on \emph{gradients}: it measures how much per-example gradients
agree with one another within a batch, and it requires access to those
per-example gradients. Equation~\eqref{eq:losssnr} instead reads coherence
off the loss curve, as the fraction of the available first-order decrease
that the optimizer actually realized, and needs only telemetry that a
training run already logs. It therefore measures agreement \emph{across
steps} within a window rather than across examples within a step, and the
two can come apart whenever the loss surface curves inside the window.

Movement directedness is a trajectory-level quantity of the kind studied
by \citet{singh2025directionality}, and it is related to, but not the
same as, the observation that optimization concentrates in a
low-dimensional subspace \citep{gur2018gradient}. A trajectory can stay
inside a small subspace while doubling back inside it, which lowers
Eq.~\eqref{eq:netpath} without changing the subspace, and this is exactly
the distinction that separates the displacement-direction family from the
subspace-residual family in \mainref{sec:methods}. Values of
Eq.~\eqref{eq:netpath} well below one over short windows are also the
signature of the diffusion-like regimes described by
\citet{mandt2017stochastic} and \citet{feng2021phases}.

None of these quantities is used here to certify another. They are
registered as independent summaries of the same stored trajectories.

\subsection{The Full Comparison}
\label{app:state_full}

Figure~\ref{fig:appE_state} shows both comparisons over the transfer
family of Appendix~\ref{app:cifar_families}, on the bulk backbone scope at
$\tau=5$: eight training conditions, five anchors each, over the seeds of
that family. The conditions differ in dataset, in whether the run is
fine-tuned or trained from scratch, in learning rate, and in schedule, so
the grid deliberately mixes regimes that a single training curve would
not connect.

\begin{figure*}[tp]
  \centering
  \includegraphics[width=0.94\textwidth]{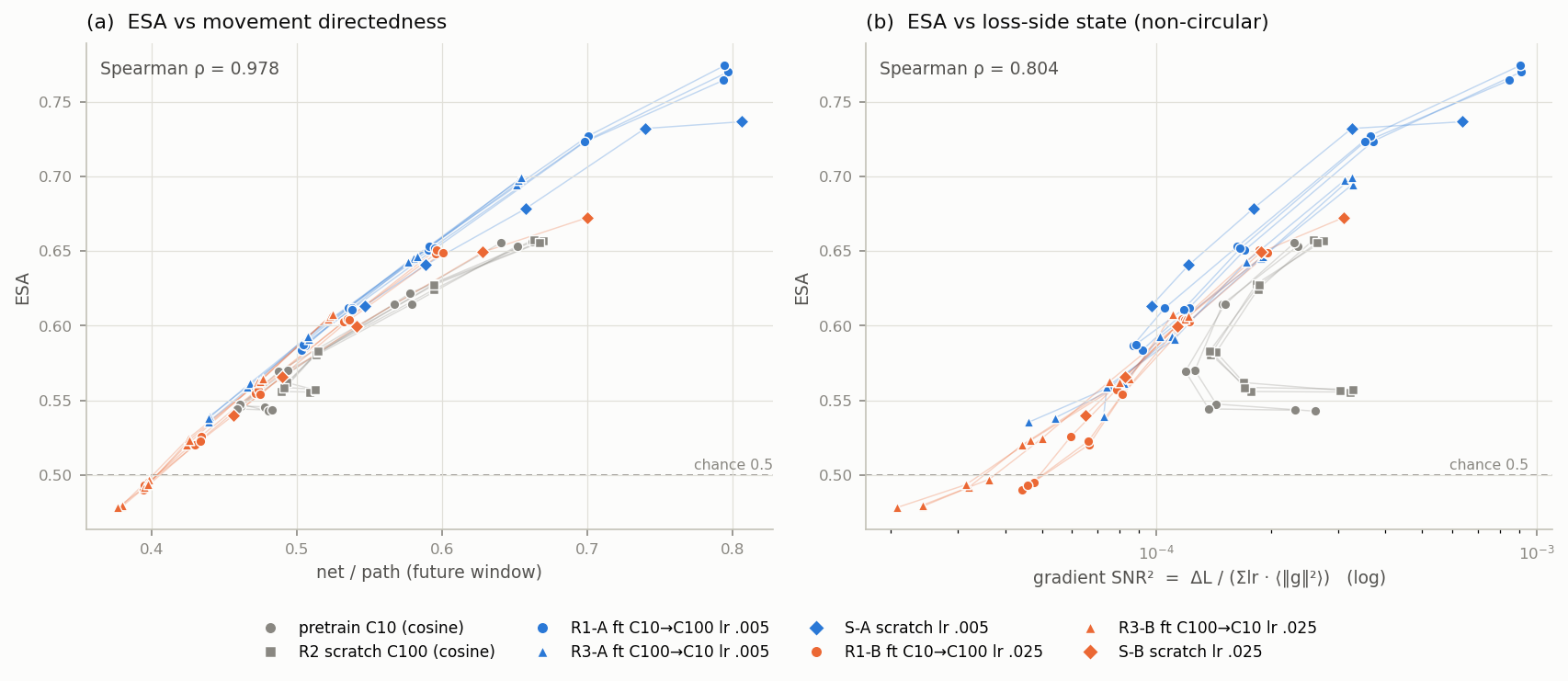}
  \caption{Probe readings against two externally defined summaries of
  realized optimization, over the eight conditions and five anchors of the
  transfer family, on the bulk backbone scope at $\tau{=}5$. (a) ESA
  against movement directedness, Eq.~\eqref{eq:netpath}, the net
  displacement of the future window divided by the path length travelled
  inside it. (b) ESA against the loss-side gradient-coherence estimate,
  Eq.~\eqref{eq:losssnr}, which is built from the loss curve, the per-step
  learning rates, and the per-step gradient norms, and carries no
  parameter-direction information. Horizontal lines mark the random
  sign-agreement level of $0.5$. Conditions differ in dataset, in whether
  the run is fine-tuned or trained from scratch, in learning rate, and in
  schedule. Both relations are retrospective correspondences between
  summaries of the same stored trajectories.}
  \label{fig:appE_state}
\end{figure*}

Across that grid ESA forms a tight monotone relation with movement
directedness ($\rho=0.978$). Trajectories whose future window covers more
net distance relative to its path length retain more coordinate-wise
directional agreement, and conditions that sit far apart in loss, learning
rate, and dataset fall on one curve. The association with the loss-side
coherence estimate is weaker but still strong ($\rho=0.804$), which is
what should be expected of a quantity that discards all directional
information and keeps only the scalar bookkeeping.

Two readings follow. The probes respond to the realized geometry of
parameter motion rather than to checkpoint index, since the grid contains
conditions at very different points of very different schedules. And the
sub-$0.5$ regimes that \mainref{sec:dynamics} reports in the pretraining
case study correspond to low directedness, that is, to windows whose steps
substantially cancel, rather than to an absence of measurable structure.

\paragraph{Scope.}
Both relations are retrospective and observational. The comparison is on
the vision side only, at one horizon and one scope, and the two quantities
are computed from the same stored trajectories as the probes, so a high
correlation registers consistency between summaries and not an independent
confirmation. The loss-side estimate additionally inherits every
confound of the loss curve it is read from, including schedule shape and
batch composition.

\section{Pythia Scale Extensions}
\label{app:pythia_scale_extensions}

\mainref{sec:dynamics} uses Pythia-70M as its case study and defers the
rest to this appendix. Six sizes enter the cross-scale registry, and every
statement below is a registration of readings and co-occurrences rather
than a directional claim.

\subsection{Coverage and Protocol}
\label{app:crossscale_coverage}

The six sizes are 14M, 31M, 70M, 160M, 410M, and 1.4B. They share the
pretraining schedule that makes step numbers comparable: $143{,}000$
optimizer steps, a single cosine decay whose decay length equals the
training length so that there is no constant tail, a $1\%$ warmup of
$1{,}430$ steps, a floor at $10\%$ of the peak rate, decoupled weight decay
$0.1$ with normalization parameters and all biases exempt, and a global
batch of $1024$ sequences of $2048$ tokens, or $2{,}097{,}152$ tokens per
step. Peak learning rate is the one recipe quantity that varies with size,
from $1.0\times10^{-3}$ at the three smallest to $2.0\times10^{-4}$ at
1.4B. Equal step therefore means equal token count, and the step-indexed
comparisons below are token-indexed comparisons as well. They are not
matched loss, matched compute, or matched training progress, and they
carry no further normalization.

The measurement protocol is identical at every size and is the one given
in Appendix~\ref{app:pythia_coverage}: $123$ anchors at steps $7000$ to
$129000$, $\npast=5$, reference $\dw_{a-1}$, Gavish--Donoho rank selection
on a $256$-unit sample, raw and column-normalized FRR in parallel, a
per-anchor membership-resample null, and the calibrated readout at
$\alpha=0.05$ reported as excess. Table~\ref{tab:appF_readings} collects
the per-size readings.

\begin{table*}[tp]
\centering
\small
\setlength{\tabcolsep}{4pt}

{\centering (a) Sizes, coverage, and external landmarks\par}
\vspace{0.3em}

\begin{tabular}{l rr c c l p{1.55in}}
\toprule
 & \multicolumn{2}{c}{Parameters} & & S2 & Train-loss minimum & Bulk ESA landmark \\
\cmidrule(lr){2-3}
Model & total & non-embed. & $L \times d_{\mathrm{model}}$ & anchors & in $[7\mathrm{k},129\mathrm{k}]$ & ($\tau{=}5$, group $p_{50}$) \\
\midrule
Pythia-14M  & 14{,}067{,}712    & 1{,}189{,}888     & $6 \times 128$   & 123 & no public telemetry & --- \\
Pythia-31M  & 30{,}494{,}720    & 4{,}739{,}072     & $6 \times 256$   & 123 & no public telemetry & --- \\
Pythia-70M  & 70{,}426{,}624    & 18{,}915{,}328    & $6 \times 512$   & 123 & step 50{,}345 ($2.828$) & min $0.426$ @ $98$k; local rise peak $0.467$ @ $77$k \\
Pythia-160M & 162{,}322{,}944   & 85{,}056{,}000    & $12 \times 768$  & 123 & step 76{,}409 ($2.465$) & min $0.424$ @ $68$k; sustained rise from $78$k to $0.45$--$0.47$ \\
Pythia-410M & 405{,}334{,}016   & 302{,}311{,}424   & $24 \times 1024$ & 123 & none in window$^{a}$ & late rise from ${\approx}85$k, peak ${\approx}0.49$ @ $110$--$118$k \\
Pythia-1.4B & 1{,}414{,}647{,}808 & 1{,}208{,}602{,}624 & $24 \times 2048$ & 123 & no verified telemetry$^{b}$ & --- \\
\bottomrule
\end{tabular}

\smallskip
{\footnotesize
Anchors are steps $7000$ to $129000$ at spacing $1000$, identical for every
size. Loss minima are recomputed from the public telemetry with a $\pm1500$-step
rolling median. $^{a}$\,The smoothed 410M loss is still descending at the right
edge of the anchor domain (step $128{,}756$, $2.175$) and reaches $2.173$ at
step $143$k, so no bottoming event occurs inside the window.
$^{b}$\,The local 1.4B telemetry file does not match the released $143$k-step cosine schedule, so no loss landmark is quoted.
Empty cells mean not read out on the full protocol; they are deliberately not
filled from the sampled-preview instrumentation.}

\vspace{1.0em}
{\centering (b) Internal transition readings\par}
\vspace{0.3em}

\begin{tabular}{l l l l l l}
\toprule
 & qkv $\mathrm{LRGF}_{\mathrm{cal}}$ & Null-line & Row pocket & Row pocket peak & Persistent qkv \\
Model & collapse & onset & onset & (frac.\ @ layer/step) & $\mathrm{LRGF}_{\mathrm{cal}}$ band \\
\midrule
Pythia-14M  & $43$k ($17$k)$^{c}$ & --- & --- & --- & none surviving in tail \\
Pythia-31M  & $28$k & --- & --- & --- & L5 \\
Pythia-70M  & $a19000 \!\to\! a20000$ & $20$k & $20$k & $42.1\%$ @ L4/$84$k & L4 (mid and tail) \\
Pythia-160M & $71$k ($12$k)$^{c}$ & $13$k ($+72$k deep) & $20$k & $32.8\%$ @ L8/$92$k & L8--L11 (mid); L6/L7 (tail) \\
Pythia-410M & $25$k & $92$k & $48$k & $43.2\%$ @ L21/$116$k & L19--L23 (tail); L21--L23 (mid) \\
Pythia-1.4B & none in window$^{d}$ & not in window & not in window$^{e}$ & $3.0\%$ @ L15/$128$k & no collapse; band persists \\
\bottomrule
\end{tabular}

\smallskip
{\footnotesize
Onsets use persistence criteria rather than visual reading. Null-line onset:
$\mathrm{null}_{q05}<0.90$ for five consecutive anchors. Row pocket onset:
peak-layer share of rows with $\mathrm{FRR}<0.7$ above $5\%$ for three
consecutive anchors, on $D{=}512$ qkv rows, with row-length invariance checked
over $D\in[128,3072]$ on three sizes. Segment-level collapse onset: one anchor
after the last anchor at which the pooled excess reaches half its early-phase
mean, declared only when the tail-phase mean has returned to the null level.
The 70M collapse is a single-anchor step: head-level layer-mean excess
$0.304\to-0.029$, a change of $0.333$ that is $3.7\times$ the next largest in
the series, with layers above chance falling from $5$ to $1$; across it the real
head groups are unchanged ($\mathrm{FRR}_{p50}=0.994\pm0.001$) while the null
moves ($\mathrm{null}_{q05}$ from $0.994$ to $0.67$--$0.85$).
$^{c}$\,Where the series dips and recovers, the first crossing is given in
parentheses. $^{d}$\,1.4B has a mid-layer extinction window at $95$ to $118$k,
with the calibrated readout at L11/L15/L21/L23 falling to $0.02$--$0.27$ near
$96$k and recovering from about $120$k, but no global collapse.
$^{e}$\,The 1.4B row-level population reaches $3.0\%$ at the last anchor and is
still rising there.
\textbf{The two onset columns are different instruments and must not be
merged.} The null-line and row-pocket columns are read on the row-level lane and
order with size; the segment-level collapse column is read on the calibrated
pooled lane and does not ($70$M $20$k, $410$M $25$k, $31$M $28$k, $14$M $43$k,
$160$M $71$k). Any size-ordering statement names which instrument it uses.}

\vspace{1.0em}
{\centering (c) Model excluded from every cross-scale reading\par}
\vspace{0.3em}

\begin{tabular}{l rr c p{0.40\textwidth}}
\toprule
Excluded model & total & non-embed. & $L \times d_{\mathrm{model}}$ & Grounds for exclusion \\
\midrule
Pythia-1B & 1{,}011{,}781{,}632 & 805{,}736{,}448 & $16 \times 2048$ &
(i) the only model of the suite trained in bfloat16 upstream, following an
fp16 loss spike; (ii) depth not monotone in size, at $16$ layers between two
$24$-layer models; (iii) an upstream defect in revision \texttt{step116000},
which forces masked windows at $116$ to $121$k. \\
\bottomrule
\end{tabular}

\smallskip
{\footnotesize
Pythia-1B contributes to no row, ordering, or trend in this appendix. Its
released checkpoints are stored in fp16 like every other size, so the
difference is in the training arithmetic and not in the container of the
weights we read; the size of any residual effect on the probes is not measured
here. Reasons (i) to (iii) are independent facts and none is evidence for the
others.}

\caption{Cross-scale readings. \textbf{(a)} sizes, coverage, and the
external landmarks available per size. \textbf{(b)} the internal transition
readings, all under persistence criteria rather than visual reading; the
two onset instruments are different lanes and are not merged.
\textbf{(c)} the one model excluded from every cross-scale reading, with
its three independent grounds.}
\label{tab:appF_readings}
\end{table*}

\paragraph{Why Pythia-1B is excluded.}
The suite contains a seventh model in this range, and it is left out of
every cross-scale statement below for three reasons that are separate
facts and should not be conflated. First, it is the only model of the
suite trained in bfloat16 upstream, following a half-precision loss spike
late in its run; the released checkpoints are stored in fp16 like every
other size, so the difference is in the training arithmetic and not in the
container we read, and the size of any residual effect on the probes is
not measured here. Second, its depth is not monotone in size: at $16$
layers of width $2048$ it is shallower than both the $24\times1024$ model
below it and the $24\times2048$ model above it in parameter count, and the
depth and width reading of Appendix~\ref{app:crossscale_pockets} is
precisely a statement about depth and width. Third, one of its released
revisions carries the upstream defect described in
Appendix~\ref{app:pythia_coverage}, which forces masked windows across the
very region that Appendix~\ref{app:crossscale_events} discusses. The first
reason alone would keep it out of a scale-ordered comparison; the three
together make the exclusion unambiguous. One cost of the exclusion is
recorded where it falls due, in Appendix~\ref{app:crossscale_pockets}.

\paragraph{Reading discipline.}
The calibrated readout detects within-group coherent motion against a
membership-redrawn null whose threshold adapts to the local geometry.
Horizontal comparisons of it \emph{between segments} are therefore
downgraded, and only three comparison types are used: time shape at fixed
grouping, cross-size comparison under one grouping convention, and
between-layer comparison inside one segment. Sampled-preview
instrumentation exists for the larger sizes and no number here is taken
from it; where a full readout does not exist the cell is left empty rather
than filled.

\subsection{The Auxiliary--Bulk Partition Holds at Every Size}
\label{app:crossscale_auxbulk}

Figure~\ref{fig:appF_auxbulk} reads the partition of
Appendix~\ref{app:pythia_partition} at all six sizes under three probes,
one from each family. The separation is not a tendency that survives
aggregation: the median auxiliary series exceeds the median bulk series at
every one of the $123$ anchors, at all six sizes, under all three probes.
The smallest median gaps anywhere in the panel occur at 1.4B late in
training, at $0.036$ for ESA, $0.351$ for $1-\mathrm{FRR}$, and $0.115$ for
the DMD direction cosine.

\begin{figure*}[tp]
  \centering
  \includegraphics[width=\textwidth]{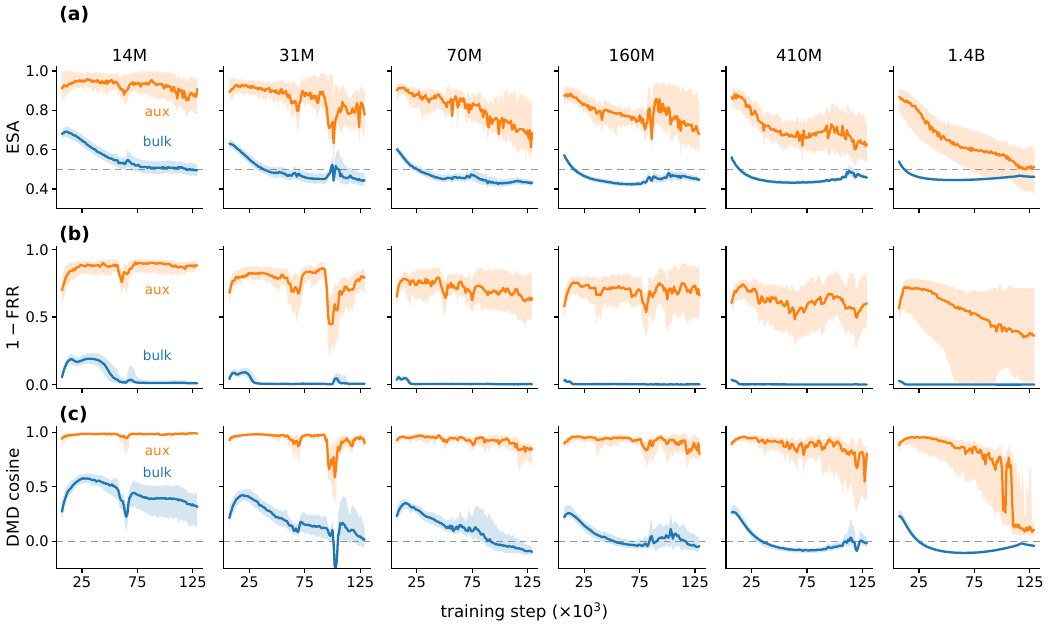}
  \caption{Auxiliary and bulk coordinates stay separated at every size and
  every anchor. Rows: (a) ESA, (b) $1-\mathrm{FRR}$, (c) DMD direction
  cosine; columns: the six sizes. The solid line is the group median over
  the pooled group set and the shaded band the interquartile range; orange
  is auxiliary, the vector segment, and blue is bulk. Grey dashed
  references mark $0.5$ in (a) and $0$ in (c). All panels use one readout
  protocol at $\tau{=}5$ over the $123$ shared anchors, and because the
  suite trains every size with the same $2$M-token batch, a matched step
  grid is also a matched token grid. Horizontal comparisons between sizes
  are therefore at matched steps and tokens, and not at matched loss,
  compute, or training progress. The median separation holds at every
  anchor at all six sizes under all three probes; the smallest median gaps
  are $0.036$, $0.351$, and $0.115$ for the three rows, all at 1.4B late in
  training. Interquartile ranges are disjoint at every anchor for (b) and
  (c) at all sizes, and for (a) at every size except 1.4B, where they
  overlap on $30\%$ of anchors. Bulk is the strict matrix set, with both
  embedding segments dropped; under the looser alternative the bulk means
  move by at most $1.2\times10^{-4}$ for (a) and (b) but by up to
  $6.3\times10^{-3}$ for (c), which is why the strict rule is used
  throughout. Pythia-1B is excluded here and in every other cross-scale
  panel, on the three independent grounds of
  Table~\ref{tab:appF_readings}(c).}
  \label{fig:appF_auxbulk}
\end{figure*}

The interquartile bands are disjoint at every anchor for the
subspace-residual and predictor-based readouts at all six sizes, and for
ESA at every size except 1.4B, where they overlap on $30\%$ of anchors in
the late window. The frozen-coordinate fraction is identically zero here,
at every segment and every anchor, so none of this contrast is carried by
coordinates that do not move.

Two conventions travel with the figure. Bulk is the strict matrix set,
$qkv$ together with the attention output projection and the MLP, with both
embedding segments dropped. Under the looser alternative that keeps
everything not in the vector segment, the bulk means move by at most
$1.2\times10^{-4}$ for ESA and $1-\mathrm{FRR}$ but by up to
$6.3\times10^{-3}$ for the DMD cosine, which is why the strict rule is
fixed throughout this appendix. And the calibrated readout does not enter
this figure at all; where it appears later it is reported as excess over
its null level.

\subsection{Segment Division of Labour}
\label{app:crossscale_segments}

\begin{figure*}[tp]
  \centering
  \includegraphics[width=\textwidth]{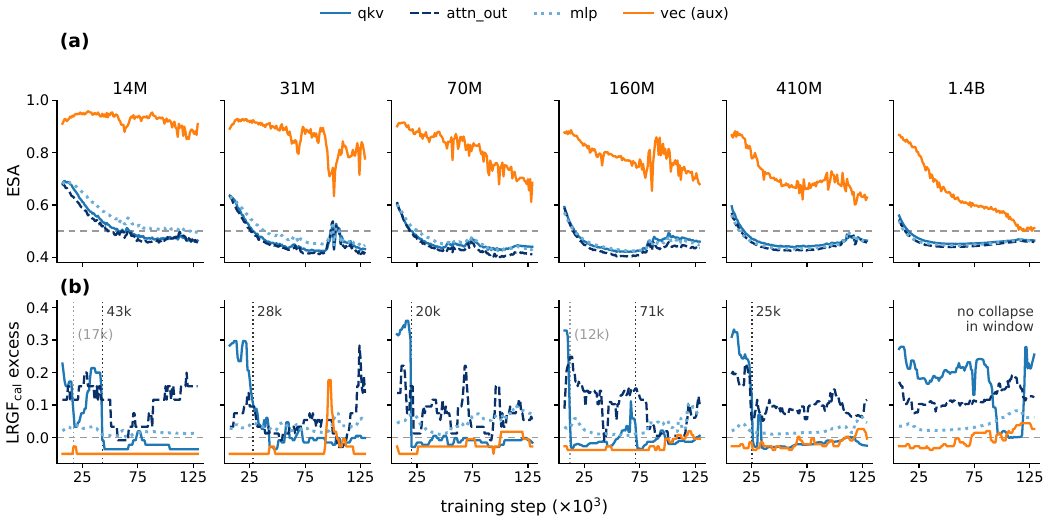}
  \caption{Segment division of labour across sizes: ESA barely separates
  the matrix segments while the calibrated readout does. Columns are the
  six sizes. (a) ESA group median per segment. (b) pooled calibrated excess
  per segment, drawn as a five-anchor centred rolling median, with the
  onset markers computed on the unsmoothed series. The three matrix
  segments share a blue system ($qkv$ solid, attention output dashed, MLP
  dotted) and the vector segment is orange; grey dashed references mark
  $0.5$ in (a) and the null level $0$ in (b), where the floor at $-0.05$ is
  the zero-fraction case. The dark dotted line is the sustained collapse
  onset of the $qkv$ series, and where the series dips and recovers the
  first crossing is given in grey parentheses. Onsets are $20$k (70M),
  $25$k (410M), $28$k (31M), $43$k (14M), and $71$k (160M), with no
  sustained collapse at 1.4B, so the ordering is not monotone in parameter
  count under either crossing. In (a) the three matrix segments stay within
  a median of $0.048$, $0.038$, $0.027$, $0.021$, $0.014$, and $0.011$ of
  one another in ascending size order, so the spread between matrix
  segments shrinks monotonically with size, while the vector segment stays
  above all of them by at least $0.034$ at every anchor and size.}
  \label{fig:appF_segment}
\end{figure*}

Figure~\ref{fig:appF_segment} puts the four segments of every size on one
axis under two readouts, and the two readouts do not agree about how much
structure there is to see.

Under ESA the three matrix segments are nearly indistinguishable, and they
become more so as the model grows: the median spread between them narrows
monotonically from $0.048$ at 14M to $0.011$ at 1.4B. The vector segment
stays above all three by at least $0.034$ at every anchor and every size.
Read through coordinate signs alone, the bulk of a larger model looks more
internally uniform, not less.

Under the calibrated readout the same segments separate clearly, and $qkv$
carries a distinct time shape: high early, then a sustained collapse to the
null level. The collapse onsets are $20$k at 70M, $25$k at 410M, $28$k at
31M, $43$k at 14M, and $71$k at 160M, with no sustained collapse at 1.4B
inside the window. That ordering is \emph{not} monotone in parameter
count, under either of the two crossing conventions recorded in
Table~\ref{tab:appF_readings}.

This is where the two instruments must be kept apart. The collapse just
described is read on the pooled calibrated lane. The null-line and
row-level pocket onsets of Appendix~\ref{app:pocket_emergence} are read on
the row-level lane, and those do order with size. A size-ordering
statement about this system is only meaningful once it names which of the
two it is about.

The corresponding segment-resolved pocket populations, including the
finding that the attention output projection carries the same wave as
$qkv$ at row granularity and that only the output half of the MLP carries
signal, are reported in Appendix~\ref{app:pocket_emergence} and are not
repeated here.

\subsection{Where the Calibrated Signal Survives}
\label{app:crossscale_pockets}

\begin{figure*}[tp]
  \centering
  \includegraphics[width=\textwidth]{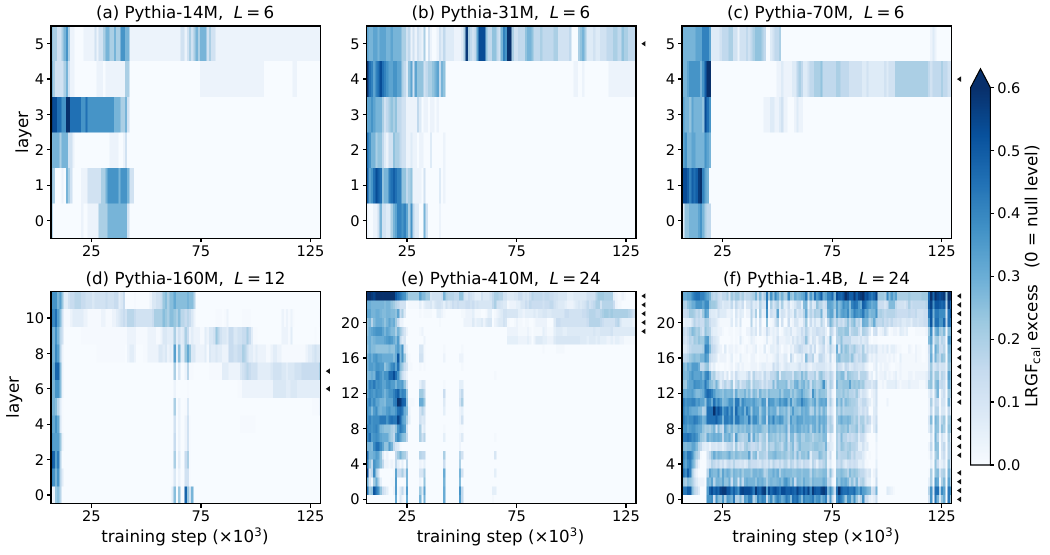}
  \caption{Where the $qkv$ calibrated signal survives: layer against step
  at six sizes on one colour scale. Each panel is one model, rows are
  transformer blocks with layer $0$ at the bottom, and columns are the
  $123$ shared anchors. The probe is reported as excess over its
  $\alpha{=}0.05$ null level, so $0$ is exactly the null and the raw pocket
  fraction is the excess plus $0.05$; the colour ramp is anchored with
  white at $0$, every cell at or below the null renders white, and the ramp
  is clipped at $0.60$, which affects $0.21\%$ of cells. Ticks on the right
  edge mark the layers whose tail-phase mean stays above the null. Early in
  training the signal covers $5/6$, $6/6$, $6/6$, $10/12$, $23/24$, and
  $23/24$ layers in ascending size order, and then contracts onto a
  persistent band whose absolute layer index moves deeper as the model
  deepens. 1.4B does not contract inside the window: $22$ of $24$ layers
  are still above the null in the tail phase, and it shows instead an
  extinction window near $96$ to $118$k followed by a recovery from about
  $120$k. The calibrated readout is identical under both bulk rules, since
  the vector segment does not enter it.}
  \label{fig:appF_depth}
\end{figure*}

Figure~\ref{fig:appF_depth} resolves the $qkv$ calibrated readout by layer
and step at all six sizes on one colour scale, with white at the null
level.

Early in training the signal is broad, covering $5/6$, $6/6$, $6/6$,
$10/12$, $23/24$, and $23/24$ layers at the six sizes in ascending order.
It then contracts onto a small persistent band whose absolute layer index
moves deeper as the model deepens: in the middle phase the band is layer 5
at 14M and 31M, layer 4 at 70M, layers 8 to 11 at 160M, and layers 21 to
23 at 410M. In the tail phase nothing survives at 14M, while the band sits
at layer 5 (31M), layer 4 (70M), layers 6 to 7 (160M), and layers 19 to 23
(410M). In relative depth the band sits in the deepest quarter of every
model that contracts at all.

1.4B does not contract inside the window: $22$ of its $24$ layers are
still above the null in the tail phase. It shows instead an extinction
window near $96$ to $118$k in which the raw fraction drops to nearly zero
in most layers, with only the deepest block holding $0.15$ to $0.44$,
followed by a recovery from about $120$k to the highest values of the run.

\paragraph{Depth and width.}
Parameter count alone does not order these transitions. Among the sizes
retained, the one contrast that survives holds depth fixed: at $24$ layers
the narrower 410M reaches its row-level pocket onset at $48$k and its
null-line onset at $92$k, while the wider 1.4B has reached neither by
$128$k. Wider at equal depth is later. The complementary contrast, deeper
at equal width, rested on the pair that the exclusion of 1B removes, so
depth and width are confounded along the suite's own scaling ladder among
the sizes retained. The depth half of a two-factor reading is therefore
registered as unsupported here rather than asserted.

\subsection{Co-location Registry for Late Events}
\label{app:crossscale_events}

This subsection records where the internal readouts and the public loss
telemetry land relative to each other on the step axis. Nothing here is a
directional claim.

At 70M the smoothed loss reaches its minimum at step $50{,}345$ and rises
afterwards, bulk ESA has a local rise peaking at $0.467$ near $77$k before
descending to its global minimum $0.426$ at $98$k, and the row-level main
wave peaks at $84$k. At 160M the loss minimum is at $76{,}409$, bulk ESA
reaches its global minimum $0.424$ at $68$k and rises persistently from
$78$k to a plateau of $0.45$ to $0.47$, and the pocket peak is at $92$k.
On both sizes the loss minimum, the ESA turn, and the main pocket wave
fall in the same stretch of training.

At 410M that coincidence does not hold, and this is the more informative
case. The smoothed loss has no minimum inside the anchor domain: it is
still descending at the right edge and remains flat or descending to the
end of training. The ESA readout nevertheless shows the same late rise, recomputed on the full-protocol extended panel: bulk group-level ESA ($\tau{=}5$, group $p_{50}$) sits on a $0.436$--$0.441$ plateau from $60$k to $85$k, climbs steadily from about $85$k, and peaks at $0.48$--$0.49$ over $111$ to $117$k. The late ESA
rise therefore occurs without a loss-bottoming event, and the co-location
seen at the two smaller sizes should not be carried to larger sizes as if
it were a general pairing.

\begin{figure*}[tp]
  \centering
  \includegraphics[width=\textwidth]{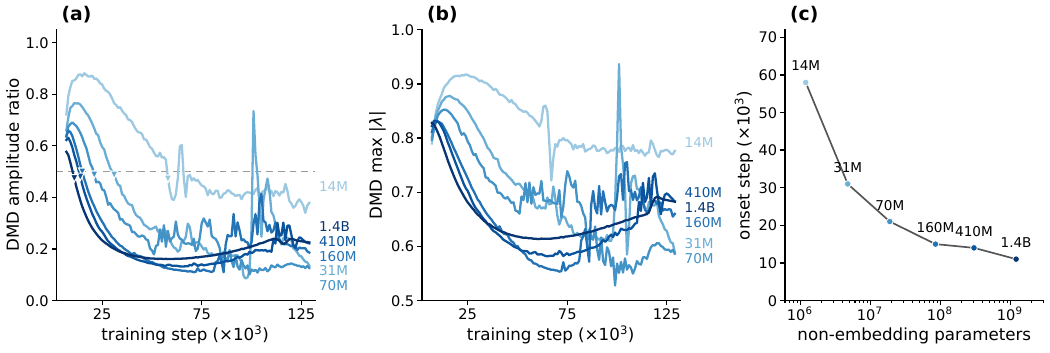}
  \caption{Internal predictor-family diagnostics on the bulk set across six
  sizes. (a) median predicted-over-realized step amplitude of the fit;
  (b) median largest eigenvalue modulus of the fitted operator; (c) the
  onset step of the amplitude deficit against non-embedding parameter count
  on a log axis. Line shade encodes size on a single ramp and every curve is
  additionally labelled at its right end, so colour is never the only
  encoding. Onset is the first anchor opening five consecutive anchors whose
  median amplitude ratio is below $0.5$, marked by the grey dashed reference
  in (a). The deficit sets in earlier the larger the model, at $58$k, $31$k,
  $21$k, $15$k, $14$k, and $11$k in ascending size order, which is the only
  cross-scale ordering in this appendix that is monotone in parameter count.
  By the last anchor the bulk median amplitude ratio has fallen to $0.13$ to
  $0.38$ at every size while the eigenvalue modulus settles into $0.59$ to
  $0.68$, with 14M the exception at $0.78$.}
  \label{fig:appF_dmd}
\end{figure*}

\paragraph{One monotone ordering.}
Figure~\ref{fig:appF_dmd} reports the internal diagnostics of the
predictor-based family on the bulk set. The fitted operator's step
amplitude falls progressively below the realized step, and the onset of
that deficit is earlier the larger the model: $58$k, $31$k, $21$k, $15$k,
$14$k, and $11$k across the six sizes in ascending order. This is the only
cross-scale ordering in this appendix that is monotone in parameter count.
By the last anchor the bulk median amplitude ratio has fallen to $0.13$ to
$0.38$ at every size while the largest eigenvalue modulus settles into
$0.59$ to $0.68$, with 14M the exception at $0.78$: the fitted operator
keeps a stable spectrum while the amplitude of the step it predicts falls
well below the step actually taken.

\paragraph{The late cluster.}
Several unrelated readouts move inside a narrow late window. On 70M and
160M the auxiliary DMD amplitude ratio falls to exactly zero and stays
there from anchors $118$k and $117$k respectively, while the bulk series
over the same anchors is unaffected. On 70M a secondary row-level wave
appears near $115$k. On 410M the ESA peak, the pocket peak at $116$k, and
an MLP output-half patch in the middle layers all fall in $110$ to $120$k.
Whether this is one phenomenon or several is open. Two constraints are
worth recording: the learning-rate schedule has no breakpoint anywhere
near it, since the decay is a single cosine with no constant tail, and the
auxiliary series that goes to zero does so exactly, which is the signature
of a degenerate fit rather than of a small value.

\subsection{Open Questions}
\label{app:crossscale_open}

Four items are registered as open rather than resolved.

The late cluster above may be one event or several, and the exact-zero
auxiliary amplitude ratio is as consistent with an instrument boundary as
with a property of the trajectory. Distinguishing the two requires a
readout that does not go through the same fit.

1.4B sits at the edge of the observation window. Its row-level population
reaches $3.0\%$ at the last anchor and is still rising, and its calibrated
band has not contracted. It is therefore a sample of a transition
beginning at the window edge, not of a transition that did or did not
happen, and it should not be read as a negative case.

The local 1.4B loss telemetry does not match the released $143{,}000$-step cosine schedule, so no loss landmark is quoted
for it and no co-location statement is made at that size.

Excluding 1B removes the only pair of sizes that share a width, so the
depth leg of the depth and width reading has no support among the sizes
retained. Recovering it would require either a fourth model at $24$ layers
and width $2048$ or an assessment of how much the training-arithmetic
difference actually moves these probes.

\section{Architecture- and Recipe-Conditioned Results}
\label{app:recipe_conditioning}

\mainref{sec:recipe} uses optimizer-family variation as its compact
example and defers the rest to this appendix: the size-controlled
architecture comparison, the controlled weight-decay and momentum dose
studies, the wider optimizer panel, and the comparison between
predictability and optimizer quality.

\subsection{Size-Controlled Architecture Comparison}
\label{app:architecture_controls}

\begin{figure}[tbp]
  \centering
  \includegraphics[width=0.86\columnwidth]{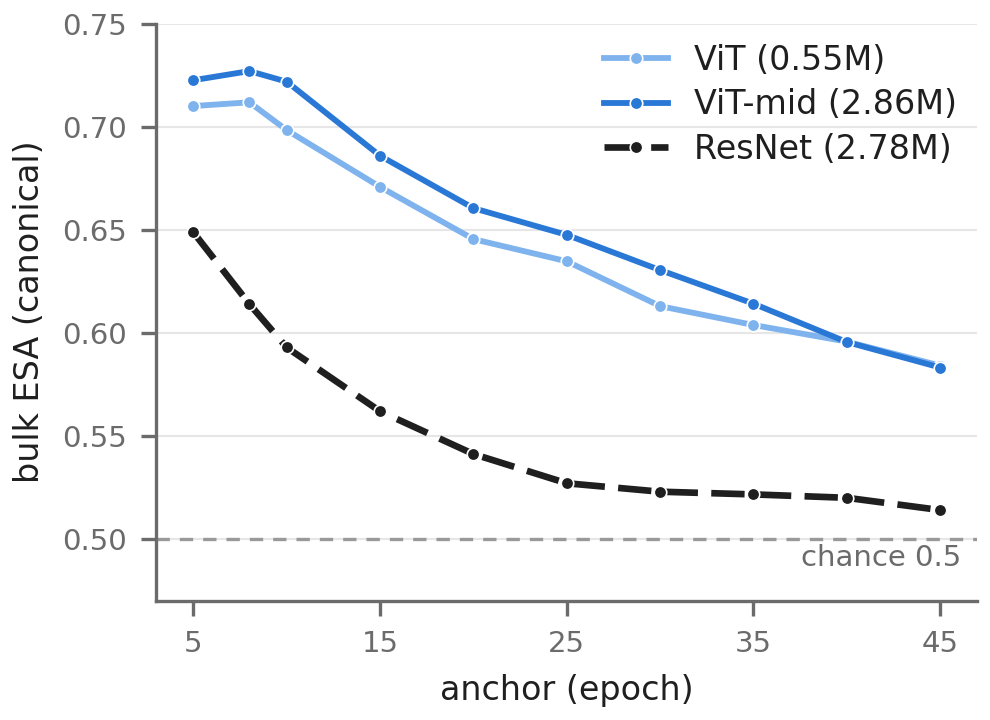}
  \caption{Size-controlled architecture comparison under SAM. Curves report
  canonical bulk ESA across ten anchors for ViT-tiny, ViT-mid, and the
  parameter-matched ResNet; ViT-mid and the ResNet have nearly equal
  parameter counts. The dashed line marks the random sign-agreement level
  of $0.5$. Each model uses its own standard learning rate, so the
  comparison controls scale but not every hyperparameter.}
  \label{fig:appG_arch}
\end{figure}

Figure~\ref{fig:appG_arch} compares three architectures trained on the
same dataset with SAM under a cosine schedule. ViT-mid and the ResNet have
nearly matched parameter counts, and ViT-mid retains higher bulk ESA at
every measured anchor. The separation is present from early training: at
epoch $5$ the two ViT scales read about $0.72$ and $0.71$ against $0.65$
for the ResNet, and around epoch $40$ the three read about $0.60$, $0.60$,
and $0.52$. The smaller ViT follows the same higher-ESA profile while
holding roughly one fifth as many parameters as ViT-mid, so parameter
count alone does not order the three.

Because each model uses its own standard learning rate, this comparison
does not isolate architecture from every hyperparameter difference. What
it establishes under the stated conditions is a stable
architecture-associated contrast: ViT-class models retain more bulk
directional persistence than the size-matched ResNet throughout the
measured trajectory.

\subsection{Optimizer Families along Training Loss}
\label{app:optimizer_panel}

Optimizer families traverse the loss trajectory at different rates, so
comparing them at matched epochs mixes the effect of the update rule with
the effect of elapsed training. \mainref{fig:opt_axis} therefore aligns
them on training loss and restricts the comparison to the loss range that
all displayed families share. Matched loss does not equate optimizer
state, and the alignment removes an elapsed-training confound rather than
all confounds.

Over the shared range the families occupy distinct profiles. SAM sits in
the highest regime, at roughly $0.58$ to $0.73$ bulk ESA over the
displayed interval. Muon stays near the $0.5$ sign-agreement reference.
Adam lies about $0.08$ below plain SGD over most of their overlap. SGD and
AdaGrad approach a similar low-loss level near $0.545$ despite following
different profiles earlier. Optimization position therefore organizes how
predictability evolves, while the update rule conditions its level and its
local directional structure.

The ordering is architecture-conditioned rather than universal. The
weight-decay and momentum ladders below are the more controlled
comparison, because each of their arms changes exactly one named
hyperparameter relative to plain SGD.

\subsection{Weight-Decay and Momentum Dose Ladders}
\label{app:recipe_dose}

\begin{figure*}[tp]
  \centering
  \includegraphics[width=0.94\textwidth]{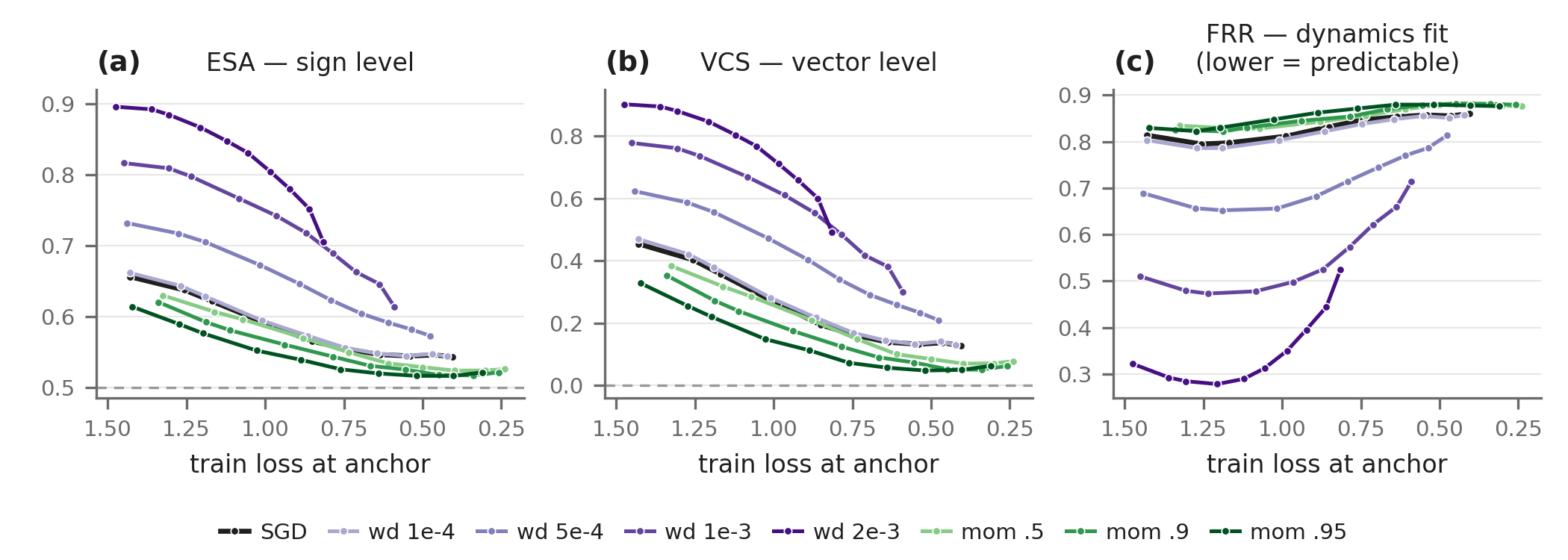}
  \caption{Controlled weight-decay and momentum dose ladders for ViT-mid.
  Panels report (a) bulk ESA, (b) bulk VCS, and (c) bulk FRR against
  reversed training loss under the canonical $K{=}16$ vision grouping. Each
  arm changes only the indicated weight-decay or momentum value relative to
  plain SGD, and curves are compared over their overlapping loss ranges.
  The two ladders order the three readouts in opposite directions, and the
  subspace-residual readout separates from plain SGD at a dose where the
  two displacement-direction readouts do not.}
  \label{fig:appG_dose}
\end{figure*}

Figure~\ref{fig:appG_dose} reports the two dose ladders for ViT-mid under
the canonical vision grouping, read through all three probe families
against reversed training loss. The ladders also show why the families are
retained jointly, since one intervention can register differently through
coordinate signs, vector direction, and containment in the recent-history
span.

\paragraph{Weight decay.}
The weight-decay ladder shows a consistent dose ordering on all three
readouts. ESA increases with the decay coefficient at shared loss
positions and approaches $0.90$ at the largest dose. VCS rises towards a
similar level, so the full displacement direction becomes more persistent,
and FRR falls to about $0.28$, so the future displacement is more strongly
contained in the span of recent updates.

The readouts differ in sensitivity at the low end. At
$\lambda=10^{-4}$, ESA and VCS remain near the plain-SGD profile while FRR
has already separated from it, at about $0.66$ against $0.79$. The
subspace-residual probe therefore resolves a low-dose change that the
displacement-direction probes do not. The mechanism is visible in the
update rule: SGD with weight decay adds a component proportional to
$-\lambda w$, which is a slow and coherent drift, and the monotone
response of all three readouts shows that this component is picked up at
the sign, vector, and subspace levels alike. These are raw probe profiles
over the realized trajectory, not a decomposition of the weight-decay
mechanism.

\paragraph{Momentum.}
The momentum ladder runs the other way. At shared loss positions, a larger
momentum coefficient depresses both displacement-direction readouts and
lifts FRR by a smaller amount. The response is clearest at the vector
level: VCS falls further than ESA, whose change is only about $-0.02$ to
$-0.03$.

Under this protocol higher momentum therefore reduces the forms of bulk
short-horizon redundancy these probes capture. The statement is about the
realized parameter trajectory and not about the smoothness of the
underlying gradient estimate; smoothing the gradient and making the
resulting parameter displacement predictable at short horizons are
distinct properties. Two common interventions can thus reorganize the
realized trajectory in opposite directions across the same probe panel.

Both ladders exclude the two collapsed momentum runs recorded in
Appendix~\ref{app:cifar_families}.

\subsection{Predictability Is Not an Optimizer Ranking}
\label{app:optimizer_quality}

\mainref{sec:recipe} cautions that higher predictability should not be
read as better optimization. Two measurements support that caution.

First, the two orderings are not aligned. Within each architecture, the
rank correlation between anchor-averaged bulk ESA and final test accuracy
across the measured optimizer cells is negative at every architecture, with
a median of $-0.71$ over the five architectures and a range from $-0.89$ to
$-0.20$. The cell counts behind those five correlations are $6$, $6$, $7$,
$7$, and $4$, so the weakest of them rests on four points and should not be
read on its own. Reading a higher probe value as a better optimizer would
therefore invert the accuracy ordering in these cells rather than
reproduce it. This is a correspondence between two summaries of the same
runs and is confounded in the obvious way, since different families reach
different loss positions in $50$ epochs; it is registered as a caution
against the ranking reading and not as a claim that predictability
predicts accuracy with either sign.

Second, the weight-decay ladder provides the same conclusion inside a
single controlled arm. Predictability increases monotonically along that
ladder, yet the $2\times10^{-3}$ arm terminates at a higher loss and
reaches about $0.707$ final accuracy against about $0.755$ for the
$10^{-3}$ arm. Temporal redundancy and training quality move in opposite
directions along a ladder in which only one hyperparameter changes.

\paragraph{Scope.}
These comparisons are representative rather than exhaustive. The
architecture and optimizer panels use the standard configuration of each
model, and the dose ladders cover one ViT-mid setting. What they establish is
not a ranking of architectures or of update rules, but that the three
probe families resolve, and resolve differently, the changes that model
structure and training recipe impose on the realized trajectory.

\section{Online Selective Prediction}
\label{app:online}

\mainref{sec:empirical} is retrospective throughout: every probe reads a
stored trajectory after the fact. The conclusion of the main paper points
to one narrow online check of whether the measured spatial organization is
useful as a prior over where to attempt prediction during training. This
appendix gives that check in full.

\begin{table*}[tp]
\centering
\small

\begin{tabular}{@{}llccc@{}}
\toprule
Architecture & Arm & Test accuracy & Gain (pp) & Accepted \\
\midrule
ViT-tiny & baseline & $0.7422 \pm 0.0057$ & --- & --- \\
 & full scope & $0.7484 \pm 0.0018$ & $+0.62 \pm 0.43$ & $23.9\%$ \\
 & auxiliary only & $\mathbf{0.7516 \pm 0.0058}$ & $\mathbf{+0.94 \pm 0.09}$ & $\mathbf{0.80\%}$ \\
\midrule
ViT-mid & baseline & $0.7425 \pm 0.0032$ & --- & --- \\
 & full scope & $0.7498 \pm 0.0032$ & $+0.73 \pm 0.18$ & $23.2\%$ \\
 & auxiliary only & $\mathbf{0.7511 \pm 0.0029}$ & $\mathbf{+0.87 \pm 0.09}$ & $\mathbf{0.32\%}$ \\
\bottomrule
\end{tabular}

\caption{Online comparison on CIFAR-10 over three seeds under pure SGD,
with momentum and weight decay both zero. All three arms share the same
predictor and the same coordinate-wise acceptance mask; the arms differ
only in which parameters are eligible for write-back. ``Accepted'' is the
fraction of parameters that actually received a written-back forecast.
Values are means over seeds with one standard deviation, and each arm is
read at the epoch chosen on a held-out split within its own run.}
\label{tab:appH_pdt}
\end{table*}

\subsection{Protocol and Selection Rule}
\label{app:online_protocol}

The online method writes a forecast back into a coordinate only when that
forecast passes a coordinate-wise acceptance test on scale control and
dynamic consistency; coordinates that fail continue to follow the
baseline optimizer. That acceptance mask is the immediate safeguard and is
unchanged in every arm below.

The variable under test is not the mask but the \emph{eligibility set} it
is applied to. Two settings are compared against ordinary training. The
full-scope arm makes every trainable parameter eligible. The restricted
arm keeps the identical predictor and identical acceptance mask but makes
only auxiliary parameters eligible, in the exact sense of
Appendix~\ref{app:partition_rule}. The restriction is deliberately coarse:
it uses the role-level separation that the retrospective measurements
report, and it makes no attempt to detect the localized, time-varying
pockets inside bulk tensors online. Nothing in the eligibility rule reads
the future, and nothing reads test labels.

The comparison uses ViT-tiny and ViT-mid under pure SGD, with momentum and
weight decay both set to zero, over three seeds. Each arm reports the test
accuracy at the epoch chosen on a held-out split of the training data, and
the gain is taken against the baseline of the same seed.

\subsection{Multi-Seed Results}
\label{app:online_results}

\begin{figure}[tbp]
  \centering
  \includegraphics[width=\columnwidth]{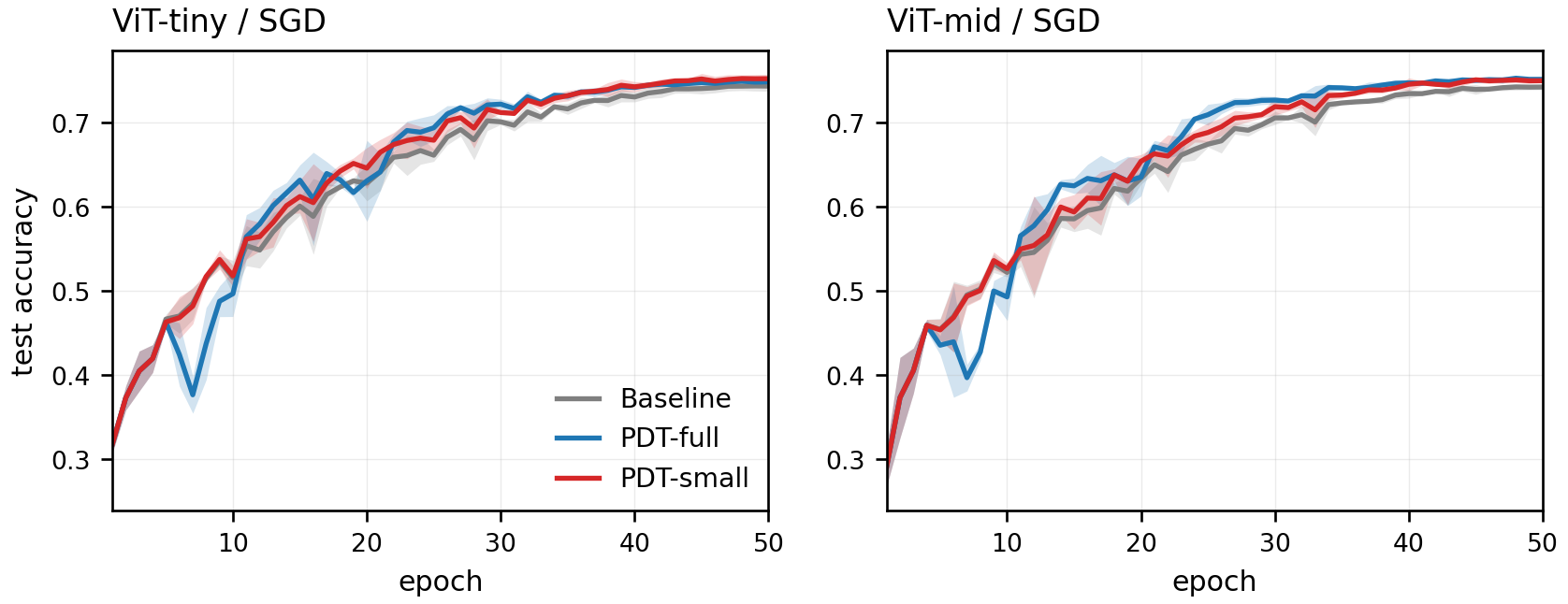}
  \caption{Multi-seed online comparison under pure SGD for ViT-tiny and
  ViT-mid. Curves show seed means with $\pm1$ standard-deviation bands for
  ordinary training, full-scope eligibility, and auxiliary-only
  eligibility.}
  \label{fig:appH_pdt}
\end{figure}

Table~\ref{tab:appH_pdt} and Figure~\ref{fig:appH_pdt} give the
comparison. On both architectures the restricted arm produces the larger
mean gain while accepting write-back for less than $1\%$ of parameters,
$0.80\%$ on ViT-tiny and $0.32\%$ on ViT-mid. The full-scope arm accepts
about $24\%$ of parameters and yields smaller mean gains, $+0.62$
percentage points on ViT-tiny and $+0.73$ on ViT-mid. The restricted arm also shows the lower across-seed variability: its standard deviation is $0.09$ percentage points at both scales, against $0.18$ to
$0.43$ for the full-scope arm.

The improvement is therefore not obtained by predicting more. Within these
two cells, narrowing eligibility to the region that the retrospective
measurements identify as higher-confidence gives a larger and more stable
gain than making everything eligible, despite an accepted fraction roughly $30$- and $70$-fold smaller.

\subsection{What This Does and Does Not Show}
\label{app:online_scope}

The result is consistent with the auxiliary--bulk separation reported in
\mainref{sec:heterogeneity}, and that is the extent of the claim. Four
limits travel with it.

It does not isolate the role restriction as the causal source of the
improvement, since the two arms differ in accepted fraction as well as in
eligibility. It does not show that auxiliary parameters are uniformly
predictable, only that the region is higher-confidence on average. It is
not a wall-clock acceleration result, because no accounting of predictor
overhead is attempted here. And it covers two model--training cells at one
optimizer setting, so it is an initial indication rather than a
demonstration that the measurement framework transfers to online use.

The coordinate-wise acceptance mask, not the role restriction, remains the
operative safeguard in every arm. The static role-level eligibility rule
implements neither the retrospective probe panel nor any detection of the
localized pockets inside bulk tensors that
Appendix~\ref{app:pocket_persistence} describes.

\end{document}